\documentclass[opre,nonblindrev]{informs3_hide}

\OneAndAHalfSpacedXI % current default line spacing
\usepackage{amsmath,amsfonts,amssymb}
\usepackage{mathtools}
    \allowdisplaybreaks
\usepackage{bm}
\usepackage[mathscr]{euscript}
\usepackage{enumitem}
\usepackage{xcolor}
\usepackage{graphicx}
\usepackage{multirow}
\usepackage{booktabs}
\usepackage{microtype}
\usepackage{fix-cm}
\usepackage[ruled,vlined]{algorithm2e}
\usepackage{physics}
\usepackage{xfrac}
\usepackage{nicematrix}

\DeclareMathOperator{\pr}{\mathbb P}
\DeclareMathOperator{\E}{\mathbb E}
\DeclareMathOperator{\Var}{\mathrm{Var}}
\DeclareMathOperator{\Cov}{\mathrm{Cov}}

\DeclareMathOperator{\ind}{\mathbb I}
\DeclareMathOperator{\supp}{\mathrm supp}

\newcommand{\subG}{\mathsf{subG}}
\newcommand{\Real}{\mathbb R}

\newcommand{\NatInt}{\mathbb N}
\newcommand{\SFc}{\mathsf c}

\newcommand{\SFp}{\mathsf p}
\newcommand{\SFq}{\mathsf q}
\newcommand{\SFB}{\mathsf B}
\newcommand{\SFG}{\mathsf G}

\newcommand{\ScrB}{\mathscr B}

\newcommand{\ScrF}{\mathscr F}
\newcommand{\ScrG}{\mathscr G}
\newcommand{\ScrH}{\mathscr H}
\newcommand{\ScrI}{\mathscr I}

\newcommand{\ScrX}{\mathscr X}
\newcommand{\CalA}{\mathcal A}
\newcommand{\CalE}{\mathcal E}

\newcommand{\CalH}{\mathcal H}

\newcommand{\CalK}{\mathcal K}
\newcommand{\CalL}{\mathcal L}
\newcommand{\CalN}{\mathcal N}
\newcommand{\CalO}{\mathcal O}
\newcommand{\CalS}{\mathcal S}

\newcommand{\CalX}{\mathcal X}

\usepackage{natbib}
 \bibpunct[, ]{(}{)}{,}{a}{}{,}%
 \def\bibfont{\small}%
 \def\BIBand{and}%

\TheoremsNumberedThrough     % Preferred (Theorem 1, Lemma 1, Theorem 2)
\ECRepeatTheorems

\EquationsNumberedThrough    % Default: (1), (2), ...
\begin{document}
%%%%%%%%%%%%%%%%

% Outcomment only when entries are known. Otherwise leave as is and
%   default values will be used.
%\setcounter{page}{1}
%\VOLUME{00}%
%\NO{0}%
%\MONTH{Xxxxx}% (month or a similar seasonal id)
%\YEAR{0000}% e.g., 2005
%\FIRSTPAGE{000}%
%\LASTPAGE{000}%
%\SHORTYEAR{00}% shortened year (two-digit)
%\ISSUE{0000} %
%\LONGFIRSTPAGE{0001} %
%\DOI{10.1287/xxxx.0000.0000}%

% Author's names for the running heads
% Sample depending on the number of authors;
% \RUNAUTHOR{Jones}
% \RUNAUTHOR{Jones and Wilson}
% \RUNAUTHOR{Jones, Miller, and Wilson}
% \RUNAUTHOR{Jones et al.} % for four or more authors
% Enter authors following the given pattern:
\RUNAUTHOR{Ding, Tuo, and Zhang}

% Title or shortened title suitable for running heads. Sample:
% \RUNTITLE{Bundling Information Goods of Decreasing Value}
% Enter the (shortened) title:
\RUNTITLE{High-Dimensional Simulation Optimization}

% Full title. Sample:
% \TITLE{Bundling Information Goods of Decreasing Value}
% Enter the full title:
\TITLE{High-Dimensional Simulation Optimization via Brownian Fields and Sparse Grids}

% Block of authors and their affiliations starts here:
% NOTE: Authors with same affiliation, if the order of authors allows,
%   should be entered in ONE field, separated by a comma.
%   \EMAIL field can be repeated if more than one author
\ARTICLEAUTHORS{%
\AUTHOR{Liang Ding, Rui Tuo}
\AFF{Department of Industrial and Systems Engineering, Texas A\&M  University, College Station, TX 77843, U.S., \EMAIL{ldingaa@tamu.edu}, \EMAIL{ruituo@tamu.edu}}
\AUTHOR{Xiaowei Zhang}
\AFF{Faculty of Business and Economics, The University of Hong Kong, Pokfulam Road, Hong Kong S.A.R., \EMAIL{xiaoweiz@hku.hk} }
% Enter all authors
} % end of the block

\ABSTRACT{%
High-dimensional simulation optimization is notoriously challenging. We propose a new sampling algorithm that converges to a global optimal solution and suffers minimally from the curse of dimensionality. The algorithm consists of two stages. First, we take samples following a sparse grid experimental design and approximate the response surface via kernel ridge regression with a Brownian field kernel. Second, we follow the expected improvement strategy---with critical modifications that boost the algorithm's sample efficiency---to iteratively sample from the next level of the sparse grid. Under mild conditions on the smoothness of the response surface and the simulation noise, we establish upper bounds on the convergence rate for both noise-free and noisy simulation samples. These upper bounds deteriorate only slightly in the dimension of the feasible set, and they can be improved if the objective function is known to be of a higher-order smoothness. Extensive numerical experiments demonstrate that the proposed algorithm dramatically outperforms typical alternatives in practice.
% Enter your abstract
}%

% Sample
%\KEYWORDS{deterministic inventory theory; infinite linear programming duality;
%  existence of optimal policies; semi-Markov decision process; cyclic schedule}

% Fill in data. If unknown, outcomment the field
% \HISTORY{}

\maketitle
%%%%%%%%%%%%%%%%%%%%%%%%%%%%%%%%%%%%%%%%%%%%%%%%%%%%%%%%%%%%%%%%%%%%%%

% Samples of sectioning (and labeling) in OPRE
% NOTE: (1) \section and \subsection do NOT end with a period
%       (2) \paragraph and lower need end punctuation
%       (3) capitalization is as shown (title style).
%
%\section{Introduction.}\label{intro} %%1.
%\subsection{Duality and the Classical EOQ Problem.}\label{class-EOQ} %% 1.1.
%\subsection{Outline.}\label{outline1} %% 1.2.
%\paragraph{Cyclic Schedules for the General Deterministic SMDP.}
%  \label{cyclic-schedules} %% 1.2.1
%\section{Problem Description.}\label{problemdescription} %% 2.

% Text of your paper here

\section{Introduction}

Decision-making problems in  management science, operations research, and machine learning,
especially those that have arisen in digital economics
with the recent explosive growth of data and
the rapid development of computing technologies,
are becoming increasingly large-scale.
The decisions that are involved in real-word scenarios are often high-dimensional.
For example, modern inventory management may be concerned with a large number of products that
require common raw materials for production or
share storage facilities \citep{MieghemRudi02}.
To minimize operating cost, managers must determine the production capacity or
order quantity jointly for each of the hundreds or even thousands of products \citep{ZhangMeiserLiuBonnerLin14}.

For another example, consider the fast-growing field of automated machine learning  \citep{AutoMLbook}.
The predictive power of many sophisticated machine learning methods is usually sensitive to a plethora of design choices,
such as the regularization parameter of the support vector machine model and
the learning rate of the stochastic gradient descent algorithm used for training a deep neural network \citep{SnoekLarochelleAdams12}.
Optimizing these  hyperparameters in a principled, algorithmic manner can save tremendous human effort and improve the reproducibility of analytical studies.

Numerous stochastic systems, including the two examples above, share a common feature.
Having no analytical form,
the objective function can be estimated based only on noisy samples from a simulation model, so the computational cost of each function evaluation is non-negligible or even substantial.
Simulation optimization (SO) is an effective approach to solving such decision-making problems.
SO algorithms and their design principles are vastly different,
depending on the nature of the  feasible set (finite, integer-ordered, or continuous) and on the nature of the solution sought (local optimal or global optimal).
In the present paper, we are concerned with SO problems involving continuous decision variables, and we seek algorithms converging to a global optimum, hereafter referred to as \emph{globally convergent} algorithms.

Surrogate-based methods have become increasingly popular for continuous global SO in recent years.
They typically begin with postulating a statistical model, referred to as \emph{surrogate} or \emph{metamodel}, to approximate the objective function---that is, the response surface representing the input-output relationship of the simulation model.
The surrogate is calibrated against simulation samples and then used to locate promising regions that may contain the optimal solution.
The search process is often iterative; that is,
it alternates between updating the surrogate and running the simulation model at suggested locations to generate additional samples.
Currently, a common practice is to employ a Gaussian process (GP) as the surrogate and determine subsequent samples via optimizing a metric that quantifies the trade-off between exploitation and exploration.
See \cite{HongZhang21} for a recent tutorial on this methodology.

Despite the remarkable success of surrogate-based methods and numerous alternatives \citep{AmaranSahinidisShardaBury16},
solving high-dimensional SO problems for global optima remains a challenge.
The curse of dimensionality is mainly manifested in the form of two challenges.
One is the \emph{statistical challenge}, stemming from the fact that estimating an unknown function in high dimensions is inherently difficult.
The number of samples needed to achieve a prescribed estimation accuracy generally grows exponentially with the dimension of the feasible set \citep{GyorfiKohlerKrzyzakWalk02}.
Another manifestation of the curse is the \emph{computational challenge}, which may arise both
in fitting a GP surrogate to a large number of samples \citep[Chapter~8]{RasmussenWilliams06} and
in identifying the best location for the next sample over the feasible set, which itself may amount to a high-dimensional, non-convex optimization problem \citep{HennigSchuler12}.
This paper addresses both challenges.

A typical strategy to tackle high-dimensional SO problems in recent years has been to impose a low-dimensional structure on the response surface and then leverage  the low \emph{effective dimensionality} to increase the search efficiency.
Nevertheless, such an assumption is often overly restrictive, for many practical SO problems are intrinsically high-dimensional.
Algorithms that rely on the assumption of low effective dimensionality may perform poorly for such problems \citep{MathesenChandrasekarLiPedrielliCandan19}.
The present paper follows a distinctive approach, assuming and exploiting certain tensor structures of the response surface, which allows a much broader scope of applications.

\subsection{Main Contributions}
First and foremost,
we propose a novel algorithm for continuous global SO that largely circumvents the curse of dimensionality.
Our algorithm consists of two stages.
In Stage~1, we take samples according to an initial experimental design to form a first approximation of the response surface.
In Stage~2, we use that approximation to construct for the response surface a GP prior with the approximation and  then follow the  criterion of
\emph{expected improvement} (EI)---which measures the gain one would obtain by sampling a new location relative to the current best solution---to iteratively choose the next sampling location, while updating the posterior distribution of the surface.

The performance of our algorithms hinges on several features as follows.
First, Stage~1 plays a vital role in our algorithm, as opposed to simply that of a ``warm-up''.
Indeed, it largely determines the convergence rate of our algorithm.
This is achieved by (i) setting the initial experimental design to be a \emph{sparse grid} \citep{bungartz_griebel_2004} and (ii) estimating the response surface via \emph{kernel ridge regression} (KRR) \citep[Chapter~6]{RasmussenWilliams06}  with a regularization parameter that is judiciously chosen.
Second, we use a Brownian field kernel
instead of the commonly used Gaussian kernels or Mat\'ern kernels,
in both the KRR in Stage~1 and the GP prior in Stage~2.
Third, when optimizing the EI criterion in each iteration of Stage~2,
the candidate solutions derive from a sparse grid instead of the entire continuous feasible set.
The first two aspects address the high-dimensional statistical challenge, while the third addresses the computational challenge.

Our second contribution is that we establish upper bounds on the convergence rate of the mean absolute error of the proposed algorithm, for both deterministic and stochastic simulation models, under the premise that the response surface lies in the reproducing kernel Hilbert space (RKHS) induced by a Brownian field kernel.
This is essentially equivalent to a mild assumption of the smoothness of the response surface.
We show that the dimensionality $d$ of the feasible set takes effect on the upper bounds only through
the exponent of $\log n$, rather than through the exponent of $n$, as is typical  \citep{YakowitzLEcuyerVazquez00,ChiaGlynn13}, where $n$ is the sample size.
Hence, the upper convergence rates deteriorate only slightly in $d$.
These rates can be further improved if stronger smoothness conditions  are imposed.
This indicates that the proposed algorithm is robust relative to model misspecification---the scenario where the response surface has a higher degree of smoothness than that induced by the Brownian field kernel.

Our third contribution is that we show, via extensive numerical experiments, that the proposed algorithm substantially outperforms the state-of-the-art approaches to solving SO problems involving as many  as 100 dimensions.
Because the performance guaranteed  by the convergence rate analysis is asymptotic, and because it may be nontrivial to verify the relevant technical conditions,
the experiments demonstrate that the proposed algorithm is indeed a practicable option for high-dimensional SO.

Lastly, in the process of analyzing the convergence rate of the proposed algorithm,
we develop a series of new technical results related to Brownian field kernels and sparse grids (see the e-companion).
Prominent examples include (i) an equivalence between the RKHS induced by a Brownian field kernel and a Sobolev-type space that is defined via function smoothness,
and (ii) an inequality that connects various function norms---including $L^2$ norm, RKHS norm, and empirical semi-norm.
These technical results are interesting in their own right and may be used to
facilitate future research that involves Brownian field kernels and sparse grids.

\subsection{Related Work}

The literature on SO is extensive.
Both locally convergent and globally convergent algorithms have been well developed for continuous SO problems.
We do not attempt to explore the former in detail, other than to mention some introductory materials, due to a lack of space and the present paper's focus on global optima.
Most of the locally convergent algorithms are gradient-based and can be classified into two categories:
stochastic approximation \citep{ChauFu15}  and sample average approximation \citep{KimPasupathyHenderson15}.
For both, a stochastic gradient estimator is used to guide the process of searching for better solutions. See \cite{Fu15} for a survey on gradient estimation.

There are a great variety of surrogate-based methods.
Common surrogates include radial basis functions and artificial neural networks \citep{BartonMeckesheimer06}, but the adoption of GPs has become prevalent thanks to their analytical tractability and their ability to  provide uncertainty quantification.
In particular, GPs permit easy updating of the posterior distribution of the response surface, which assists in selecting the next sampling location.
A great variety of selection strategies have been proposed in the literature, under the general umbrella of \emph{Bayesian optimization} (BO), including
EI \citep{jones1998}, knowledge gradient \citep{ScottFrazierPowell11},
probability of improvement \citep{SunHuHong18},
upper confidence bound (UCB) \citep{SrinivasKrauseKakadeSeeger12},  and
entropy search \citep{HennigSchuler12}.
One can even construct a \emph{meta-strategy}---a portfolio of selection strategies---to guide the search, because no single strategy outperforms the others in all problem instances \citep{HoffmanBrochuFreitas11}.
We refer to \cite{Frazier18} for a recent overview.

There are two general approaches to scaling up BO to high-dimensional spaces, both of which are agnostic to the selection strategy for determining the next sampling location.
The first is to postulate the existence of a low-dimensional embedding (i.e., the response surface evolves depending on a small set of features), search for sampling locations in the low-dimensional subspace, and then project back to the original space for sampling \citep{WangHutterZoghiMathesonFeitas16}.
The embedding is usually assumed to be linear \citep{BinoisGinsbourgerRoustant20}, but it can also be  nonlinear to cope with potentially complex optimization constraints \citep{JaquierRozo20}.
The second approach also relies on dimensionality reduction.
It assumes that the response surface can be decomposed into a set of low-dimensional components, then treats each component separately.
This effectively breaks down a high-dimensional problem into several low-dimensional problems. See, e.g., \cite{KandasamySchneiderPoczos15} and \cite{RollandScarlettBogunovicCevher18}.

In addition to surrogate-based methods, random search methods constitute another main class of globally convergent algorithms for continuous SO.
Their critical feature is to generate a population of potential candidates at each iteration from a probability distribution that is increasingly concentrated around the optimal solution.
An incomplete list of recent examples includes
model reference adaptive search \citep{hu2007model},
adaptive search with resampling \citep{AndradottirPrudius10},
gradient-based adaptive stochastic search \citep{ZhouBhatnagar18}, and
single observation search \citep{KiatsupaibulSmithZabinsky18}.
We refer to \cite{Andradottir15} and \cite{Zabinsky15} for reviews of random search methods.

While it is standard practice to prove convergence for continuous global SO algorithms,
results of the rate of convergence are available for very few of them.
\cite{ChiaGlynn13} fully characterize the convergence rate of pure random search, deriving the limit distribution of the estimator.
\cite{Bull11} investigates the convergence rate of  the EI algorithm
in a noise-free context.
GP-UCB has been extensively studied in machine learning literature.
Various upper bounds on its convergence rate are established. See, e.g., \cite{SrinivasKrauseKakadeSeeger12} and \cite{JanzBurtGonzalez20}.
These results are mostly developed under the assumption that the response surface has a certain smoothness that is induced by Gaussian kernels or Mat\'ern kernels.
Under the same setting,
\cite{Singh21} proves
minimax lower bounds on the convergence rate of an arbitrary algorithm for continuous global SO,
characterizing the intrinsic difficulty of optimizing a black-box function via noisy samples.
The lower bounds suggest that, in general, the convergence rate of a globally convergent algorithm deteriorates quickly as the dimension increases, unless the response surface is extremely smooth or an additional structure can be imposed and exploited.

% \smallskip
The remainder of this paper is organized as follows.
In Section~\ref{sec:formulation}, we formulate the SO problem, highlight the high-dimensional challenges, and discuss our algorithm design principles.
In Section~\ref{sec:Brownian} and Section~\ref{sec:sparse-grids}, respectively, we  overview the two main tools---Brownian fields and sparse grids---that are used for algorithm design and its asymptotic analysis.
We present our algorithm in Section~\ref{sec:algo} and analyze its rate of convergence in Section~\ref{sec:analysis}.
We conduct extensive numerical experiments with high-dimensional examples in Section~\ref{sec:num}
and conclude in Section~\ref{sec:conclusions}.
Additional technique results are collected in the appendix, and all proofs are provided in the e-companion to this paper.

\section{Problem Formulation}\label{sec:formulation}

The present paper concerns solving problems of the form
\begin{equation}\label{eq:SO}
\max_{\BFx\in\ScrX\subset \Real^d} \E[F(\BFx)],
\end{equation}
where $\BFx=(x_1,\ldots,x_d)\in\Real^d$ denotes the decision variable,
$\ScrX$ is the feasible set,
and $F(\BFx)$ represents the random output of a simulation model evaluated at $\BFx$.
Let $f(\BFx)\coloneqq \E[F(\BFx)]$ denote the \emph{response surface} of the simulation model.
In general,  the distribution of $F(\BFx)$ is unknown and $f(\BFx)$ has no analytical form.
But running simulation experiments can generate independent samples of $F(\BFx)$, denoted by $y(\BFx)$:
\begin{equation}\label{eq:observations}
y(\BFx) = f(\BFx)  + \varepsilon(\BFx),
\end{equation}
where $\varepsilon(\BFx)$ is the  zero-mean simulation noise at $\BFx$, and its distribution may depend on $\BFx$.

We are tasked with finding a globally optimal solution
$\BFx^*$ to problem~\eqref{eq:SO},
and we must do so subject to a simulation budget $N$ because running simulation models is often costly.
The goal is to develop a sampling algorithm $\pi$---which determines a sequence of design points $\{\BFx_1,\ldots,\BFx_N\}$ at which the simulation model is executed---to
learn the response surface $f$ over time and construct an estimate of $\BFx^*$, denoted by $\widehat{\BFx}^*_N$, upon termination of the sampling process.
We measure the performance of $\pi$ by the expected \emph{optimality gap}
\begin{equation}\label{eq:OG}
\E[f(\BFx^*) - f(\widehat{\BFx}^*_N)],
\end{equation}
where the expectation is taken with respect to the distribution of the samples
$\{y(\BFx_1),\ldots,y(\BFx_N)\}$
that are  generated by the algorithm $\pi$.
Throughout this paper, we impose the following assumptions.

\begin{assumption}\label{assump:optimum}
$\ScrX = (0,1)^d$, and $f$ has a global maximum $\BFx^*\in\ScrX$.
\end{assumption}

\begin{assumption}\label{assump:sub-Gaussian}
For any $n\in\NatInt$ and any sequence of design points $\{\BFx_{i}\}_{i=1}^n\subset\ScrX$, the noise terms $\{\varepsilon(\BFx_i)\}_{i=1}^n$ are independent zero-mean sub-Gaussian random variables with variance proxy $\sigma^2$, denoted by $\subG(\sigma^2)$.
That is, $\E\bigl[e^{t\varepsilon(\BFx_i)} \bigr] \leq e^{t^2\sigma^2/2}$ for all $t\in\Real$ and $i=1,\ldots,n$.
\end{assumption}

Typical examples of $\subG(\sigma^2)$ random variables include bounded random variables and normal random variables.
If a random variable $X$ is normal, then $\sigma^2$ may be taken as $\Var[X]$.
In general, however, $\sigma^2$ is not identical to but is rather an upper bound of $\Var[X]$.
Thus, under Assumption~\ref{assump:sub-Gaussian},
the simulation noise is allowed to be heteroscedastic, but $\Var[\varepsilon(\BFx)] \leq \sigma^2$ for all $\BFx\in\ScrX$.

\subsection{Two Challenges in High Dimensions}\label{sec:challenges}

We are particularly interested in solving problem~\eqref{eq:SO} in high dimensions---for example, $d>10$.
The task is demanding due to two essential challenges, with one being \emph{statistical} while the other \emph{computational}.
First, to estimate a global optimum with high confidence,
one needs to learn the response surface globally with high confidence.
Users of a simulation model usually have little prior knowledge  about $f$,
and thus impose minimal assumptions on its form to estimate it in a nonparametric fashion.
For example, it is considered restrictive to assume $f$ to be a quadratic function or a linear combination of a set of basis functions.
Nonparametric estimation of an unknown function in high dimensions, however, generally suffers from the curse of dimensionality;
that is, the sample complexity---the number of samples necessary for estimating the function as a whole to a prescribed level of accuracy---grows exponentially with the dimensionality.
The issue is further exacerbated by the high cost of simulation samples.
We refer to \cite{GyorfiKohlerKrzyzakWalk02} for an introduction to nonparametric estimation and to \cite{Singh21} for a recent discussion that reveals the statistical challenge of global optimization of unknown functions with noisy samples.

Suppose the simulation budget is large enough for a high-dimensional problem.
Then, one often needs to address a second---somewhat less severe, but still substantial---challenge.
Namely, it can be computationally burdensome to (i) process a large number of samples for constructing an estimate of $f$ and (ii) select subsequent design points based on that estimate.
For an example of the former, consider stochastic kriging \citep{AnkenmanNelsonStaum10}, a popular method based on GP regression to estimate $f$.
It involves numerical matrix inversion, which requires a time complexity that scales cubically with the sample size,
quickly becoming prohibitive, even on modern computing platforms \citep[Chapter~8]{RasmussenWilliams06}.
Sequential sampling algorithms for solving problem~\eqref{eq:SO} mostly need to repeat similar but increasingly demanding computations on a growing set of samples.
Eventually, it may be computationally more expensive to process the samples than to run simulation to acquire them  \citep{HuangAllenNotzZeng06}.

Moreover, to demonstrate the computational challenge associated with  selecting the design points,
we note that in each iteration of a sequential sampling algorithm, the next design point is usually determined by optimizing certain metric--which is often called \emph{acquisition function} in BO literature---that measures the prospect of a candidate location.
This itself is a non-convex optimization problem in high dimensions \citep{Frazier18}.
The design points can also be generated randomly from a probability distribution that basically approximates the likelihood of global optima.
But these high-dimensional distributions are multimodal in general, and generating samples from them is also computationally difficult \citep{SunHuHong18}.

\subsection{Principles and Tools for Algorithm Design}

To address the two challenges in high dimensions while maintaining a wide scope of application,
we adopt the following principles for designing our sampling algorithm.
First, the assumptions that we impose on $f$  should be general enough so that the induced functions space $\ScrF$ includes functions of practical interest;
at the same time, they should not be too general, thus largely alleviating the curse of dimensionality on the sample complexity when estimating and optimizing an unknown function in $\ScrF$.
From this perspective,
the space of all Lipschitz continuous functions is excessively broad, for in this space, function optimization  requires a sample size that grows exponentially with the dimensionality, regardless of the algorithms \citep{MalherbeVayatis17}.

A second algorithm design principle is that the design points should be determined with a computational cost that is negligible relative to that of running the simulation model.
Although approximations are a common choice for reducing computational complexity, and
a plethora of approximation methods are indeed available---for example, for computing GP regression from large datasets \citep{LiuOngShenCai20}---we seek to achieve fast computations with no approximations involved.
This is because the optimality gap of the returned solution  in the presence of approximate computations---whether they are used in processing the simulation samples to construct an estimate of the response surface, or in  optimizing an acquisition function---is difficult to quantify,
thereby demonstrating that the algorithm falls short of the theoretical guarantees of its statistical properties.

To implement the preceding design principles, we primarily employ two mathematical tools: Brownian fields and truncated sparse grids,
which are introduced in Sections~\ref{sec:Brownian} and \ref{sec:sparse-grids}, respectively.
We integrate both tools with
the KRR method (Appendix~\ref{sec:RKHS}) and
the EI criterion (Appendix~\ref{sec:EI}) to devise a sampling algorithm that achieves both low sample complexity and low computational complexity in high dimensions without resorting to approximation schemes.

\section{Brownian Fields}\label{sec:Brownian}

In this section, we first introduce Brownian fields, and the class of kernels that are associated with them.
We then discuss the function space that is induced by a Brownian field kernel,
characterizing the differentiability of the functions in the space and providing concrete examples that arise from management science and operations research.

\subsection{Definition}

Let us begin with the one-dimensional case.
Suppose that $\{\SFB(x):x\geq 0\}$ is a one-dimensional standard Brownian motion and $Z$ is an independent standard normal variable.
Then, $\SFG(x) \coloneqq Z+ \SFB(x) $ defines a Brownian motion that is initialized with the standard normal distribution:  $\SFG(0)\sim \mathtt{Normal}(0,1)$.
This process is a zero-mean Gaussian process with kernel (i.e., covariance function) $\Cov[\SFG(x), \SFG(x')] = 1 + x\wedge x'$ for all $x,x'\geq 0$,
where $x\wedge x' = \min(x,x')$.
More generally, we may consider a kernel of the form $k(x,x') = \theta + \gamma (x\wedge x') $ for some positive constants $\theta$ and $\gamma$.
Then, the zero-mean Gaussian process that corresponds to this kernel has the same distribution as $\{\theta^{\frac{1}{2}} Z + \gamma^{\frac{1}{2}} \SFB(x):x\geq 0\}$.

In the multi-dimensional case, a $d$-dimensional Brownian field (BF) is a zero-mean Gaussian process on $\Real^d_+\coloneqq \{\BFx=(x_1,\ldots,x_d)\in\Real^d: x_j\geq 0\mbox{ for all } j=1,\ldots,d\}$ having kernel
\begin{equation}\label{eq:BM-kernel}
    k_{\mathsf{BF}}(\BFx,\BFx')=\prod_{j=1}^d\bigl[\theta_j+\gamma_j(x_j\wedge x_j')\bigr]
\end{equation}
for all $\BFx,\BFx'\in\Real_+^d$,
where $\theta_j$ and $\gamma_j$ are positive constants for all $j=1,\ldots,d$.
We call $k_{\mathsf{BF}}(\BFx,\BFx')$ a BF kernel.
Because $k_{\mathsf{BF}}$ is defined on $\Real_+^d$, which includes $\ScrX=(0,1)^d$ as a subset,
in the sequel we shall consider the restriction of $k_{\mathsf{BF}}$ on $\ScrX$ unless otherwise specified.

Note that BF kernels are in the form of a tensor product.
This form, in conjunction with the Markov property of BFs \citep{SalemiStaumNelson19},
turns out to be critical for addressing both the statistical and computational challenges reviewed in Section~\ref{sec:challenges}.
In Section~\ref{sec:mixed-sobolev} we introduce
the function spaces that will facilitate the statistical analysis of our algorithm.
See the e-companion for a discussion of fast computations.

\subsection{Function Spaces}\label{sec:mixed-sobolev}

Let $k$ be a BF kernel.
In the present paper, we suppose that the objective function $f$ lies in the RKHS induced by $k$.
This means that $f$ can be expressed as a (possibly infinite) linear combination of kernel functions: $f=\sum_{i=1}^\infty \beta_i k(\BFx_i,\cdot)$ for some sequences
$\{\beta_i\}_{i=1}^\infty \subset \Real$ and $\{\BFx_i\}_{i=1}^\infty\subseteq \ScrX$.
A particularly attractive feature of RKHSs is that they allow us to construct an estimate of $f$ in a nonparametric yet analytically tractable fashion via KRR. We provide an overview of RKHS theory in Appendix~\ref{sec:RKHS}.

However, in practice it may be difficult to verify a priori that the response surface of a simulation model takes the linear combination form, thereby certifying its membership of a RKHS.
To this end,
we show that the RKHS induced by $k$ is equivalent to a function space that is defined via a mild condition on function smoothness (i.e., level of differentiability).
Practitioners can safely and easily assume this smoothness condition.
In addition, the use of the latter function space facilitates the theoretical analysis of our algorithm.

\begin{definition}[Sobolev Spaces with Dominating Mixed Smoothness]
Let
$L^2(\ScrX)$ be the space of square-integrable functions on $\ScrX$.
For each $m\in\NatInt$,
the order-$m$ Sobolev space with dominating mixed smoothness is defined as
\begin{equation}\label{eq:mixed-sobolev-def}
\ScrH^m_{\mathsf{mix}}\coloneqq \Biggl\{
g\in L^2(\ScrX) : \|g\|_{\ScrH^m_{\mathsf{mix}}}\coloneqq \biggl\|\frac{\partial^{md}g}{\partial x_1^{m}\cdots \partial x_d^{m}} \biggr\|_2 < \infty
\Biggr\},
\end{equation}
where $\frac{\partial^{md}g}{\partial x_1^{m}\cdots \partial x_d^{m}}$ denotes the weak partial derivative and $\|\cdot\|_2$ denotes the $L^2$ norm.
\end{definition}

\begin{proposition}
\label{prop:RKHS-norm-equivalence}
Let $k:\ScrX\times \ScrX\mapsto \Real$ be a BF kernel and ${\ScrH_k}$ be the RKHS induced by $k$. Then, $\ScrH_k$ is norm-equivalent to $\ScrH_{\mathsf{mix}}^1$. That is,
$\ScrH_k = \ScrH_{\mathsf{mix}}^1$
as a set of functions; moreover, there exist  some positive constants $C_1$ and $C_2$ such that
$C_1 \| g\|_{\ScrH_{\mathsf{mix}}^1} \leq \|g\|_{\ScrH_k} \leq C_2 \| g\|_{\ScrH_{\mathsf{mix}}^1}$ for all $g\in\ScrH_k$.
\end{proposition}

We shall develop an algorithm to optimize functions in $\ScrH_{\mathsf{mix}}^1$---which is equivalent to $\ScrH_k$---and analyze its rate of convergence.
Because a higher value of $m$ represents a higher-order smoothness,
$\ScrH_{\mathsf{mix}}^2$ is a subset of   $\ScrH_{\mathsf{mix}}^1$.
If such information about function smoothness is known,
the algorithm may achieve a faster rate of convergence,
as the function space of interest is smaller.
We also investigate the convergence rate of the algorithm if the objective function is in $\ScrH_{\mathsf{mix}}^2$.

\begin{remark}
Let $\BFalpha=(\alpha_1,\ldots,\alpha_d)$ be a multi-index,
$|\BFalpha| = \sum_{j=1}^d \alpha_j$,
and $D^\BFalpha g = \frac{\partial^{|\BFalpha|}g}{\partial x_1^{\alpha_1}\cdots \partial x_d^{\alpha_d}}$ denote the $\BFalpha$-th weak partial derivative of $g$.
Let $\BFone = (1,1,\ldots,1)$ be the vector of all 1s.
By the definition in \eqref{eq:mixed-sobolev-def}, $\ScrH_{\mathsf{mix}}^1$ consists of functions $g$ such that the $\BFone$-th weak partial derivative $D^{\BFone}g = \frac{\partial^d g}{\partial x_1\cdots \partial x_d}$ exists.
We stress that this condition is much weaker than requiring $g$ to be weakly differentiable up to order $d$.
The latter condition means that $D^{\BFalpha}g$ should exist for all $\BFalpha$ such that $|\BFalpha|=d$,
which requires, for example, the existence of $\frac{\partial^d g}{\partial x_j^d}$ for all $j$ (i.e., $g$ is $d$ times weakly differentiable in each coordinate).
\end{remark}

We note that if $d=1$, $\ScrH_{\mathsf{mix}}^1$ includes not only differentiable functions, such as polynomials, but also functions that are differentiable except on sets of zero measure such as $|x-\frac{1}{2}|$.
To better illustrate this, we will provide several multi-dimensional examples of functions in $\ScrH_{\mathsf{mix}}^1$.

\begin{example}
\label{exmp:ProdAssort}
Consider a product assortment problem from \cite{Aydin08}.
The problem involves $d$ products with joint inventory and pricing decisions in a newsvendor model.
For each product $j$, let
$x_j$ be its price and
$c_j$ be its unit procurement cost.
Suppose that, given a prive vector $\BFx=(x_1,\ldots,x_d)$, the demand of product $j$ is $D_j(\BFx)= \varepsilon_j Q_j(\BFx)$, where $\varepsilon_j$'s are i.i.d. uniform random variables on $(a,b)$, and
$Q_j(\BFx)=e^{\alpha_j - x_j} / [1+ \sum_{i=1}^d e^{\alpha_i-x_i}]$ for some positive parameters $\alpha_1,\ldots,\alpha_d$.
Then, the expected profit under the optimal inventory decision is
\[
f(\BFx) = \frac{1}{2(b-a)}\sum_{j=1}^d \left[ (b-a)\left(\frac{x_j-c_j}{x_j}\right) + a \right]^2 Q_j^2(\BFx),
\]
for all $x_j\in (l_j,u_j)$, $j=1,\ldots,d$, where the interval is a given price range of interest.
% A two-dimensional projection of $f$ is shown in the left panel of Figure~\ref{fig:jackson}.
Since each term $j$ in the above summation is infinitely differentiable for $x_j\in(l_j,u_j)$,
the weak partial derivative $D^{\BFone}f$  exists with a finite $L^2$ norm.
Thus, $f\in \ScrH_{\mathsf{mix}}^1$ up to a change of variables.
\end{example}

\begin{figure}[ht]
\FIGURE{
\includegraphics[width=0.4\textwidth]{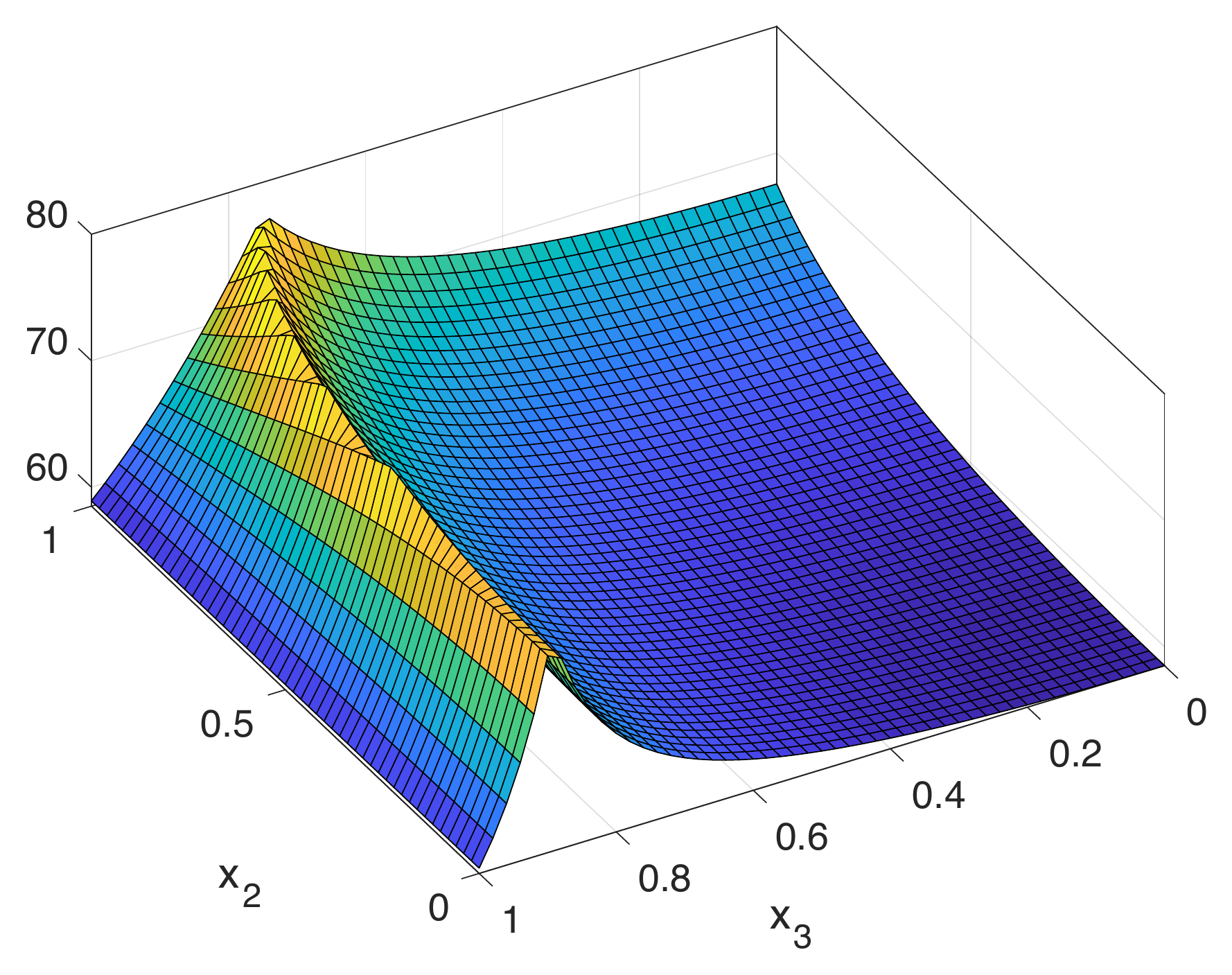}
}
{Two-Dimensional Projection of the Expected Cycle Time in a Jackson Network. \label{fig:jackson}}
{}
\end{figure}

\begin{example}\label{exmp:Jackson}
Consider an open Jackson network with $n$ stations and $d$ classes of jobs.
For all $i=1,\ldots,n$ and $j=1,\ldots,d$,
let $\mu_i$ be the service rate of station $i$,
$\alpha_j$ be the fraction of jobs of type $j$,
and $\delta_{i,j}$ be the expected number of visits to station $i$ by jobs of type $j$.
Moreover, let $\rho$ be the utilization of the bottleneck station, which is the one having the largest utilization among all stations.
The response surface of interest is the steady-state mean cycle time (CT)  (i.e., the time that an individual job takes to traverse a given routing in the system) of jobs of---for example, type 1---as a function of $(\alpha_1,\ldots,\alpha_d,\rho)$.
\cite{YangLiuNelsonAnkenmanTongarlak11} demonstrate that
\[
\E[\text{CT}_1]
= \sum_{i=1}^n \delta_{1i} \left[\mu_i - \rho \left(\frac{\sum_{j=1}^d \alpha_j \delta_{ij} }{\max_{1\leq \ell\leq n} \sum_{j=1}^d \alpha_j \delta_{\ell j} \mu_\ell^{-1}}\right)\right]^{-1}.
\]
Note that by definition, $\alpha_j$'s must satisfy the constraint $\sum_{j=1}^d \alpha_j=1$.
To construct the design space in the form of a Cartesian product,
we let $x_1 = \sqrt{\alpha_1}$, $x_j=\sqrt{\alpha_j/(1-\sum_{h=1}^{j-1}\alpha_h)}$ for $j=2,\ldots,d-1$, and $x_d=\rho$.
Let $f(\BFx)$ be the function after replacing $(\alpha_1,\ldots,\alpha_d,\rho)$ with $\BFx=(x_1,\ldots,x_d)$ in $\E[\text{CT}_1]$.
As $(x_1,\ldots,x_{d-1})$ varies, the bottleneck may vary between stations, leading to non-differentiability in the response surface.
A two-dimensional projection of $f$ is shown in Figure~\ref{fig:jackson}.
It can be verified via direct calculation that $D^{\BFone}f$ exists with a finite $L^2$ norm on $(l,u)^d$ for some $0<l<u<1$ that represent the design space of interest. Hence, $f\in \ScrH_{\mathsf{mix}}^1$ up to a change of variable.
\end{example}

\begin{example}\label{exmp:SumProdFunc}
Let $n\in\NatInt$, $\{c_i\}_{i=1}^n\subset \Real$,
and $f_{i,j}:(0,1)\mapsto\Real$ be a weakly differentiable function for all $i=1,\ldots,n$ and $j=1,\ldots,d$.
Then,
\[f(\BFx)=\sum_{i=1}^n c_i \prod_{j=1}^d f_{i,j}(x_j)\]
is a function in $\ScrH_{\mathsf{mix}}^1$.
Many test functions for global optimization are of this form up to a possible change of variables, such as the Griewank, Schwefel-2.22, and Rosenbrock functions:
\begin{align*}
& f_{\text{Griewank}}(\BFx)=50\bigg[\sum_{j=1}^d\frac{x_j^2}{4000}-\prod_{j=1}^d\cos(\frac{x_j}{j})+1\bigg], \\
& f_{\text{Schwefel-2.22}}(\BFx)=\sum_{j=1}^d|x_j|+\prod_{j=1}^d|x_j|+100, \\
&f_{\text{Rosenbrock}}(\BFx) =  \sum_{j=1}^{d-1}\big[100(x_{j+1}-x_j^2)^2+(x_j-1)^2\big].
\end{align*}
\end{example}

\section{Sparse Grids }\label{sec:sparse-grids}

One of the simplest experimental designs is a \emph{full grid} design, also known as a lattice design.
It takes in the form of a Cartesian product: $\CalX=\CalX_1\times \cdots\times  \CalX_d$,
where each $\CalX_j$ is a set of $n_j$ one-dimensional points
in the $j$-th dimension.
Namely, $\CalX$ is composed of all the points $\BFx=(x_1,\ldots,x_d)$ such that $x_j\in\CalX_j$ for all $j=1,\ldots,d$.
Thus, the size of a full grid design is $|\CalX|=\prod_{j=1}^d n_j$.

Another advantage of a full grid design, in addition to its simplicity, is that it may facilitate the computation of the inverse of a kernel matrix when the kernel is in a tensor product form.
Specifically, if $k(\BFx,\BFx') = \prod_{j=1}^d k_j(x_j,x'_j)$, where each $k_j$ is a kernel function defined in a one-dimensional space,
then the kernel matrix $\BFK=(k(\BFx, \BFx'))_{\BFx,\BFx'\in\CalX}$ takes the form of a tensor product of matrices. That is,
$\BFK = \bigotimes_{j=1}^d \BFK_j$, where $\BFK_j$ denotes the matrix composed of entries $k_j(x,  x')$ for all $x, x'\in \CalX_j$.
It follows immediately that $\BFK^{-1} = \bigotimes_{j=1}^d \BFK_j^{-1}$,
meaning that the computation of $\BFK^{-1}$ is reduced to the inversion of a sequence of smaller matrices instead of a large matrix of size $\bigl(\prod_{j=1}^d n_j\bigr)\times \bigl(\prod_{j=1}^d n_j\bigr)$.

However, full grid designs scale poorly for high-dimensional problems, as the number of design points in a full grid grows exponentially with the dimensionality.
For example, a 10-dimensional full grid with $n_j=7$ for all $j$ is of size $7^{10}\approx 2.82\times 10^8$ (see Table~\ref{tab:Lattice-SG}).

\subsection{Classical Sparse Grids}

Sparse grids (SGs) are a class of experimental designs that---while retaining the computational convenience of full grids---are significantly smaller in high dimensions.
An SG design has a hierarchical structure, and it is specified through the notion of \emph{level}, which we denote as $\tau\geq 1$.
To construct an SG of level $\tau$, we begin with a \emph{nested} sequence of one-dimensional designs $\emptyset = \CalX_{j,0} \subseteq \CalX_{j,1} \subseteq \CalX_{j,2}\subseteq \cdots \subseteq \CalX_{j,\tau}$ for each dimension $j=1,\ldots,d$.
For example, if the design space is $\ScrX=(0,1)^d$,
we may specify $\CalX_{j,l}$'s by recursively partitioning the interval $(0,1)$ in the dyadic fashion as follows: $\CalX_{j, 1}=\{\sfrac{1}{2}\}$, $\CalX_{j, 2} = \{\sfrac{1}{4},\; \sfrac{1}{2},\; \sfrac{3}{4}\}$, $\CalX_{j, 3} = \{\sfrac{1}{8},\; \sfrac{1}{4},\; \sfrac{3}{8},\; \sfrac{1}{2},\; \sfrac{1}{5},\;  \sfrac{3}{4},\; \sfrac{7}{8} \}$, etc.
That is,
\begin{equation}\label{eq:dyadic}
\CalX_{j, l} = \{1\cdot 2^{-l},\; 2\cdot  2^{-l},\ldots, \; (2^l-1)\cdot 2^{-l} \},\quad \forall j=1,\ldots,d,\; l=1,\ldots,\tau.
\end{equation}

\begin{figure}[ht]
\FIGURE
{
$\begin{array}{c}
\includegraphics[width=0.3\textwidth]{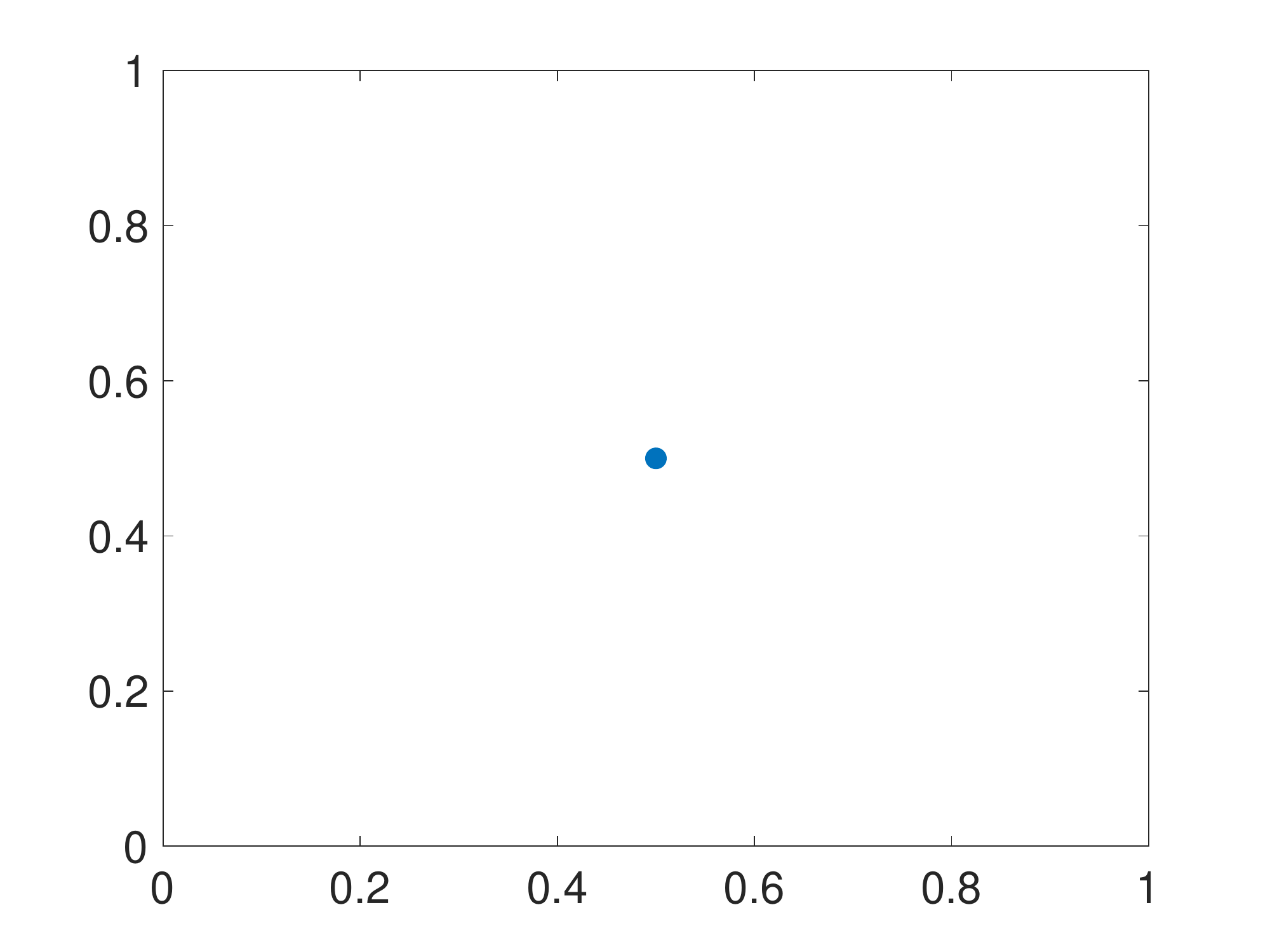}
\includegraphics[width=0.3\textwidth]{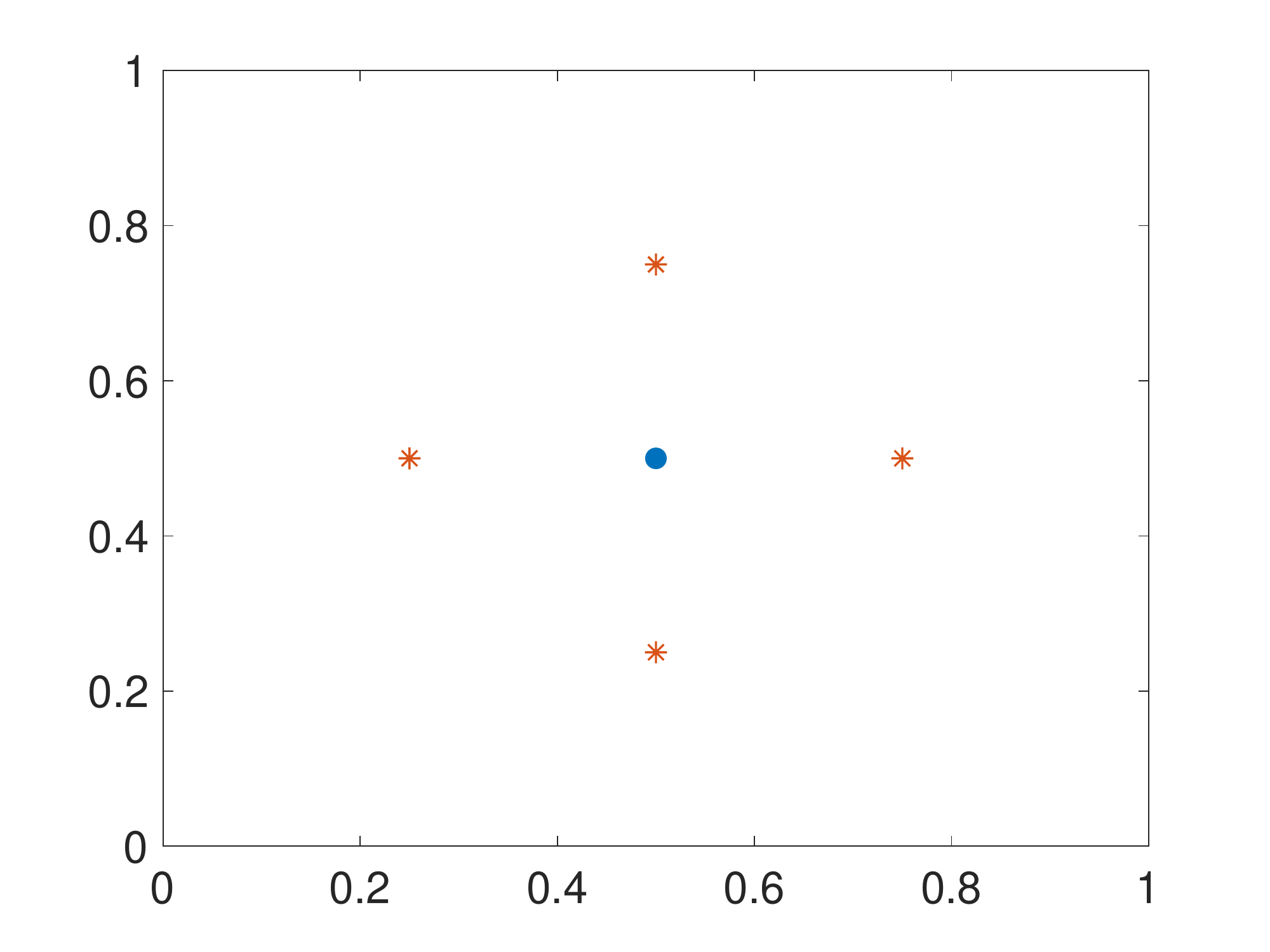}\\
\includegraphics[width=0.3\textwidth]{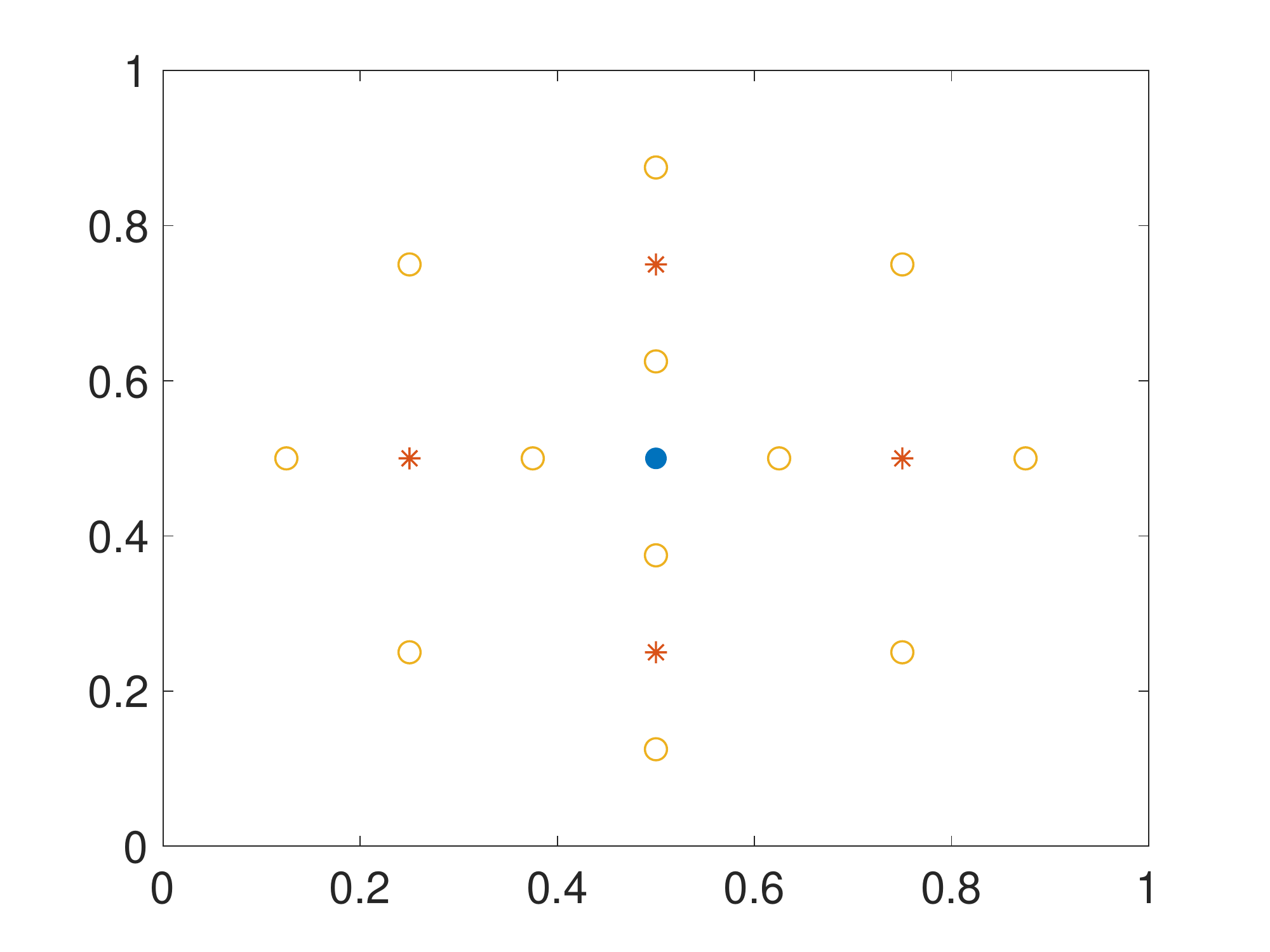}
\includegraphics[width=0.3\textwidth]{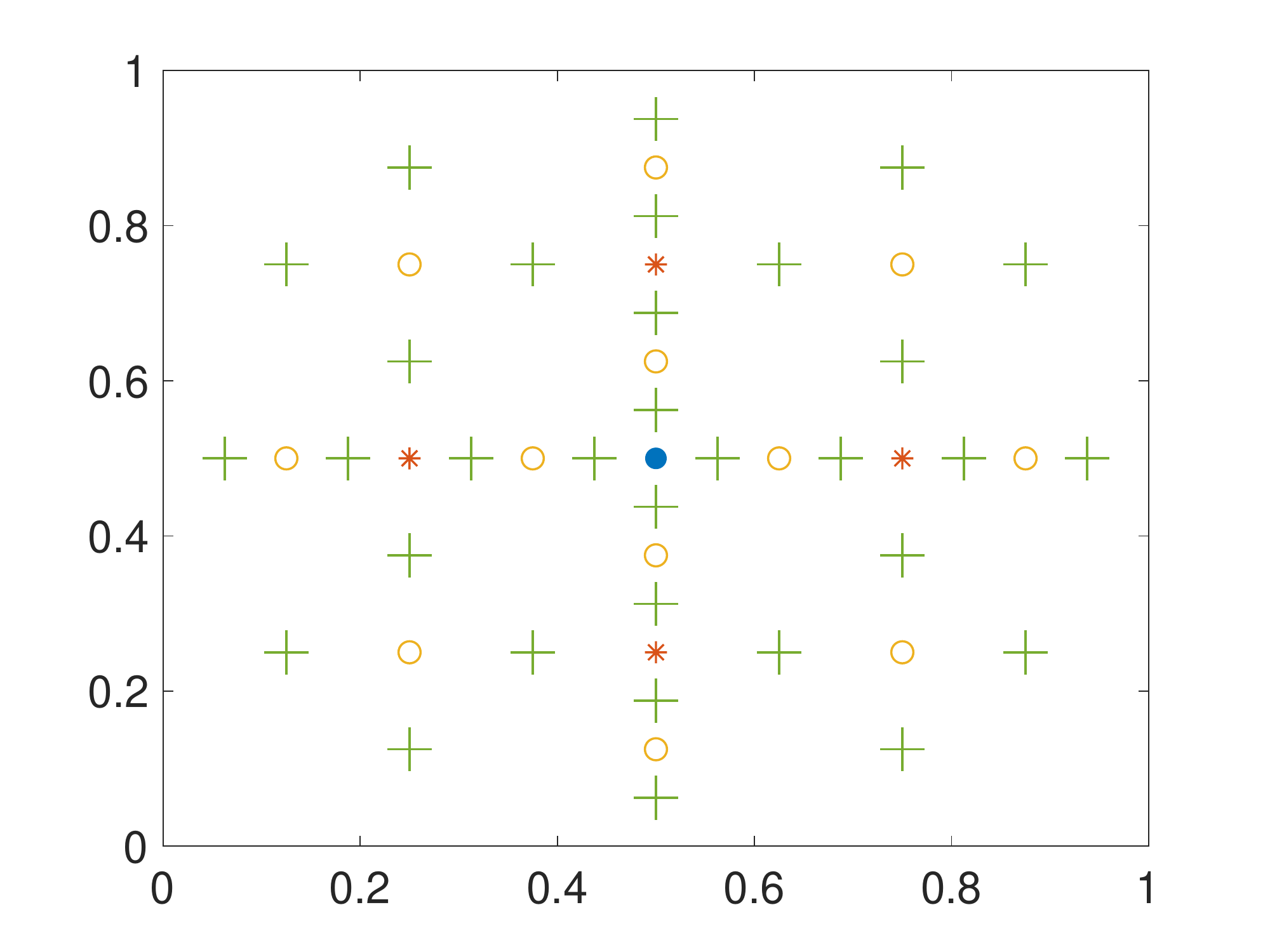}
\end{array}
$
}
{Classical Sparse Grids  of Levels 1 to 4 in Two Dimensions.  \label{fig:SG}}
{The new points added as the level increases are denoted by a different symbol and color.}
\end{figure}

Then, we may build an SG of level $\tau$ via
\begin{equation}\label{eq:SG}
    \CalX^{\mathsf{SG}}_\tau = \bigcup_{\abs{\BFl}\leq \tau+d-1} \CalX_{1,l_1} \times \cdots \times \CalX_{d,l_d},
\end{equation}
where $\BFl=(l_1,\ldots,l_d)\in \NatInt^d$ and $\abs{\BFl}=\sum_{j=1}^d l_j$.
An SG defined via the nested sequence in \eqref{eq:dyadic} is hereafter referred to as a \emph{classical SG} (see Figure~\ref{fig:SG}).

By definition, an SG  forms a union of many smaller full grids.
Taking advantage of this representation, \cite{Plumlee14} develops fast algorithms for computing $\BFK^{-1}$ for tensor product kernels and SG designs (see the e-companion).

Lemma~3.6 in \cite{bungartz_griebel_2004} asserts that the size of a classical SG of level $\tau$ is
\begin{equation}\label{eq:SG-size}
|\CalX^{\mathsf{SG}}_\tau| = \sum_{l=0}^{\tau-1} 2^l \cdot \binom{d-1+l}{d-1} \asymp 2^\tau \tau^{d-1},
\end{equation}
where $a_n\asymp b_n$ denotes the relationship such that $\limsup_{n\to\infty}a_n/b_n<\infty$ and $\limsup_{n\to\infty}b_n/a_n<\infty$.
By contrast, the full grid of the same refinement level is $\bigtimes_{j=1}^d \CalX_{j,\tau}$, and its size is $\prod_{j=1}^d|\CalX_{j,\tau}| = (2^\tau-1)^d\asymp 2^{\tau d}$.
Table~\ref{tab:Lattice-SG} compares the size of classical SGs  and full grids in different dimensions.

\begin{table}[ht]
\TABLE
{Full Grids Versus Classical Sparse Grids. \label{tab:Lattice-SG}}
{
    \begin{tabular}{ c c r c  r }
    \toprule
    Dimension $d$ && Full Grid && Sparse Grid of Level 3 \\
    \midrule
    1 && 7 && 7\\
    2 && 225 && 17\\
    5 && 16,807 && 71\\
    10 && $2.82\times10^{8}$ && 241\\
    20 && $7.98\times10^{16}$ && 881\\
    50 && $1.80\times10^{42}$ && 5,201\\
    100 && $3.23\times 10^{84}$  && 20,401\\
    \bottomrule
    \end{tabular}
}
{The refinement level of each full grid is $\tau=3$, so its size is $(2^\tau-1)^d = 7^d$.
The size of a classical SG is calculated by \eqref{eq:SG-size}.}
\end{table}

\subsection{Truncated Sparse Grids}

However, classical SGs may be inflexible to use,
because algorithms developed for them usually require that the design should be \emph{complete}  with respect to the level parameter.
The algorithms may break down if only a subset of the design points of an SG are used.
Because SGs  are specified via the level parameter instead of the size,
if the simulation budget $N$ lies between $\abs{\CalX_{\tau}^{\mathsf{SG}}}$ and $\abs{\CalX_{\tau+1}^{\mathsf{SG}}}$ for some $\tau$,
then
we may be forced to take repeated samples on the lower-level SG
because taking samples on the design points in $\CalX_{\tau+1}^{\mathsf{SG}}\setminus \CalX_{\tau}^{\mathsf{SG}}$ may render fast computation of $\BFK^{-1}$ infeasible.
Restricting the tensor product kernels to a smaller class that includes BF kernels as a special case,
\cite{DingZhang21} develop fast kernel matrix inversion algorithms that allow the design points to constitute a specific incomplete form of an SG, which is called a \emph{truncated SG} (TSG) (see the e-companion).
TSGs can be defined for any arbitrary sample size, thereby
substantially increasing their flexibility.

To define a TSG, we first note that a classical SG can be  represented as a union of disjoint sets of design points.
Let $\SFc_{l, i}\coloneqq i\cdot 2^{-l}$ for $l\geq 1$ and $i=1,\ldots,2^l -1$,
and let
$\BFc_{\BFl,\BFi}\coloneqq (\SFc_{l_1, i_1},\ldots, \SFc_{l_d, i_d})$.
The design points in \eqref{eq:dyadic} are then written as $\CalX_{j, l} = \{\SFc_{l, i}: i=1,\ldots,2^l-1 \}$ for all $j=1,\ldots,d$ and $l=1,\ldots,\tau$.
For any multi-index $\BFl=(l_1,\ldots,l_d)\in \NatInt^d$, we define a set for the  multi-index $\BFi=(i_1,\ldots,i_d)$ as follows:
\begin{equation}\label{eq:index-set-rho}
\rho(\BFl) \coloneqq \bigtimes_{j=1}^d \{i_j: i_j\mbox{ is an odd number between 1 and }2^{l_j} \}  = \bigtimes_{j=1}^d  \{1, 3, 5, \ldots, 2^{l_j}-1 \}.
\end{equation}
Then, the hierarchical structure of the classical SG of level $\tau$ indicates that it can be represented as
\begin{equation}\label{eq:disjoint-rep}
    \CalX^{\mathsf{SG}}_\tau = \bigcup_{\abs{\BFl}\leq \tau+d-1} \{\BFc_{\BFl,\BFi}: \BFi\in\rho(\BFl)\},
\end{equation}
where  $\abs{\BFl}=\sum_{j=1}^d l_j$.
Moreover, the set of design points that augments a classical SG from level $\tau-1$ to level $\tau$ is
\[\CalX_{\tau+1}^{\mathsf{SG}}\setminus \CalX_{\tau}^{\mathsf{SG}} = \{\BFc_{\BFl,\BFi}: \abs{\BFl}=\tau+d-1, \BFi\in\rho(\BFl)\}.\]
Hence, \eqref{eq:disjoint-rep} expresses $\CalX^{\mathsf{SG}}_\tau$ as a union of disjoint sets.

\begin{definition}[Truncated Sparse Grid]\label{def:TSG}
Given an integer $n\in\NatInt$, there exists $\tau\geq 1$ such that $\abs{\CalX_{\tau}^{\mathsf{SG}}} \leq n <  \abs{\CalX_{\tau+1}^{\mathsf{SG}}}$.
Let $\Tilde{n} = n-\abs{\CalX_{\tau}^{\mathsf{SG}}}$ and
$\mathcal{A}_{\Tilde{n}}$ be a size-$\Tilde{n}$ subset of $\CalX_{\tau+1}^{\mathsf{SG}} \setminus \CalX_{\tau}^{\mathsf{SG}}$. Then,
$\CalX_n^{\mathsf{TSG}} \coloneqq \CalX_{\tau}^{\mathsf{SG}} \cup \mathcal{A}_{\Tilde{n}}$ is said to be a TSG of size $n$.
\end{definition}

A particular feature of the algorithm that we propose in the present paper to solve continuous SO problems is that
the design points are selected from a classical SG instead of the entire design space, and
they will form a TSG after the simulation budget $N$ is exhausted.
Specifically,
we first take one sample from each design point of a classical SG of level $\tau$, where $\tau$ is such that $\abs{\CalX_{\tau}^{\mathsf{SG}}} \leq N < \abs{\CalX_{\tau+1}^{\mathsf{SG}}}$.
Then, we follow the EI strategy to sequentially select design points from $\CalX_{\tau+1}^{\mathsf{SG}}$ until the remaining $N-\abs{\CalX_{\tau}^{\mathsf{SG}}}$  simulation budget is exhausted.
Clearly, although samples  may be repeatedly  taken from the lower-level classical SG,
the design points that are selected  will eventually form a TSG.

\section{Algorithm}\label{sec:algo}

In this section, we first describe the general structure of our algorithm and then provide details.

\subsection{Structure}

Our algorithm has a simple structure, consisting of two stages. In Stage~1
the design points (i.e., sampling locations) are determined all at once, whereas in Stage~2 they are determined in a sequential fashion.
\begin{enumerate}[label=\emph{Stage \arabic*.}, left=0pt]
\item
(i) Identify the largest SG---for example, of level $\tau$---that does not exceed the sample size limited by the simulation budget, (ii) take samples from each design point on this SG, and (iii) compute an estimate of $f$ using KRR in conjunction with a BF kernel.
\item (i) Assign to $f$ a GP prior with the BF kernel and the KRR estimate as the mean function, and
(ii) iteratively determine the subsequent design point by optimizing the EI criterion over---instead of the entire (continuous) design space---a (discrete) set of candidate points formed by the SG of  level $\tau+1$.
\end{enumerate}

This two-stage structure greatly resembles that of a typical BO algorithm \citep{Frazier18}.
Nevertheless, remarks on several subtle yet critical differences are warranted here.
First, if a BO algorithm formally has a stage prior to sequential sampling,
this stage is generally treated as a ``warm-up'' phase, and its purpose is to obtain a  basic exploration of the design space.
In this stage, a relatively small number of design points are determined in an ad hoc manner because doing to  does not affect the algorithm's performance---at least not asymptotically for a large sample size.
By contrast, being integral to the task of combating the curse of dimensionality, Stage~1 of our algorithm has  a well-planned experimental design.
In particular, the sample size in Stage~1 constitutes  much  of the total simulation budget, depending on the size of the SG relative to the budget.

Second, in a BO algorithm, all samples---irrespective of when they are collected---are pooled for Bayesian updating of the posterior distribution of $f$.
Our algorithm, however, differentiates samples from the two stages.
While Stage~2 samples are used for Bayesian-like updating,
Stage~1 samples are instead processed using KRR.
Despite the close relationship between GP regression---on which BO algorithms are usually based---and KRR \citep[Chapter~6]{RasmussenWilliams06},
the latter permits flexibility in selecting a regularization parameter with care.
This is critical for accelerating our algorithm's rate of convergence.

Third, in a BO algorithm, the acquisition function, which is generally non-convex, is optimized over the entire design space to produce subsequent design points.
Although theoretical analysis often assumes that a globally optimal solution can be computed with relative ease for this intermediate optimization problem, it is
common practice to run a continuous optimization method multiple times, say, of a quasi-Newton type, using a random starting point each time.
As the dimension of the design space grows,
this practice grows more computationally demanding, and the quality of the solution it returns becomes more difficult to control, possibly because of the increase in local optima.
Our algorithm takes a vastly different approach: it optimizes the acquisition function over a discrete set of points that is formed by an SG and that grows mildly with the dimensionality.
Thus, our treatment is computationally fast, and the solution to the intermediate optimization problem is precise.

Lastly, BO algorithms mostly use Gaussian or Mat\'ern kernels, whereas our algorithm uses a BF kernel to drive both computations and theoretical analysis.
The significance of BF kernels is twofold.
On the one hand, their associated RKHS has a tensor product structure that is crucial for reducing the sample complexity in high dimensions.
On the other, when used jointly with SGs, they induce sparsity in the inverse kernel matrices that facilitates fast computation via sparse linear algebra.

\subsection{Details}

Let $N$ be the simulation budget and $k$ be the BF kernel defined in \eqref{eq:BM-kernel}.
To facilitate the presentation, we fix the following notations.
For any $n\geq 1$ and $\BFx\in\ScrX$, we let $\BFk_n(\BFx)=(k(\BFx_1,\BFx),\ldots,k(\BFx_n, \BFx))^\intercal $,
and
$\BFy_n(\BFx)=(y(\BFx_1),\ldots,y(\BFx_n))^\intercal $;
moreover,
we use $\BFK_n \in \Real^{n\times n}$ to denote the kernel matrix that is composed of $k(\BFx_i,\BFx_j)$ for all $i,j=1,\ldots,n$,
and use $\BFI_n$ to denote the $n\times n$ identity matrix.
Let $\Phi$ and $\phi$ denote the cumulative distribution function and the probability density function of the standard normal distribution, respectively.

\subsubsection{Stage~1: Batch Sampling}

We first identify a level-$\tau$ classical SG such that $\big|{\ScrX}_{\tau}^{\mathsf{SG}}\big|\leq N < \big|{\ScrX}_{\tau+1}^{\mathsf{SG}}\big|$.
Let $N_{\tau}\coloneqq \big|{\ScrX}_{\tau}^{\mathsf{SG}}\big|$ and
$\{\BFx_1,\ldots,\BFx_{N_\tau}\}$ be all the design points in ${\ScrX}_{\tau}^{\mathsf{SG}}$.
We take one sample at each $\BFx_i$, resulting in observations
$y(\BFx_i)=f(\BFx_i) + \varepsilon(\BFx_i)$ for $i=1,\ldots,N_\tau$.
Given the data $\{(\BFx_i, y(\BFx_i)\}_{i=1}^{N_\tau}$,
we construct an estimate of $f$ via KRR, which solves the regularized least-squares:
\[
\min_{g\in\ScrH_k}\frac{1}{N_\tau}\sum_{i=1}^{N_\tau} (y(\BFx_i) - g(\BFx_i))^2 + \lambda \|g\|_{\ScrH_k}^2,
\]
where $\lambda>0$ is the chosen regularization parameter, and $\ScrH_k$ denotes the RKHS induced by $k$.
The solution is
\begin{align}\label{eq:KRR-solution}
\widehat{f}_{N_\tau,\lambda}(\BFx) \coloneqq
\BFk_{N_\tau}^\intercal(\BFx)\left(\BFK_{N_\tau}+N_{\tau}\lambda\BFI_{N_{\tau}}\right)^{-1}\BFy_{N_\tau}.
\end{align}
We refer to Appendix~\ref{sec:RKHS} for an introduction to KRR.

In general, the computation of KRR---which involves matrix inversion---is high in time complexity when the sample size is large.
For example, if $k$ is chosen to be a Mat\'ern kernel, or if the design points form
a random design or a Latin hypercube design, then
one usually needs to use generic matrix inversion algorithms whose computational complexity is cubic in the sample size.
Nevertheless, fast algorithms for computing the KRR estimate are available if $k$ is a BF kernel and the design points form a TSG (see the e-companion).

\subsubsection{Stage~2: Sequential Sampling}

Upon completion of Stage~1, the remaining simulation budget is $N-N_\tau$
For each $n=N_\tau,\ldots, N-1$,
we select the next design point by maximizing the EI acquisition function:
\begin{align}\label{eq:modifed-EI}
\BFx_{n+1} = \argmax_{\BFx\in\ScrX_{\tau+1}^{\mathsf{SG}}}\biggl\{ s_n(\BFx)\eta\biggl(\frac{\widetilde{f}_n(\BFx)-\max_{1\leq i\leq n}\widetilde{f}_n(\BFx_i)}{ s_n(\BFx)}\biggr)\biggr\},
\end{align}
where $\eta(z)\coloneqq z\Phi(z)+\phi(z)$. Also,
\begin{align}
  \widetilde{f}_n(\BFx)={}&\widehat{f}_{N_\tau,\lambda}(\BFx)+\delta_n^2\BFk_n^\intercal(\BFx)\left(\delta_n^2\BFK_n+\sigma^2 \BFI_n \right)^{-1} (\BFy_n-\widehat{\BFf}_{N_\tau,\lambda,n}), \label{eq:mean-proxy}\\
 s^2_n(\BFx)={}&\delta_n^2 k(\BFx,\BFx)-\delta_n^2\BFk_n^\intercal(\BFx)\left(\delta_n^2\BFK_n+\sigma^2 \BFI_n \right)^{-1}\delta_n^2\BFk_n(\BFx),\label{eq:var-proxy}
\end{align}
where $\widehat{\BFf}_{N_\tau,\lambda,n} = (\widehat{f}_{N_\tau,\lambda}(\BFx_1),\ldots, \widehat{f}_{N_\tau,\lambda}(\BFx_n))^\intercal$, and $\delta_n>0$ is a tuning parameter that  controls the ratio between exploration and exploitation.

Notably, there are several critical modifications relative to the standard EI algorithm (see Appendix~\ref{sec:EI}).
First, when maximizing the acquisition function in \eqref{eq:modifed-EI},
we restrict the feasible set from $\ScrX$ to $\ScrX_{\tau+1}^{\mathsf{SG}}$.
The computational benefit is evident: the former set is a high-dimensional continuous set, whereas the latter is a discrete set of a moderate size: $|\ScrX_{\tau+1}^{\mathsf{SG}}|=\CalO(2^{\tau+1}(\tau+1)^{d-1})=\CalO(N(\log(N))^{d-1})$, which can be shown using \eqref{eq:SG-size} and the fact that $|\ScrX_{\tau}^{\mathsf{SG}}|\leq N < |\ScrX_{\tau+1}^{\mathsf{SG}}|$.
This is one of the two key elements that address the computational challenge in high dimensions discussed in Section~\ref{sec:challenges}. (The other key element is fast matrix inversion,  which is enabled by the joint use of BF kernels and TSGs.
The matrices that need to be inverted in  \eqref{eq:mean-proxy}--\eqref{eq:var-proxy} are  structurally similar to the matrix in \eqref{eq:KRR-solution},
so they can also be computed using the fast matrix inversion algorithms in the e-companion.)

In addition, the standard EI algorithm works under the premise that
(i) $f$ is assigned a GP prior with a \emph{fixed} kernel and
(ii) the samples have a normal distribution.
Nevertheless,
comparing \eqref{eq:mean-proxy}--\eqref{eq:var-proxy}  with  \eqref{eq:posterior-mean}--\eqref{eq:posterior-var} suggests that we are acting as if
the kernel of the GP prior is $\delta_n^2 k$ that varies with $n$.
Moreover, we assume in the present paper that the samples are sub-Gaussian
(see Assumption~\ref{assump:sub-Gaussian}).
Hence,  $\widetilde{f}_n(\BFx)$ is not technically the posterior mean of $f$, and  $s_n^2(\BFx)$  is not the posterior variance.
In other words, we adopt an instrumental view of the EI strategy.
We do not attempt to substantiate a Bayesian interpretation of
the computations in
\eqref{eq:modifed-EI}--\eqref{eq:var-proxy}---nor would such an interpretation be necessary for the purpose of optimizing $f$---but instead, we treat them as merely a means to find a suitable subsequent design point.
Our algorithm is a frequentist method rather than a Bayesian one.

We summarize the discussion thus far in Algorithm~\ref{alg:EI} and name it \textbf{K}ernel \textbf{E}xpected \textbf{I}mprovement via \textbf{B}rownian Fields and \textbf{S}parse Grids  (\texttt{KEIBS}).

\begin{algorithm}[t]
\caption{\texttt{KEIBS}} \label{alg:EI}
\SingleSpacedXI
\SetAlgoLined
\DontPrintSemicolon
\SetKwInOut{Input}{Input}
\SetKwInOut{Output}{Output}
\SetKwInOut{StageOne}{Batch Sampling (KRR)}
\SetKwInOut{StageTwo}{Sequential Sampling (EI)}
\Input{Budget $N$, tuning parameters $\lambda$ and $\delta_n$. \;}
\StageOne{
Select level $\tau$ such that $N_{\tau}\coloneqq \big|{\ScrX}_{\tau}^{\mathsf{SG}}\big|\leq N < \big|{\ScrX}_{\tau+1}^{\mathsf{SG}}\big|$. \;
Set the design points $\{\BFx_1,\ldots,\BFx_{N_\tau}\} = {\ScrX}_{\tau}^{\mathsf{SG}}$. \;
Take a sample $y(\BFx_i)$ at each $\BFx_i$, $i=1,\ldots,N_\tau$.\;
% resulting in $\BFy_{N_\tau}=(y(\BFx_1),\ldots,y(\BFx_{N_\tau}))^\intercal$\;
Construct the KRR estimate:
\[\widehat{f}_{N_\tau,\lambda}(\BFx) = \BFk_{N_\tau}^\intercal(\BFx)\left(\BFK_{N_\tau}+N_{\tau}\lambda\BFI_{N_{\tau}}\right)^{-1}\BFy_{N_\tau}.\]
}
\StageTwo{
\For{$N_{\tau}\leq  n < N$}{
Select $\BFx_{n+1}\in\ScrX_{\tau+1}^{\mathsf{SG}}$ that maximizes the EI acquisition function:
\[\BFx_{n+1}=\argmax_{\BFx\in\ScrX_{\tau+1}^{\mathsf{SG}}}\biggl\{ s_n(\BFx)\eta\biggl(\frac{\widetilde{f}_n(\BFx)-\max_{1\leq i\leq n}\widetilde{f}_{N_\tau,\lambda,n}(\BFx_i)}{ s_n(\BFx)}\biggr)\biggr\},\]
where
\begin{align*}
  \widetilde{f}_n(\BFx)={}&\widehat{f}_{N_\tau,\lambda}(\BFx)+\delta_n^2\BFk_n^\intercal(\BFx)\left(\delta_n^2\BFK_n+\sigma^2 \BFI_n \right)^{-1} (\BFy_n-\widehat{\BFf}_{N_\tau,\lambda,n}),\\
 s^2_n(\BFx)={}&\delta_n^2 k(\BFx,\BFx)-\delta_n^2\BFk_n^\intercal(\BFx)\left(\delta_n^2\BFK_n+\sigma^2 \BFI_n \right)^{-1}\delta_n^2\BFk_n(\BFx).
\end{align*}
Take a sample $y(\BFx_{n+1})$ at $\BFx_{n+1}$.\;
}
}
\Output{$\widehat{\BFx}^*_N = \argmax\limits_{x\in\ScrX}\widetilde{f}_N(\BFx_i)$ and  $\widetilde{f}_N(\widehat{\BFx}^*_N) = \max\limits_{x\in\ScrX}\widetilde{f}_N(\BFx_i)$.
}
\end{algorithm}

\section{Asymptotic Analysis}\label{sec:analysis}

In this section, we establish upper bounds on the rate of  convergence of the optimality gap \eqref{eq:OG}.
This gap may be decomposed as follows:
\begin{align}
f(\BFx^*)-f(\widehat{\BFx}^*_N)={}& [f(\BFx^*)-\widetilde{f}_N(\BFx^*)] + [\widetilde{f}_N(\BFx^*) - \widetilde{f}_N(\widehat{\BFx}^*_N)]
+ [\widetilde{f}_N(\widehat{\BFx}^*_N)  - f(\widehat{\BFx}^*_N)] \nonumber \\
\leq{}& \|f-\widetilde{f}_N\|_\infty + 0 + \|f-\widetilde{f}_N\|_\infty  = 2 \|f-\widetilde{f}_N\|_\infty, \label{eq:main-decomp}
\end{align}
where $\|g\|_\infty = \inf\{C\geq 0:|g(\BFx)|\leq C \mbox{ for almost all }\BFx\in\ScrX \}$ denotes the $L^\infty$ norm of $g$,
and $\widetilde{f}_N(\BFx^*) - \widetilde{f}_N(\widehat{\BFx}^*_N) \leq 0$ because  $\widehat{\BFx}^*_N$ maximizes $\widetilde{f}_N$.
Thus,
our theoretical analysis of Algorithm~\ref{alg:EI} will focus on the error term $\|f-\widetilde{f}_N\|_\infty$.

Because $\widetilde{f}_N$ takes a different form depending on whether the samples are noise-free or noisy,
the analysis of $\|f-\widetilde{f}_N\|_\infty$ demands a different set of techniques.
We present the results for these two scenarios in Section~\ref{sec:noise-free} and Section~\ref{sec:noisy}, respectively.
We further discuss the scenario when the objective function is of a higher-order smoothness
in Section~\ref{sec:smoother}.
However, the analysis is intricate, so the details are deferred to the e-companion.

\subsection{Noise-free Samples}\label{sec:noise-free}

In the absence of simulation noise, Assumption~\ref{assump:sub-Gaussian} is satisfied with $\sigma=0$,
and we set the tuning parameters in Algorithm~\ref{alg:EI} as $\lambda=0$  and $\delta_n=1$.
Then,
each observation $y_i$ is identical to the function value $f(\BFx_i)$.
Moreover,
the KRR estimate $\widehat{f}_{N_\tau,\lambda}(\BFx)$ at the end of Stage~1 is reduced to the kernel interpolation (KI) estimator
\[\breve{f}_{N_\tau}(\BFx) \coloneqq \BFk_{N_\tau}^\intercal(\BFx)\BFK_{N_\tau}^{-1} f(\CalS_{N_\tau}),\]
where $\CalS_n \coloneqq \{\BFx_i\}_{i=1}^n$ denotes the set of design points selected by Algorithm~\ref{alg:EI} for each $n$,
and $f(\CalS_n) = (f(\BFx_1),\ldots, f(\BFx_n))^\intercal$ (see Appendix~\ref{sec:RKHS}).
Hence,
\begin{align*}
     \|f-\widetilde{f}_N\|_\infty ={}& \| f- [\breve{f}_{N_\tau} + \BFk_N^\intercal(\cdot) \BFK_N ^{-1}(f(\CalS_N) - \breve{f}_{N_\tau}(\CalS_N)) ] \|_\infty \nonumber \\
     ={}& \| f- \BFk_N^\intercal(\cdot) \BFK_N ^{-1}f(\CalS_N)  + \BFk_N^\intercal(\cdot) \BFK_N ^{-1} \breve{f}_{N_\tau}(\CalS_N) - \breve{f}_{N_\tau} \|_\infty \\
     ={}& \| \underbrace{f- \breve{f}_N}_{J_1}  + \underbrace{\BFk_N^\intercal(\cdot) \BFK_N ^{-1} \breve{f}_{N_\tau}(\CalS_N) - \breve{f}_{N_\tau}}_{J_2}  \|_\infty. \label{eq:decomp-J1-J2}
\end{align*}

Taking advantage of both the hierarchical structure of  classical SGs and the tensor structure of the function space $\ScrH_{\mathrm{mix}}^1$, we show that if $f\in \ScrH_{\mathrm{mix}}^1$,
then $f$ can be expressed as an orthogonal expansion of basis functions that are jointly determined by the BF kernel and the design points on classical SGs.
Next, we show that the KI estimator $\breve{f}_N$ exactly equals the sum of the first $N$ terms of this expansion,
and thus $J_1$ becomes the expansion's remainder.
The orthogonality of the basis functions allows us to calculate both the $L^2$ norm and the RKHS norm---which is equivalent to the $\ScrH_{\mathsf{mix}}^1$ norm by Proposition~\ref{prop:RKHS-norm-equivalence}---of $J_1$.
Lastly, we apply the Gagliardo–Nirenberg interpolation inequality \citep{haroske2017gagliardo} to link the $L^\infty$ norm with the other norms, yielding
\[\|J_1\|_\infty =  \CalO\left(N^{-\frac{1}{2}}(\log N)^{\frac{3(d-1)}{2}}\right). \]

Also leveraging the expansion, we show that $J_2=0$, which leads us to the following theorem.

\begin{theorem}\label{thm:EI_noiseless}
Suppose that $f\in\ScrH^1_{\mathsf{mix}}$, Assumption~\ref{assump:optimum} holds, and Assumption~\ref{assump:sub-Gaussian} holds with $\sigma=0$.
Let $\lambda=0$  and $\delta_n = 1$ in Algorithm~\ref{alg:EI}.
Then,
\[f(\BFx^*)-f(\widehat{\BFx}^*_N)=\CalO\left(N^{-\frac{1}{2}}(\log N)^{\frac{3(d-1)}{2}}\right).\]
\end{theorem}

\begin{remark}
When the simulation samples are noise-free, Algorithm~\ref{alg:EI} becomes deterministic, involving no random variables.
Hence, the upper bound in Theorem~\ref{thm:EI_noiseless} is established without taking an expectation.
\end{remark}

\begin{remark}
For $d=1$, $\ScrH^1_{\mathsf{mix}}$ is identical to the classical first-order Sobolev space $\ScrH^1$, and the upper bound in Theorem~\ref{thm:EI_noiseless} is reduced to $\CalO(N^{-\frac{1}{2}})$.
This upper bound then matches the known minimax lower bound on the optimality gap for maximizing a function in $\ScrH^1$ in one dimension without observation noise \citep{Bull11}.
\end{remark}

We conclude this subsection with a companion result,
which shows that the function space $\ScrH_{\mathsf{mix}}^1$ is broad,
in the sense that maximizing a function in that space without a good algorithm may suffer severely from the curse of dimensionality.

\begin{proposition}
\label{prop:FG-convergence}
Suppose that $f\in\ScrH^1_{\mathsf{mix}}$, Assumption~\ref{assump:optimum} holds, and Assumption~\ref{assump:sub-Gaussian} holds with $\sigma=0$.
Let the design points  $\{\BFx_1,\ldots,\BFx_N\}$  form a full grid
$\bigtimes_{j=1}^d \CalX_{j,\tau} $ for some $\tau\geq 1$ where $\CalX_{j,\tau} = \{i\cdot 2^{-\tau}:1\leq i\leq 2^\tau-1\}$.
Then, for any $0<L<\infty$, there exists a constant $c>0$ such that for all sufficiently large $N$ and any interpolation estimator $\breve{f}_N$,
\[
\sup_{\|f\|_{\ScrH_{\mathsf{mix}}^1}\leq L}\biggl\{\max_{\BFx\in\ScrX}f(\BFx)-\max_{\BFx\in\ScrX}\breve{f}_N(\BFx)\biggr\} \geq  c N^{-\frac{1}{2d}}.
\]
\end{proposition}

\subsection{Noisy Samples}\label{sec:noisy}

When the simulation samples are indeed noisy, we have
\begin{align*}
     \|f-\widetilde{f}_N\|_\infty
      ={}  & \|\underbrace{(f-\widehat{f}_{N_\tau,\lambda})}_{U_1}-\underbrace{\BFk_N^\intercal(\cdot) (\BFK_N+ \delta_N^{-2} \sigma^2  \BFI_N )^{-1}(y(\CalS_N)-\widehat{f}_{N_\tau,\lambda}(\CalS_N))}_{U_2}\|_\infty,
\end{align*}
where $y(S_N) = (y(\BFx_1),\ldots, y(\BFx_N))^\intercal$ and $\widehat{f}_{N_\tau,\lambda}(\CalS_N)) = (\widehat{f}_{N_\tau,\lambda}(\BFx_1),\ldots,\widehat{f}_{N_\tau,\lambda}(\BFx_N))^\intercal $.
A key observation is that
$y(\CalS_N)-\widehat{f}_{N_\tau,\lambda}(\CalS_N)$ is a collection of samples of $f-\widehat{f}_{N_\tau,\lambda}$, and thus
$U_2$ may be recognized as the KRR estimator of $U_1$ based on these samples.
It follows that both $U_1$ and $U_1-U_2$ are in the form of the difference between a function and its KRR estimator.
Consequently, to bound $\|f - \widehat{f}_N\|_\infty$,
it suffices to study the convergence rate of the KRR estimator $\widehat{f}_{n,\lambda}$ under the $L^\infty$ norm.
This demands a distinct set of analytical tools from those employed for the noise-free case.

In particular,
we apply empirical process theory to bound the empirical semi-norm of $f-\widehat{f}_{n, \lambda}$ in the presence of sub-Gaussian noise.
We then
apply approximation theory in Sobolev spaces to connect the empirical semi-norm and the $L^\infty$ norm, using the RKHS norm (equivalently, the $\ScrH_{\mathsf{mix}}^1$ norm) as a bridge.
This leads to an upper bound---which involves the tuning parameters $\lambda$ and $\delta_n$---on $\|f-\widehat{f}_{n, \lambda}\|_\infty$.
Choosing them carefully to improve the bound results in the following theorem.

\begin{theorem}
\label{thm:Convergence_EI}
Suppose that $f\in\ScrH^1_{\mathsf{mix}}$, Assumption~\ref{assump:optimum} holds, and Assumption~\ref{assump:sub-Gaussian} holds with $\sigma>0$.
Let $\lambda \asymp \left(\sigma^4 N_{\tau}^{-2}|\log (\sigma N_\tau)|^{{2d-1}}\right)^{\frac{1}{3}}$ and $\delta_n\asymp\left( \sigma^2 n^{-1}|\log (\sigma n)|^{1-2d}\right)^{\frac{1}{6}}$  in Algorithm~\ref{alg:EI}.
Then,
\begin{align*}
\E[f(\BFx^*)-f(\widehat{\BFx}^*_N)]
 =   \CalO\left(\sigma^{\frac{1}{3}}N^{-\frac{1}{6}}|\log (\sigma N)|^{\frac{2d-1}{12}}(\log N)^{\frac{3(d-1)}{4}}\right).
\end{align*}
\end{theorem}

Comparing Theorems~\ref{thm:EI_noiseless} and \ref{thm:Convergence_EI}, one can  clearly see that the dominating term in the upper bound in the noisy case is $N^{-\frac{1}{6}}$,
whereas that in the noise-free case is $N^{-\frac{1}{2}}$.
The former is significantly slower than the latter, demonstrating the effect of simulation noise on the convergence rate.

\subsection{Higher-order Smoothness}\label{sec:smoother}

In practice, the objective function of an SO problem may be smoother than that indicated by~$\ScrH_{\mathrm{mix}}^1$. (Recall the product assortment problem in Example~\ref{exmp:ProdAssort}.)
In this subsection, we discuss the convergence rates of Algorithm~\ref{alg:EI} when $f$ is in $\ScrH_{\mathsf{mix}}^2$, the order-2 Sobolev space with dominating mixed smoothness.

If the simulation samples are noise-free,
our algorithm benefits automatically from the higher-order smoothness without any modification, exhibiting a faster convergence rate.

\begin{theorem}\label{thm:EI_noiseless_misspecification}
Suppose that $f\in\ScrH^2_{\mathsf{mix}}$, Assumption~\ref{assump:optimum} holds, and Assumption~\ref{assump:sub-Gaussian} holds with $\sigma=0$.
Let $\lambda=0$  and $\delta_n = 1$ in Algorithm~\ref{alg:EI}.
Then,
\[\E[f(\BFx^*)-f(\widehat{\BFx}^*_N)]=\CalO\left(N^{-\frac{3}{2}}(\log N)^{\frac{5(d-1)}{2}}\right).\]
\end{theorem}

If the simulation samples are noisy,
our algorithm can also benefit from the higher-order smoothness.
However,
such information needs to be incorporated into Algorithm~\ref{alg:EI}, which is reflected through the tuning parameters $\lambda$ and $\delta_n$, in order to achieve a faster convergence rate.

\begin{theorem}
\label{thm:Convergence_EI_misspecification}
Suppose that $f\in\ScrH^2_{\mathsf{mix}}$, Assumption~\ref{assump:optimum} holds, and Assumption~\ref{assump:sub-Gaussian} holds with $\sigma>0$.
Let $\lambda \asymp \left(\sigma^{4}N_{\tau}^{-2}|\log (\sigma N_\tau)|^{{2d-1}}(\log N_\tau)^{6(1-d)}\right)^{\frac{1}{5}}$
and $\delta_n\asymp\left( \sigma^6 n^{-3} |\log (\sigma n)|^{{1-2d}}(\log n)^{6(d-1)}\right)^{\frac{1}{10}}$ in Algorithm~\ref{alg:EI}.
Then,
\begin{align*}
\E[f(\BFx^*)-f(\widehat{\BFx}^*_N)]
= \CalO\left(\sigma^{\frac{3}{5}}N^{-\frac{3}{10}}|\log (\sigma N)|^{\frac{3(2d-1)}{20}}(\log N)^{\frac{9(d-1)}{10}}\right).
\end{align*}
\end{theorem}

\subsection{Summary of Convergence Rates}

Table~\ref{tab:summary} presents the results of Theorems~\ref{thm:EI_noiseless}--\ref{thm:Convergence_EI_misspecification}.
We use the notation $\widetilde{\CalO}$, which ignores the logarithmic terms to highlight the dominating terms.
Namely, $a_n = \widetilde{\CalO}(b_n)$ if
$a_n = \CalO(b_n (\log(n))^c)$ for some $c>0$.
Notably, the dimensionality affects only the logarithmic terms, and it is thus hidden from the polynomial decaying terms.
\begin{table}[ht]
\TABLE
{Convergence Rate of \texttt{KEIBS}.\label{tab:summary}}
{
\begin{tabular}{ccccc}
\toprule
Smoothness Parameter && Simulation Noise? && Convergence Rate \\
\midrule
$m=1$ && No && $\widetilde{\CalO}\bigl(N^{-\frac{1}{2}}\bigr)\;$ \\
$m=1$ && Yes &&  $\widetilde{\CalO}\bigl(N^{-\frac{1}{6}}\bigr)\;$ \\
$m=2$ && No && $\widetilde{\CalO}\bigl(N^{-\frac{3}{2}}\bigr)\;$ \\
$m=2$ && Yes &&  $\widetilde{\CalO}\bigl(N^{-\frac{3}{10}}\bigr)$ \\
\bottomrule
\end{tabular}
}
{The rates are achieved by properly specifying $\lambda$ and $\delta_n$; see Theorems~\ref{thm:EI_noiseless}--\ref{thm:Convergence_EI_misspecification}.}
\end{table}

\section{Numerical Experiments}\label{sec:num}

In this section, we numerically assess the performance of \texttt{KEIBS} (Algorithm~\ref{alg:EI}) for high-dimensional SO problems.
In all the experiments below, we set the parameters of the BF kernel \eqref{eq:BM-kernel} as $\theta_j=\gamma_j=1$ for all $j=1,\ldots,d$.
In the last step of \texttt{KEIBS}---after  all simulation budget is exhausted---we
solve $\widehat{\BFx}_N^* = \argmax_{\BFx\in\ScrX} \widetilde{f}_N(\BFx)$ using
the function \texttt{spcompsearch} in the
Sparse Grid Interpolation Toolbox of \textsc{Matlab} \citep{Klimke07}.
(Note that $\widetilde{f}_N(\BFx)$ is
a piecewise multilinear function,
because it is expressed as a linear combination of functions of the form $k(\BFx_i,\BFx)$ which is a piecewise multilinear function by definition.
This \textsc{Matlab} function is specifically designed for optimizing such functions.)

We compare \texttt{KEIBS} (Algorithm~\ref{alg:EI})  with three alternatives  that are popular Bayesian optimization (BO) methods in machine learning literature for solving black-box optimization problems.
\begin{enumerate}[label=(\roman*)]
    \item \texttt{EI-plus}:  A refinement of the EI strategy \citep{Bull11} via an adaptive adjustment of the kernel variance to avoid over-exploiting a particular area of the design space. The implementation of this algorithm is available through the function \texttt{bayesopt} in the Statistics and Machine Learning Toolbox\textsuperscript{\texttrademark} of \textsc{Matlab}.
    \item \texttt{REMBO}: Random Embedding Bayesian Optimization \citep{WangHutterZoghiMathesonFeitas16}. This algorithm is developed specifically to address high-dimensional BO problems. It is similar to \texttt{EI-plus} but works under the additional assumption that the function of interest takes the form of a low-dimensional embedding; that is, $f(\BFx)=f(\BFA \BFu)$ where $\BFu\in\Real^\ell$  and $\BFA\in\Real^{d\times \ell}$ with $\ell\ll d$ is a matrix with unknown entries, representing linear low-dimensional mapping. In each iteration of the algorithm, the matrix $\BFA$ is updated adaptively, and
    the EI criterion is evaluated and maximized as a function of $\BFu$. The \textsc{Matlab} code of $\texttt{REMBO}$ is released by the authors of the paper at \url{https://github.com/ziyuw/rembo}.
     \item \texttt{GP-UCB}: Gaussian Process Upper Confidence Bound \citep{SrinivasKrauseKakadeSeeger12}. This BO algorithm iteratively chooses the next design point by maximizing an upper confidence bound of the unknown function.
    It is also available through the \textsc{Matlab} function \texttt{bayesopt}.
\end{enumerate}
Following a common practice \citep{SnoekLarochelleAdams12},
for all the three methods, we use the Mat\'ern($\sfrac{5}{2}$) kernel with parameters estimated from the data.
This is also the default choice in \textsc{Matlab}.

The numerical comparison is performed on three test problems in ascending order of  dimensionality.
We consider a 20-dimensional production line problem in Section~\ref{sec:prod-line},
a 50-dimensional product assortment problem in Section~\ref{sec:assortment}, and
two 100-dimensional artificial test functions in Section~\ref{sec:test-func}.
In all three examples, \texttt{KEIBS} outperforms the competing methods by a substantial margin both in terms of the optimality gap between the returned solution and the global optimum and in terms of the computational speed.
All the experiments are implemented in \textsc{Matlab} (version 2018a) on a laptop computer with macOS, 3.3 GHz Intel Core i5 CPU, and 8 GB of RAM (2133Mhz).

\subsection{20-Workstation Production Line}\label{sec:prod-line}
In this subsection, we consider a production line problem from the SimOpt Library (\url{www.simopt.org}).
The production line is modeled as a tandem queueing system having a sequence of $d=20$  workstations.
Each workstation is modeled as a single-server finite-capacity queueing system with  the first-in-first-out discipline and exponentially distributed service times.
Suppose that each of the workstations has  a capacity $K$ but they may have different service rates, which are the design variables of interest, denoted by $\BFx = (x_1,\ldots,x_d)$.
Suppose also that parts arrive at workstation 1 following an exogenous Poisson process with rate $\alpha$.
Upon completing the service at workstation $i$, a part is moved to workstation $i+1$ unless there are already $K$ parts at the downstream workstation,
in which case the part will stay at workstation $i$, occupying the server there and blocking other parts from receiving service.

The manager of the production line may increase the service rates to boost the throughput---the number of parts that complete the service from all the workstations---but at a higher operating cost.
The objective is to maximize the expected revenue from running the production line for a duration of $T$ time units.
Specifically, the objective function is modeled as
\[f(\BFx) = \E\biggl[\frac{r\cdot \mathsf{Th}(\BFx)}{c_0 + \BFc^\intercal \BFx} \biggr],\]
where $\mathsf{Th}(\BFx)$ denotes the throughput during $T$ time units, $r$ represents the revenue from each completed part,
$\BFc=(c_1,\ldots,c_d)$ denotes the cost parameters associated with each workstation, and $c_0$ is some fixed cost.

The parameters are specified as follows. The design space is $\ScrX =  (0, 2)^d$,
$K=10$, $r=2\times 10^5$, $c_0=1$, and $c_i=i$ for $i=1,\ldots,d$.
Moreover, we consider the following two scenarios having different levels of simulation noise.
\begin{enumerate}[label=(\roman*)]
    \item Low-noise: $\alpha=0.5$ and $T=1000$.
    \item High-noise: $\alpha=2$ and $T=200$.
\end{enumerate}
The simulation budget is set to be $N=100, 200, \ldots, 500$.

For each experimental setup, we run the four algorithms $R=20$ times and call time a \emph{macro-replication}.
For the $r$-th macro-replication, let  $\widehat{\BFx}^*_{N,r}$ denote the solution that an algorithm returns after $N$ simulation samples are collected.
The performance of each algorithm is assessed via the average estimated optimal value (AEOV):
\[\widehat{\text{AEOV}}\coloneqq\frac{1}{R}\sum_{r=1}^R f(\BFx^*_{N,r}).\]
Because $f$ has no analytical form, we use 100 simulation replications to estimate $f(\BFx^*_{N,r})$ and treat the sample mean as the true value. (The standard deviation (SD) of these replications is sufficiently small.)
Moreover, we use the sample SD of these quantities to assess the stability of the performance:
\[\widehat{\rm SD}=\biggl(\frac{1}{R-1}\sum_{r=1}^R\Bigl(f(\BFx^*_{N,r})-\widehat{\text{AEOV}}\Bigr)^2\biggr)^{\frac{1}{2}}.\]

\begin{figure}[ht]
\FIGURE{
$\begin{array}{cc}
     \includegraphics[width=0.4\textwidth]{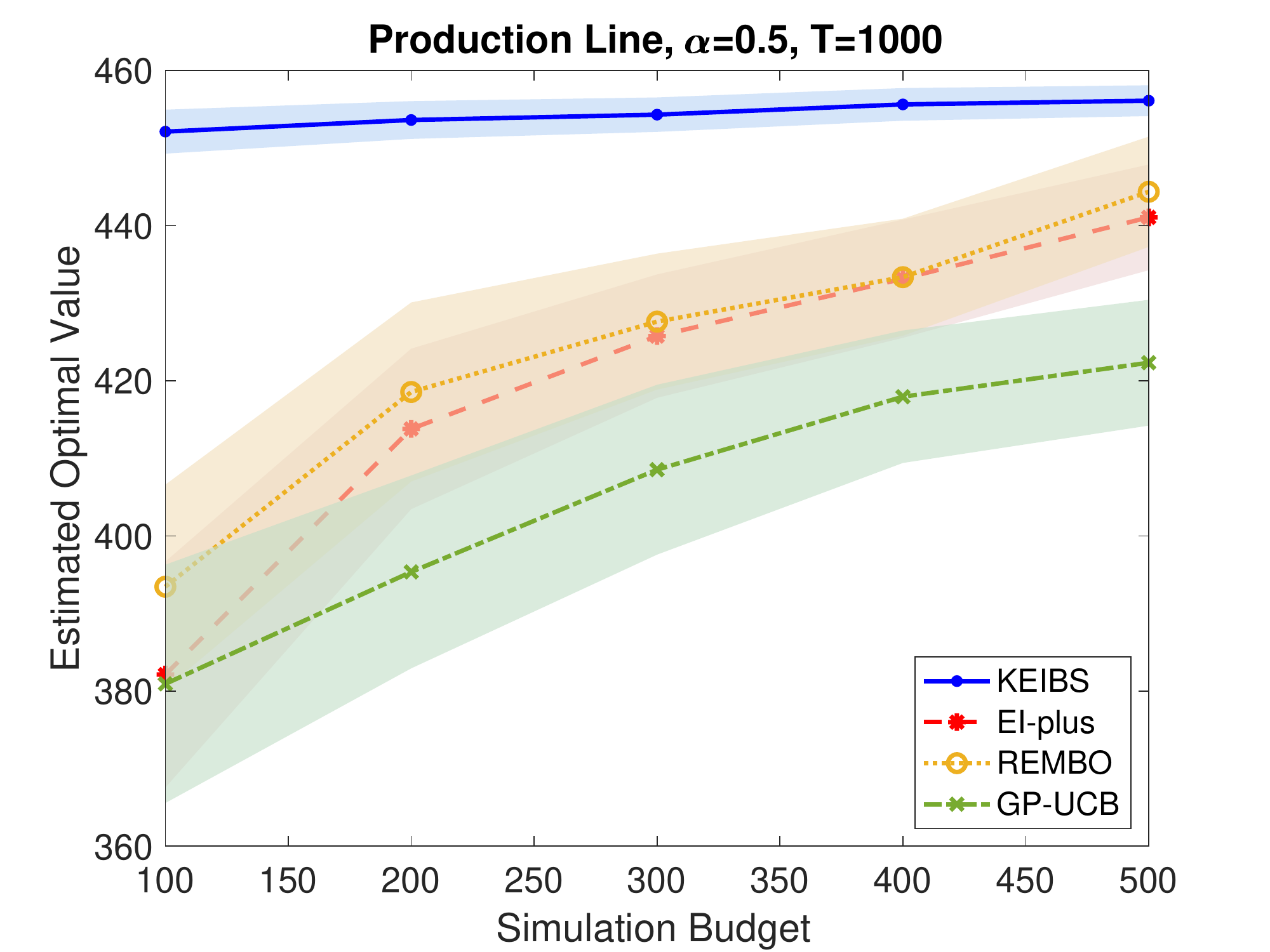} &
     \includegraphics[width=0.4\textwidth]{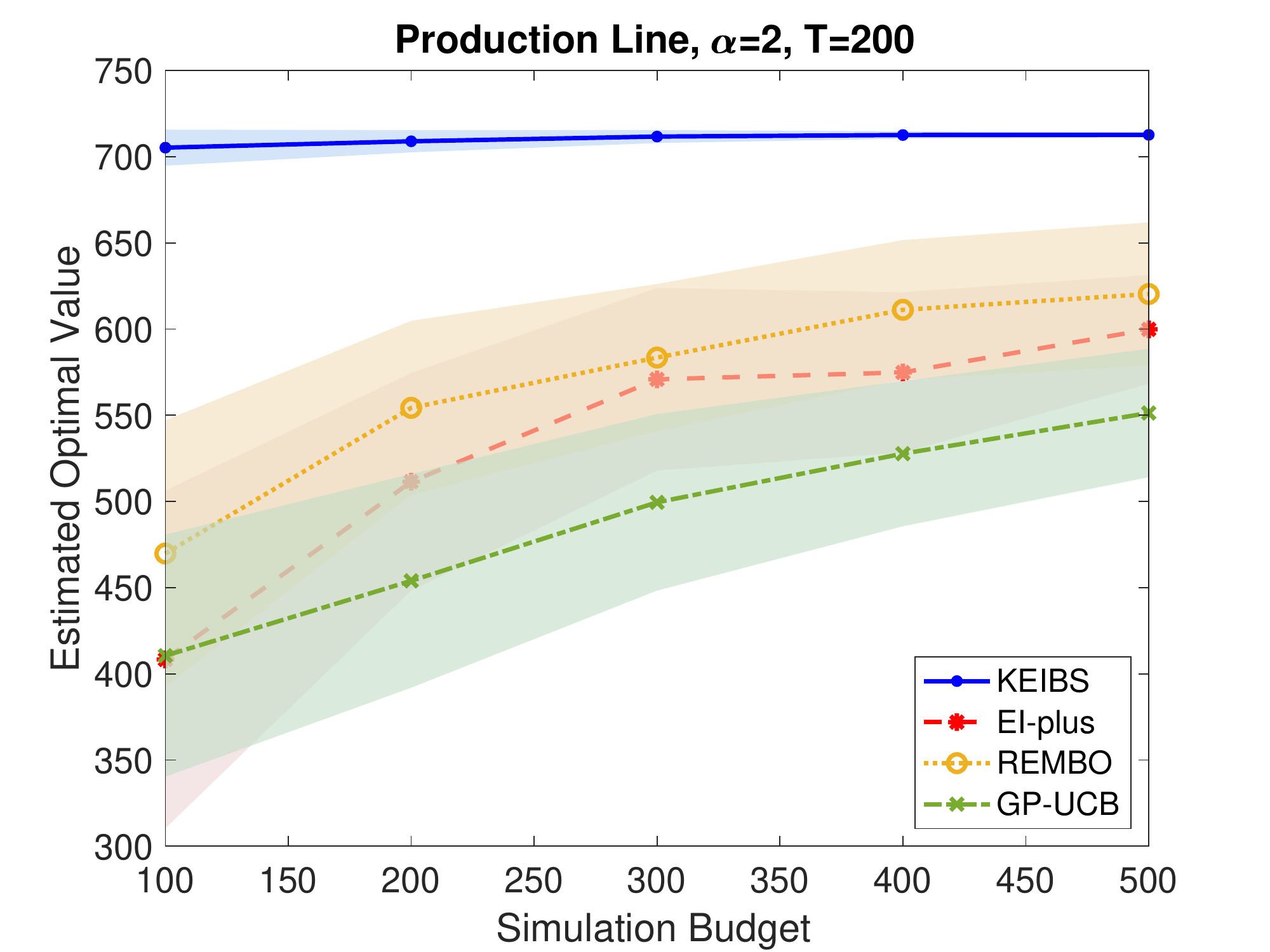}
\end{array}
$
}
{Estimated Optimal Values for the Production Line Problem. \label{fig:prod-line} }
{
The shaded areas have a half-width that equals $\widehat{\text{SD}}$.
}
\end{figure}

The numerical results are shown in  Figure~\ref{fig:prod-line}.
In terms of $\widehat{\text{AEOV}}$,
\texttt{KEIBS} significantly outperforms the others
in all the
tested cases of different simulation budgets and noise levels.
In particular, in this 20-dimensional SO problem, \texttt{KEIBS} is able to find a
solution that is  very close to the true optimum with a simulation budget of less than 100.
Being designed for high-dimensional problems, \texttt{REMBO} performs better than both \texttt{EI-plus} and \texttt{GP-UCB} in general.
Nevertheless, the quality of the solutions returned by \texttt{REMBO} still
falls far behind those returned by \texttt{KEIBS}.
This may be because the underlying assumption of \texttt{REMBO}---that the objective function has a low-dimensional embedding---is invalid for the production line problem.

In addition, the shaded areas that correspond to \texttt{KEIBS} in both the left and right panels of Figure~\ref{fig:prod-line} are substantially narrower than those of the other algorithms.
Since the half-width of a shaded area represents $\widehat{\text{SD}}$,
this indicates that \texttt{KEIBS} can identify a good solution with a much smaller budget, and it does so in a much more stable manner.

\begin{figure}[ht]
\FIGURE{
\includegraphics[width=0.4\textwidth]{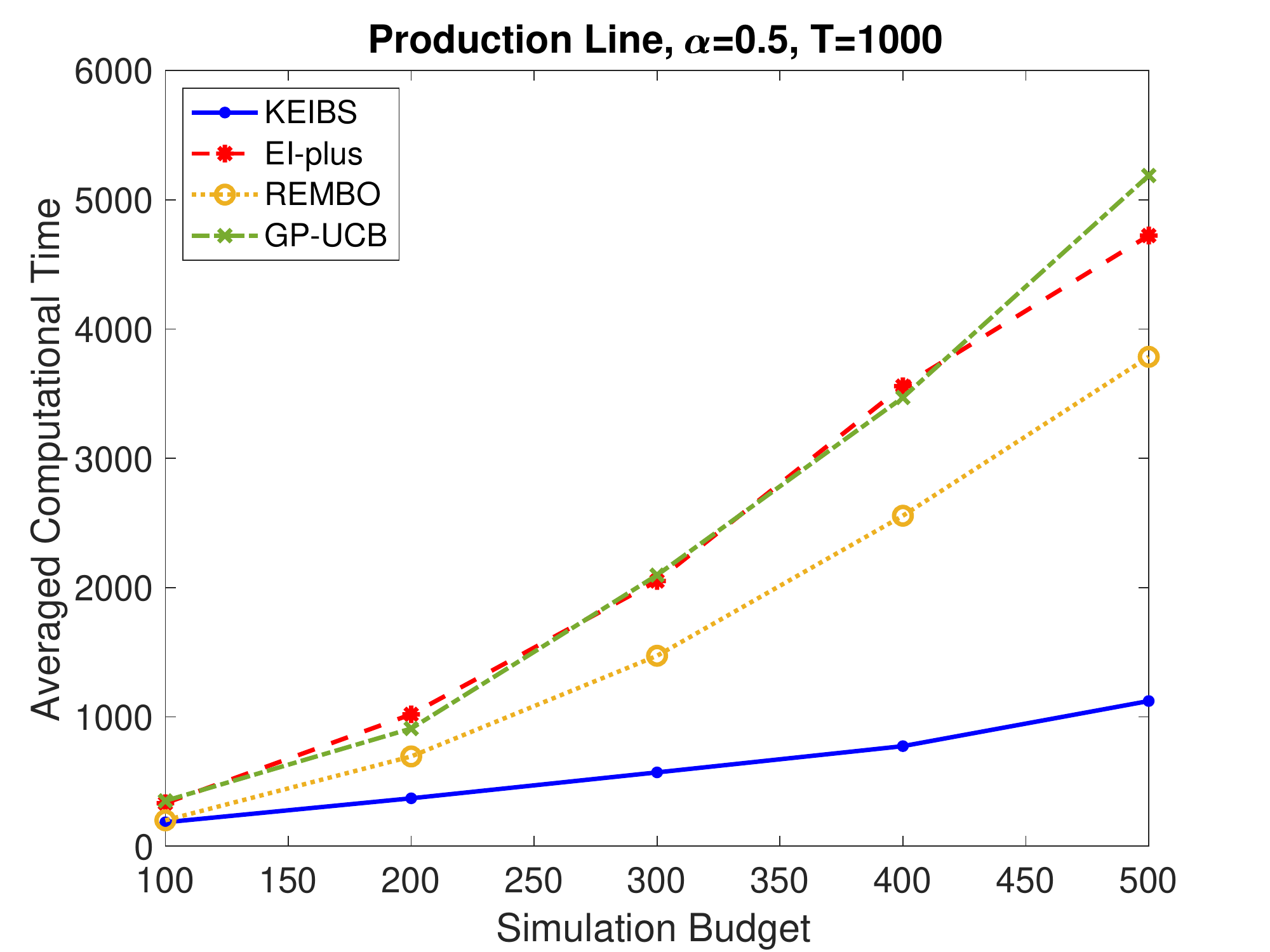}
}
{Computational Time for the Production Line Problem. \label{fig:compu-time} }
{Time unit: second.}
\end{figure}

In Figure~\ref{fig:compu-time}, we plot for the case of $\alpha=0.5$ and $T=1000$ the average computational time used by each algorithm to complete a search process with a given simulation budget.
Note that the elapsed time reported here includes the time spent running the simulator to generate noisy samples. But the simulation time for the production line time is much shorter, so it does not change the conclusion that \texttt{KEIBS} has an obvious advantage over the other algorithms in terms of computational efficiency.

Recall that there are two computationally heavy tasks in each iteration of \texttt{EI-plus}, \texttt{REMBO}, and \texttt{GP-UCB}.
First, kernel matrices need to inverted numerically to compute the GP posterior of the objective function, and the matrices grow with the sample size.
Second, the acquisition function (EI or UCB) needs to be optimized numerically over the design space, which is high-dimensional, to determine the next design point.
By contrast, \texttt{KEIBS} is computationally much faster, thanks to (i) the joint use of BF kernels and SGs that accelerates the computation of large-scale linear algebra, and (ii) the fact that the next design point is chosen among a discrete set of candidates from an SG.

\subsection{50-Product Assortment}\label{sec:assortment}

In this subsection, we consider the problem described in Example~\ref{exmp:ProdAssort} in Section~\ref{sec:Brownian} with $d=50$.
Specifically, we maximize the following expected profit as a function of the price vector:
\[
f(\BFx) = \frac{1}{2(b-a)}\sum_{j=1}^d \left[ (b-a)\left(\frac{x_j-c_j}{x_j}\right) + a \right]^2 Q_j^2(\BFx).
\]
We set the relevant parameters as follows: $a=100$, $b=400$, $\alpha_j = 10.5 + 0.5(j-1)$, and $c_j=6.5+0.5(j-1)$, for $j=1,\ldots,d$. Also,
the design space is $\ScrX = \bigtimes_{j=1}^d (h_j, h_j + 10)$, where $h_j = 9 + 0.5(j-1)$. (Note that the global optimum is included in this region.)

We set the simulation budget to be $N=500, 1000, 1500, 2000$.
Because the objective function $f$ in this example has an analytical form,
for simplicity, we generate noisy samples by adding an artificial zero-mean Gaussian noise term to $f$.
Specifically, we assume that the observations are
normal random variables with heterogeneous variances:
$y(\BFx_i) \sim \mathsf{Normal}\bigl(f(\BFx_i), \zeta |f(\BFx_i)| \bigr)$, where the variance is proportional to $|f(\BFx_i)|$.
We set the parameter $\zeta$ to be 0.01 or 0.1 to represent different noise levels.

\begin{figure}[ht]
\FIGURE{
$\begin{array}{cc}
     \includegraphics[width=0.4\textwidth]{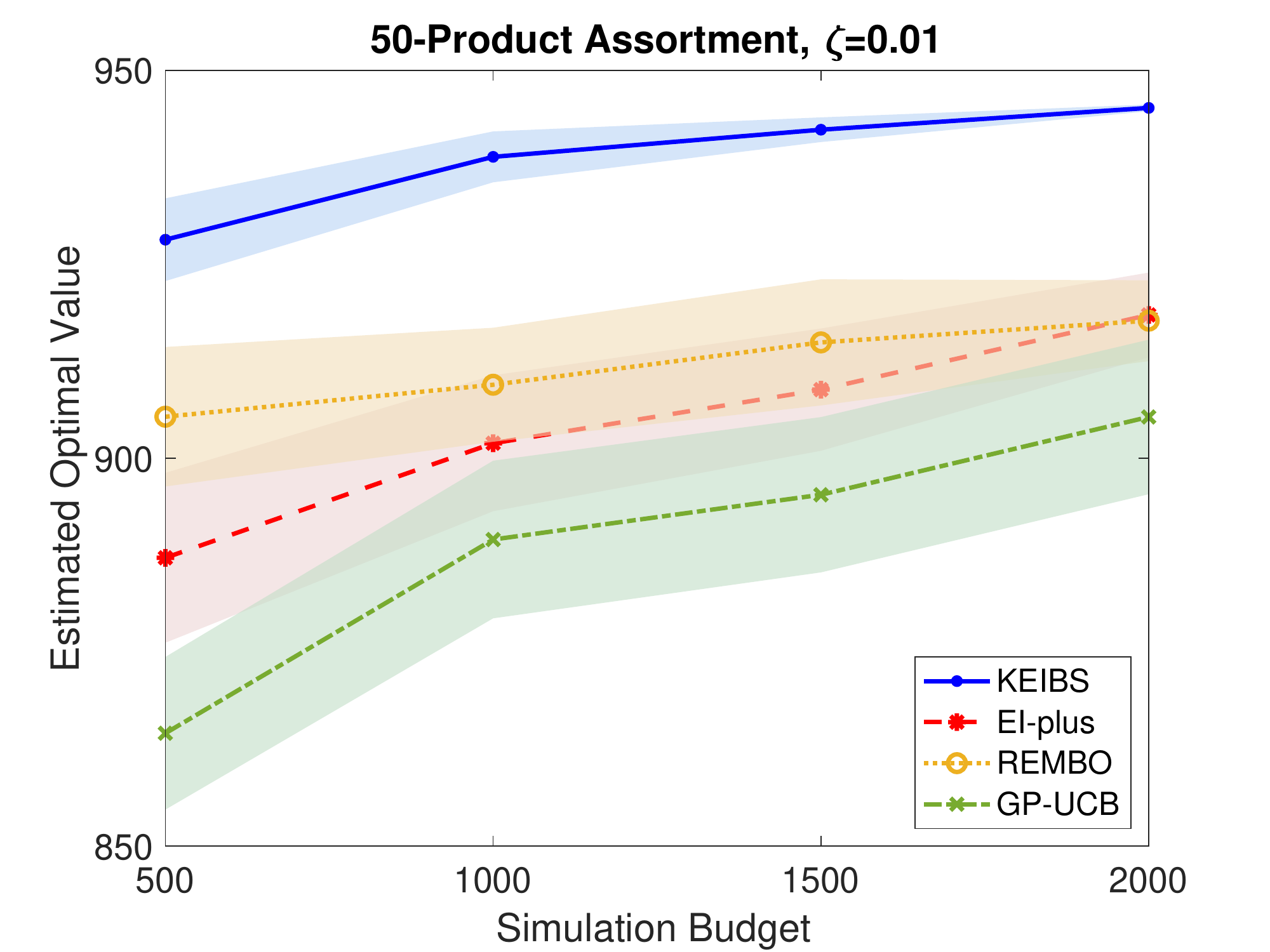} &
     \includegraphics[width=0.4\textwidth]{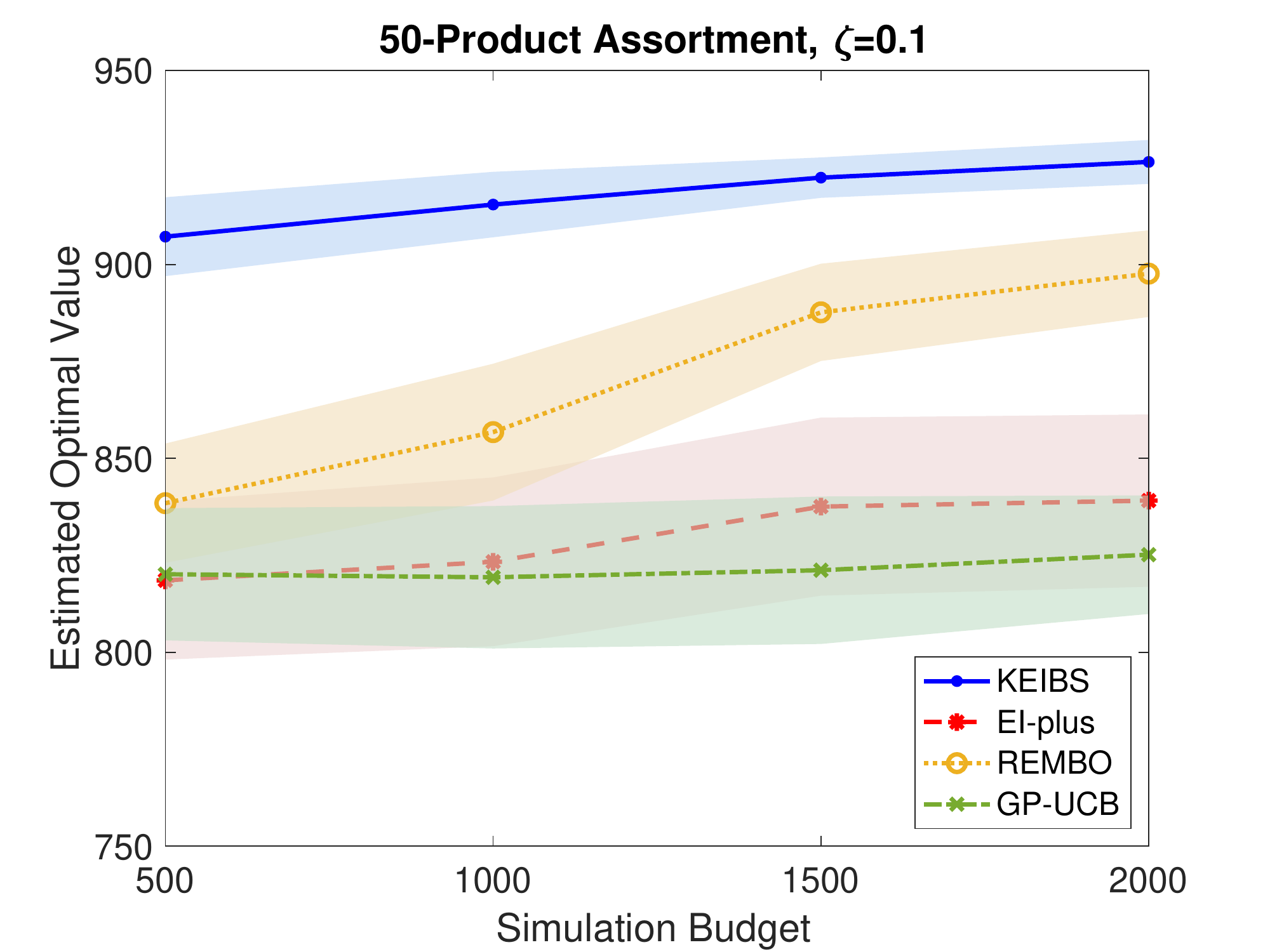}
\end{array}
$
}
{Estimated Optimal Values for the Assortment Problem. \label{fig:Product_Assortment} }
{
The shaded areas have a half-width that equals $\widehat{\text{SD}}$.}
\end{figure}

Similar to the experimental setup in Section~\ref{sec:prod-line},
we use $R=20$ macro-replications to compute
$\widehat{\text{AEOV}}$ and $\widehat{\text{SD}}$ for the four algorithms.
The results are presented in Figure~\ref{fig:Product_Assortment}.
Again,
\texttt{KEIBS} exhibits in this high-dimensional problem a substantial advantage relative to the other algorithms in terms of both the quality of the returned solution and the stability of the performance.

\subsection{Test Functions in 100 Dimensions}\label{sec:test-func}
\begin{figure}[ht]
\FIGURE{
\includegraphics[width=0.4\textwidth]{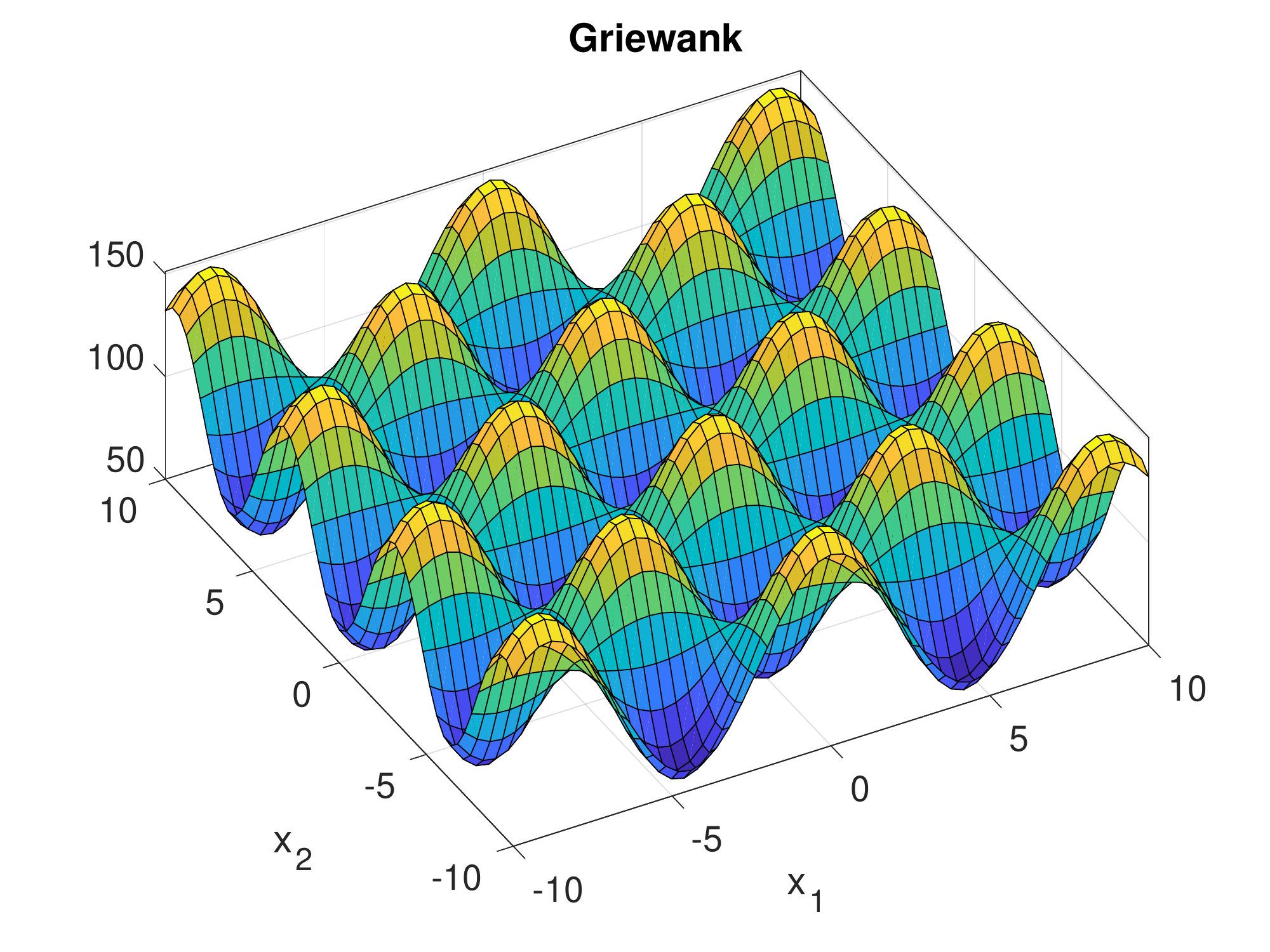}
\includegraphics[width=0.4\textwidth]{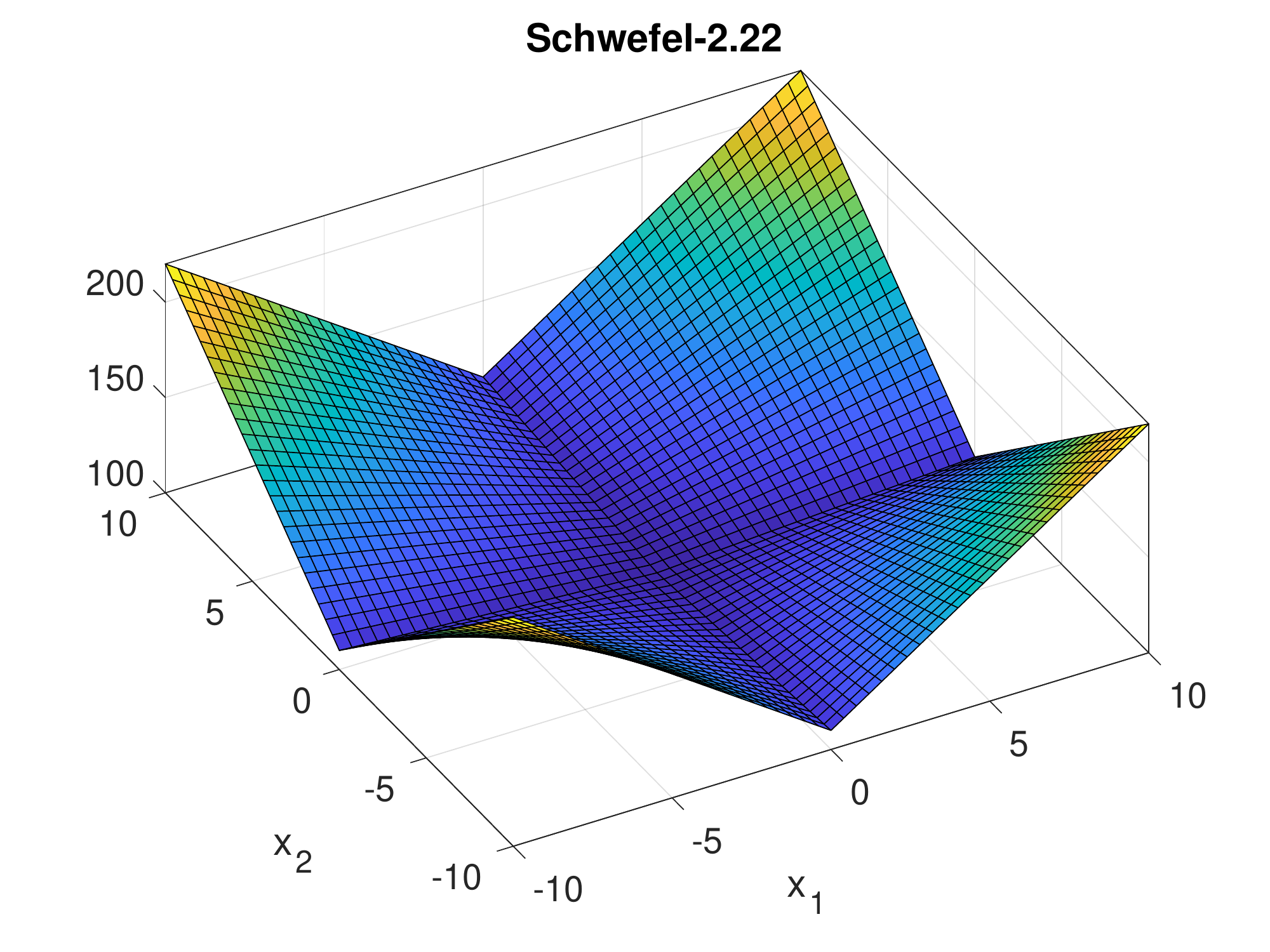}
}
{The Griewank and  Schwefel-2.22 Functions in Two Dimensions. \label{fig:test-func-surfaces}}
{}
\end{figure}

In this subsection, we challenge \texttt{KEIBS} using a very high-dimensional setting.
We consider the problem of minimizing the following two test functions in $d=100$ dimensions (see Figure~\ref{fig:test-func-surfaces} for their two-dimensional projections):
\begin{align*}
    & f_{\text{Griewank}}(\BFx)=50\bigg[\sum_{j=1}^d\frac{x_j^2}{4000}-\prod_{j=1}^d\cos(\frac{x_j}{j})+1\bigg], \quad \BFx\in(-10,10)^d, \\
    & f_{\text{Schwefel-2.22}}(\BFx)=\sum_{j=1}^d|x_j|+\prod_{j=1}^d|x_j|+100, \quad \BFx\in(-10,10)^d.
\end{align*}

The two test functions are chosen because they are qualitatively different. The Griewank function is infinitely differentiable and has many local minima, whereas the Schwefel-2.22 function is non-differentiable and has a unique local minimum.

Note that both functions are minimized at the origin $\BFx=\BFzero$.
The use of sparse grids in \texttt{KEIBS} guarantees that the origin is always sampled because it is the center of the design space $(-10,10)^d$ (see Figure~\ref{fig:SG}).
This creates an unfair advantage for \texttt{KEIBS} relative to the alternatives if these two test functions are used in the original form.
Therefore, for each test function, we generate $R=50$ random problem instances via a simple change of variables, which effectively moves the global minimum to a random position while retaining the shape of the function.
Specifically, for each $r=1,\ldots,R$, we define
\[
f_{\text{G},r}(\BFx) \coloneqq f_{\text{Griewank}}\biggl(\BFx + \frac{\BFu_r}{\sqrt{d}}\biggr) \qq{and}
f_{\text{S},r}(\BFx) \coloneqq f_{\text{Schwefel-2.22}}\biggl(\BFx + \frac{\BFu_r}{\sqrt{d}}\biggr),\quad \BFx\in(-10,10)^d,
\]
where $\BFu_r\in\Real^d$ is a vector of independent random variables uniformly distributed on $(-1,1)$.
In other words, each $f_{\text{G},r}$ (resp., $f_{\text{S},r}$) is a random variation of $f_{\text{Griewank}}$ (resp., $f_{\text{Schwefel-2.22}}$) with the global minimum being relocated to the position $\frac{\BFu_r}{\sqrt{d}}$.

For each problem instance, we assume that the function value is observed with heterogeneous Gaussian noise and a variance proportional to the function value.
For example, if $f = f_{\text{G},r}$, then the sample at $\BFx_i$ is
$y(\BFx_i) \sim \mathsf{Normal}\bigl(f(\BFx_i), \zeta^2 f^2(\BFx_i)\bigr)$.
We set $\zeta$ to be $0.1$ or $1$.

\begin{figure}[ht]
\FIGURE{
$\begin{array}{cc}
     \includegraphics[width=0.4\textwidth]{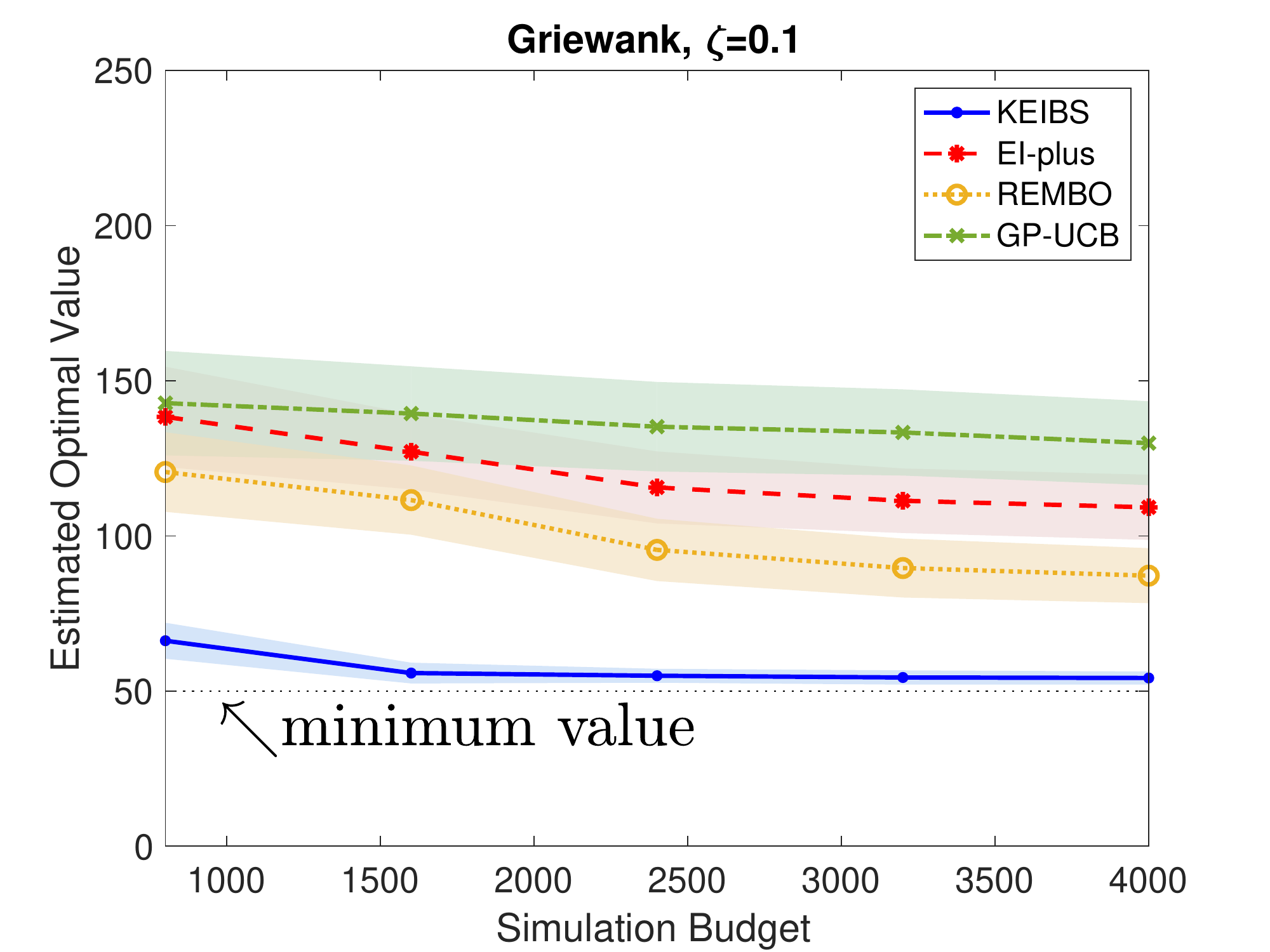} &
     \includegraphics[width=0.4\textwidth]{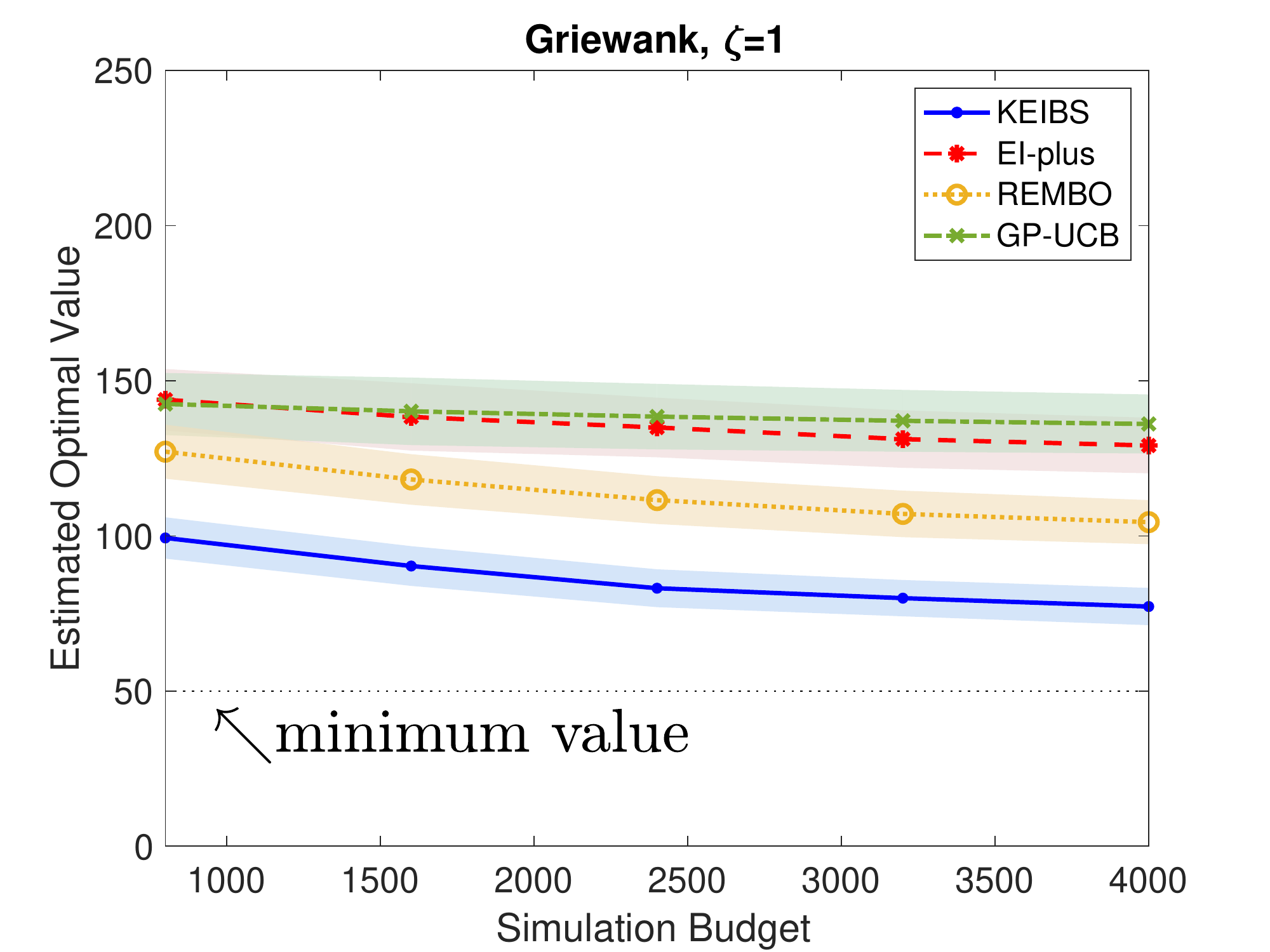} \\
     \includegraphics[width=0.4\textwidth]{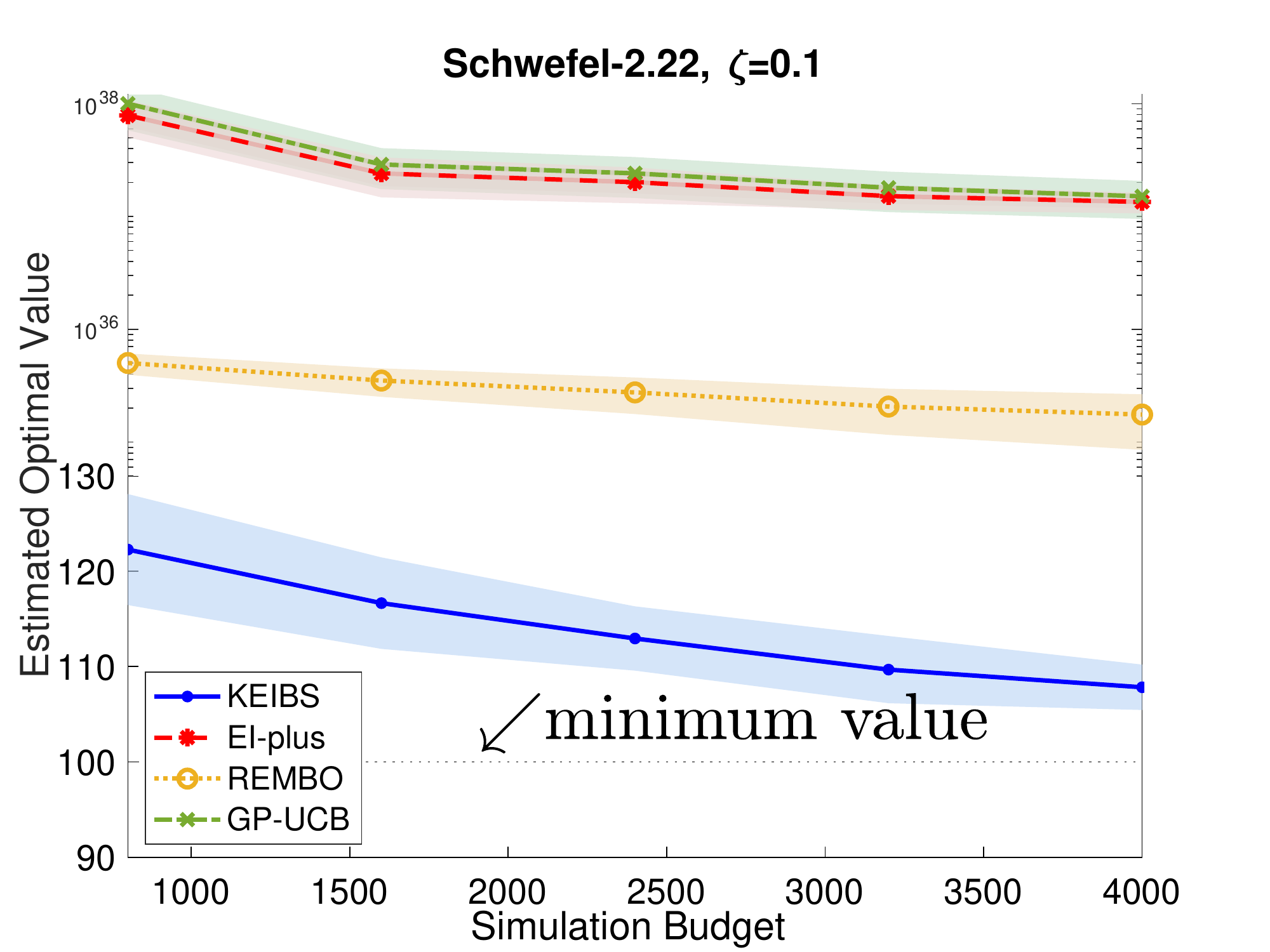} &
     \includegraphics[width=0.4\textwidth]{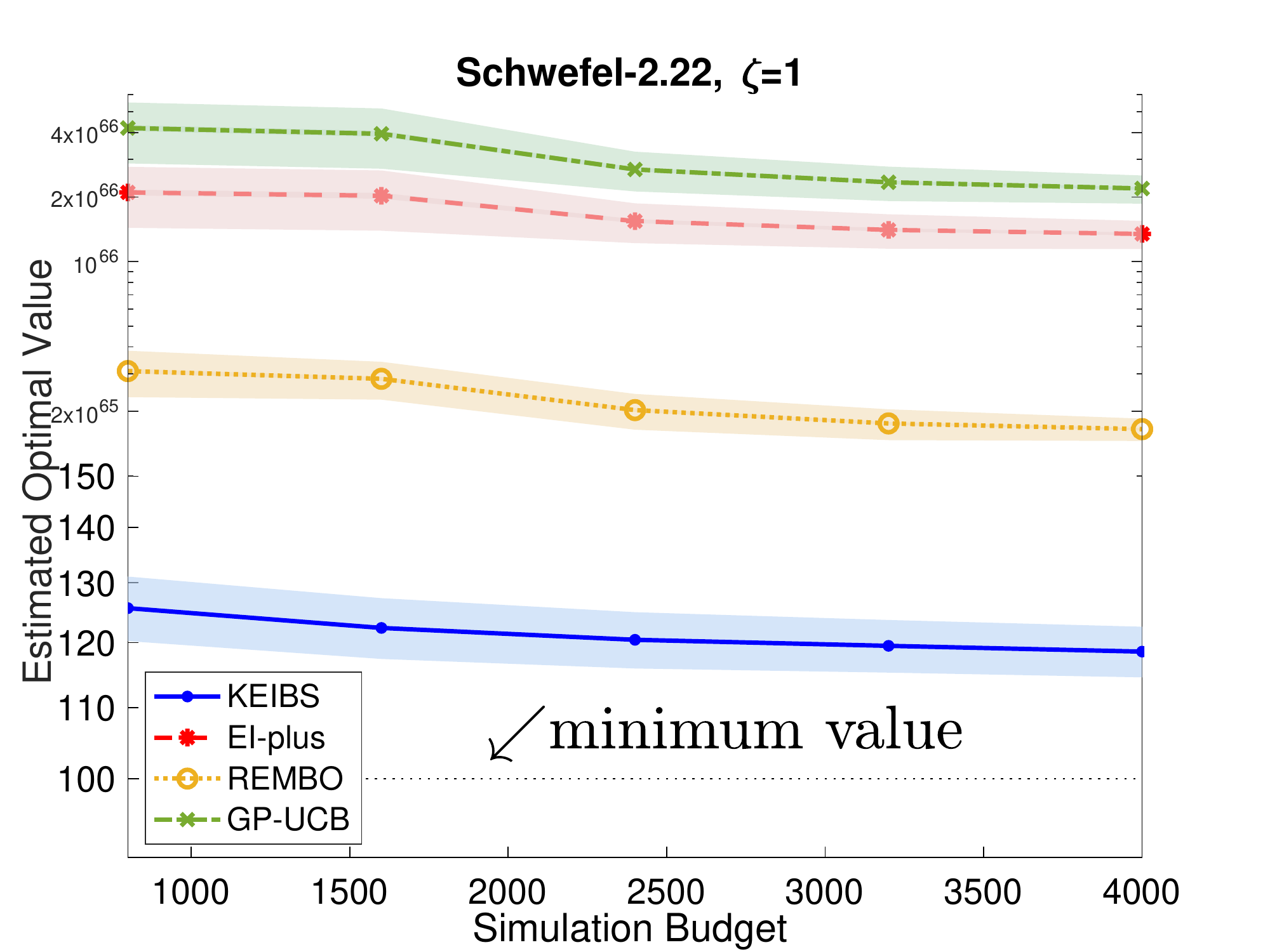}
\end{array}
$
}
{Estimated Optimal Values for the Test Functions. \label{fig:test_func}}
{
The shaded areas have a half-width that equals $\widehat{\text{SD}}$.}
\end{figure}

Then, we run the four algorithms on each problem instance with a simulation budget $N=800,1600,\ldots,4000$.
Lastly, we compute $\widehat{\text{AEOV}}$ and $\widehat{\text{SD}}$ based on the results from the $R$ problem instances.
Doing so ensures a fair comparison of the algorithms.
Their performances are visualized in Figure~\ref{fig:test_func}.

First,  similar to the experiments in Section~\ref{sec:prod-line} and Section~\ref{sec:assortment}, \texttt{KEIBS} exhibits a dominating performance over \texttt{EI-plus}, \texttt{REMBO}, and \texttt{GP-UCB} for both test functions and under both low-noise and high-noise scenarios.
Second, the competing algorithms are able to find reasonable solutions for the Griewank function, but they fail completely for the Schwefel-2.22 function. (Note that the upper part of the vertical axis of the lower panel of Figure~\ref{fig:test_func} is at a logarithmic scale.)
The failure for the latter is largely because the Schwefel-2.22 function is non-differentiable, having a very rough landscape in high dimensions.
Third, under the low-noise scenario ($\zeta=0.1$)---where the standard deviation of the noise is 10\% of the true function value---\texttt{KEIBS} can quickly approach the minimum.
By contrast, under the high-noise scenario ($\zeta=1$), the solution returned by  \texttt{KEIBS}  after collecting 4000 simulation samples still falls short of the optimal by a sizable gap, demonstrating the impact of noise the quality of the solution.
Finally, we remark that, for simulation budget $N>4000$, the three alternatives, especially \texttt{EI-plus}, are excessively demanding in terms of computation in this very high-dimensional setting.
By contrast, \texttt{KEIBS} is computationally fast enough to handle an even larger sample size.

\section{Concluding Remarks}\label{sec:conclusions}

It is challenging to solve high-dimensional SO problems due to two fundamental difficulties.
From a statistical viewpoint, simulation samples are usually expensive,  yet the number of samples that is needed in constructing an accurate estimate of the response surface  generally grows exponentially with the dimensionality.
Meanwhile, from a computational viewpoint, given a large number of simulation samples,
it may still be time-consuming to process them and to determine subsequent design points in a principled, dynamic fashion, because this often involves large-scale linear algebra and high-dimensional numerical optimization.

In this paper we propose a novel SO algorithm (i.e., \texttt{KEIBS}) based on the joint use of Brownian fields and sparse grids to address these two challenges simultaneously.
In particular, we establish upper bounds on the convergence rate of \texttt{KEIBS} for both the cases of matched smoothness and higher-order smoothness.
These upper bounds indicate mild dependence of the sample complexity of \texttt{KEIBS} on the dimensionality.
The theoretical findings are corroborated by extensive numerical experiments.

In addition, \texttt{KEIBS} is computationally fast thanks to two features that distinguish it from other algorithms.
First, in \texttt{KEIBS} the numerical inversion of large kernel matrices---which is common for most BO algorithms---is done by exploiting the sparse structure of the inverse kernel matrix associated with Brownian field kernels and sparse grids.
This gives rise to fast, exact computations without resorting to kernel approximation methods.
Second, in the sequential sampling stage of \texttt{KEIBS}, selecting subsequent design points is formulated as a discrete optimization problem over a sparse grid, rather than a continuous one over a high-dimensional space.
Not only does this avoid the considerable computational overheads involved in performing numerical optimization,
but it also obviates the concern of the numerical solver converging to a local optimum, as such an optimization problem is usually very non-convex.

Looking forward, we believe the following research problems will be  of great impact.
First, the asymptotic analysis in the present paper focuses on the optimality gap between the returned solution and a global optimum.
Another important metric---especially in an online environment---is the cumulative regret, which basically measures the sum of the optimality gap for all the intermediate solutions visited by an algorithm.
It is of great interest to perform such analysis on \texttt{KEIBS}.

Second, it is known in BO literature that no acquisition functions are better than the others for all problem instances.
\texttt{KEIBS} can easily be  modified to use other acquisition functions (e.g., UCB) in the sequential sampling stage.
Fast computations would still be available.
Our theoretical analysis might also be  extended to cover the modified algorithms.

Lastly, we  prove only upper bounds on the convergence rate of \texttt{KEIBS}. A minimax lower bound is needed to deepen the theoretical understanding of the algorithm. This would reveal whether our upper bounds are tight,
and it might shed new light on how to further improve the algorithm.
However, new analysis techniques appear to be needed for this purpose.

% Appendix here
% Options are (1) APPENDIX (with or without general title) or
%             (2) APPENDICES (if it has more than one unrelated sections)
% Outcomment the appropriate case if necessary
%
% \begin{APPENDIX}{<Title of the Appendix>}
% \end{APPENDIX}
%
%   or
%

\begin{APPENDICES}

\section{Reproducing Kernel Hilbert Spaces }\label{sec:RKHS}

In this section, we provide an overview of reproducing kernel Hilbert spaces and refer to \cite{BerlinetThomas-Agnan04} for an extensive treatment of the subject.

\subsection{Definition}
Let $\ScrX\subseteq \Real^d$ be a nonempty set
and
$k:\ScrX\times \ScrX\mapsto \Real$ be a positive definite kernel; that is, it is a symmetric function, and
\[\sum_{i,j=1}^n \beta_i \beta_j k(\BFx_i,\BFx_j) \geq 0,\quad \forall n\in\NatInt,\ \{\BFx_i\}_{i=1}^n \subset\ScrX,\ \{\beta_i\}_{i=1}^n\subset\Real.\]

\begin{definition}
A Hilbert space $\ScrH_k$ of functions on $\ScrX$ that is endowed with an inner product $\langle\cdot,\cdot \rangle_{\ScrH_k}$ is said to be a reproducing kernel Hilbert space (RKHS) with kernel $k$ if
\begin{enumerate}[label=(\roman*)]
    \item $k(\cdot, \BFx)\in \ScrH_k$ for all $\BFx\in\ScrX$ and
    \item the reproducing property holds; that is, $g(\BFx) = \langle g, k(\cdot, \BFx) \rangle_{\ScrH_k}$ for all $\BFx\in\ScrX$ and $g\in\ScrH_k$.
\end{enumerate}
\end{definition}

The Moore-Aronszajn theorem \citep{Aronszajn50} implies that there is a one-to-one relationship between RKHSs and positive definite kernels: for each positive definite kernel $k$ there exists a unique RKHS $\ScrH_k$ that is induced by $k$, and vice versa.

Given a positive definite kernel $k$, the associated RKHS can be constructed as follows.
Let $\ScrF_k$ denote the linear space spanned by $\{k(\cdot, \BFx):\BFx\in\ScrX\}$:
\[\ScrF_k\coloneqq\biggl\{g=\sum_{i=1}^n\beta_i k(\cdot,\BFx_i):\  n\in\NatInt,\ \{\BFx_i\}_{i=1}^n \subset\ScrX,\ \{\beta_i\}_{i=1}^n\subset\Real \biggr\}.\]
We endow $\ScrF_k$ with the following inner product.
For any $g=\sum_{i=1}^{n}\beta_i k(\cdot,\BFx_i)$ and $h=\sum_{j=1}^m \gamma_j k(\cdot,\BFx_j')$ with $n,m\in\NatInt$,  $\{\BFx_i\}_{i=1}^n, \{\BFx_j'\}_{j=1}^m \subset\ScrX$ and $\{\beta_i\}_{i=1}^n, \{\gamma_j\}_{j=1}^m\subset\Real$, the inner product $\langle\cdot,\cdot\rangle_{\ScrF_k}$ is defined as
\[\langle g, h\rangle_{\ScrF_k}
= \sum_{i=1}^{n}\sum_{j=1}^{m} \beta_i\gamma_j k(\BFx_i,\BFx_j').
\]
Let $\|\cdot\|_{\ScrF_k}$ denote the norm of $\ScrF_k$, i.e., $\|g\|_{\ScrF_k} = \langle g, g\rangle_{\ScrF_k}$.
Then, the RKHS $\ScrH_k$ induced by $k$ is the closure of $\ScrF_k$ with respect to $\|\cdot\|_{\ScrF_k}$; that is,
\begin{align*}
\ScrH_k = \biggl\{
g=\sum_{i=1}^\infty \beta_i k(\cdot, \BFx_i): \  &{} \{\BFx_i\}_{i=1}^\infty \subset\ScrX, \ \{\beta_i\}_{i=1}^\infty \subset\Real  \mbox{ such that } \\
& \|g\|_{\ScrH_k}^2 \coloneqq \lim_{n\to\infty} \biggl\|\sum_{i=1}^n \beta_i k(\cdot,\BFx_i) \biggr\|_{\ScrF_k}^2 = \sum_{i,j=1}^\infty \beta_i\beta_j k(\BFx_i,\BFx_j) <\infty
\biggr\}.
\end{align*}

RKHSs include a variety of function spaces of interest, depending on the choice of the kernel. See \citet[Chapter 7]{BerlinetThomas-Agnan04}.
For example, if  $k$ is
the linear kernel, then $\ScrH_k$ is norm-equivalent to the space of all linear functions.
If $k$ is a Mat\'ern kernel with smooth parameter $\nu$, then $\ScrH_k$ is norm-equivalent to the Sobolev space of order $m=\nu+\frac{d}{2}$, which consists of functions that are weakly differentiable up to order~$m$ \citep{tuo2016theoretical}.

For a subset $\ScrX_0\subset \ScrX$,
we may define the restriction of $\ScrH_k$ on $\ScrX_0$ as
\[\ScrH_k(\ScrX_0) \coloneqq \left\{g:\ScrX\mapsto \Real: g = h|_{\ScrX_0} \mbox{ for some }h\in\ScrH_k \right\},\]
where $g=h|_\ScrX$ denotes $g(\BFx) = h(\BFx)$ for all $\BFx\in\ScrX_0$.
We equip $\ScrH_k(\ScrX_0)$ with norm
\[\|g\|_{\ScrH_k(\ScrX_0)} \coloneqq \inf_{\{h\in\ScrH_k: h|_{\ScrX_0} = g\}} \|h\|_{\ScrH_k}. \]
Then, $\ScrH_k(\ScrX_0)$ is a RKHS with norm $\|\cdot\|_{\ScrH_k(\ScrX_0)}$.
See \citet[page 351]{Aronszajn50}.
We may also write $\ScrH_k(\ScrX) = \ScrH_k$ if it is necessary to stress the dependence on $\ScrX$.

\subsection{Kernel Ridge Regression}

A main reason for the wide adoption of RKHSs in statistics and machine learning is the representer theorem \citep{ScholkopfHerbrichSmola01}.
It asserts that, given a set of training data---although seeking a function in a RKHS that best fits the data is an infinite-dimensional optimization problem---the optimal solution can be represented as a linear combination of a \emph{finite} number of functions.

\begin{lemma}[Representer Theorem]\label{lemma:representation-thm}
Suppose that $k:\ScrX\times\ScrX\mapsto \Real$ is a positive definite kernel,
$\ScrH_k$ is its associated RKHS,
$L:\Real \times \Real \mapsto \Real_+$ is an arbitrary loss function,
$\Omega:\Real_+ \mapsto \Real$ is a strictly increasing function,
and $\{(\BFx_i,y_i)\}_{i=1}^n$ is a given set of training data, where $\BFx_i\in\ScrX$ and $y_i\in\Real$.
Then, each optimal solution to the  optimization problem
\begin{equation}\label{eq:reg-risk-min}
    \min_{g\in\ScrH_k}\frac{1}{n}\sum_{i=1}^n L(y_i, g(\BFx_i)) + \Omega(\|g\|_{\ScrH_k})
\end{equation}
admits a representation of the form $g^*=\sum_{i=1}^n \beta_i^* k(\BFx_i,\cdot)$ for some constants $\beta_i^*\in\Real$, $i=1,\ldots,n$.
\end{lemma}

Suppose that a function $f$ is observed at $\BFx_i$, and the observation is $y_i = f(\BFx_i) + \varepsilon_i$, where $\varepsilon_i$ is the noise $i=1,\ldots,n$.
Kernel ridge regression (KRR) is also known as regularized least-squares.
It estimates $f$ in an RKHS by solving the optimization problem
\begin{equation}\label{eq:RLS}
\widehat{f}_{n,\lambda}\coloneqq \argmin_{g\in\ScrH_k}\frac{1}{n}\sum_{i=1}^n (y_i - g(\BFx_i))^2 + \lambda \|g\|_{\ScrH_k}^2,
\end{equation}
where $\lambda>0$ is a regularization parameter.
Clearly,
this is a special case of \eqref{eq:reg-risk-min}, by setting $L(y, y') = (y-y')^2$ and $\Omega(z) = \lambda z^2$.
By the representer theorem, the KRR estimator takes the form
$\widehat{f}_{n,\lambda}=\sum_{i=1}^n\beta_i^*k(\BFx_i,\cdot)$, and
$\BFbeta^*=(\beta_1^*,\ldots,\beta_n^*)^\intercal$ is given by
\begin{align*}
\BFbeta^*=\argmin_{\BFbeta\in\Real^n}\frac{1}{n}\sum_{i=1}^n\biggl(\sum_{j=1}^n\beta_i k(\BFx_i,\BFx_j) - y(\BFx_i)\biggr)^2+\lambda \sum_{i,j=1}^n\beta_i\beta_jk(\BFx_i,\BFx_j).
\end{align*}
This is a quadratic optimization problem.
A direct calculation yields $\BFbeta^* = (\BFK_n+n \lambda \BFI_n )^{-1}\BFy_n$.
Thus,
\begin{equation}\label{eq:KRR-estimator}
\widehat{f}_{n,\lambda}(\BFx) = \BFk_n^\intercal(\BFx) (\BFK_n+n \lambda \BFI_n )^{-1}\BFy_n,
\end{equation}
where  $\BFk_n(\BFx) = (k(\BFx_1,\BFx),\ldots, k(\BFx_n,\BFx))^\intercal$, $\BFK_n$ is the kernel matrix $(k(\BFx_i,\BFx_j))_{i,j=1}^n\in\Real^{n\times n}$, $\BFI_n $ is the $n\times n$ identity matrix, and $\BFy_n=(y_1,\ldots,y_n)^\intercal$.

\subsection{Kernel Interpolation for the Noise-free Case}
If the observations are noise-free---that is, if $y_i = f(\BFx_i)$, $i=1,\ldots,d$---then the regularized least-squares problem \eqref{eq:RLS} is reduced to
\[\min_{g\in\ScrH_k} \|g\|_{\ScrH_k} \qq{ subject to } g(\BFx_i) = f(\BFx_i),\  i=1,\ldots,n.\]
The solution is called the kernel interpolation (KI) estimator:
\begin{equation*}
    \label{eq:KI}
    \breve{f}_n(\BFx)\coloneqq \BFk_n^\intercal(\BFx)\BFK_n^{-1}\BFy_n,
\end{equation*}
which is identical to the KRR estimator \eqref{eq:KRR-estimator} with $\lambda=0$.

\section{Expected Improvement }\label{sec:EI}

Suppose that one is interested in designing a sequential sampling process for the purpose of maximizing a function $f$ of an unknown form.
EI is among the most popular BO algorithms for this task.
It first  assigns a GP prior on $f$.
Let $\mu(\BFx)$ and $k(\BFx,\BFx')$ denote the mean function and the kernel function of the prior GP, respectively.
Namely, for an arbitrary finite set of design points $\{\BFx_1,\ldots,\BFx_n\}\subset \ScrX$,
the prior distribution of $\{f(\BFx_1),\ldots,f(\BFx_n)\}$  is multivariate normal,  with a mean vector that is composed of $\mu(\BFx_i)$ and a covariance matrix that is composed of $k(\BFx_i,\BFx_j)$ for all $i,j=1,\ldots,n$.

Let $y_i=f(\BFx_i) + \varepsilon_i$ be a noisy observation of $f(\BFx_i)$.
If the noise terms are independent normal random variables with a \emph{known} variance $\varsigma^2$, then a straightforward use of Bayes' rule results in that for all $\BFx\in\ScrX$ the posterior distribution of $f(\BFx)$ is  normal, with a mean $\widetilde{f}_n(\BFx)$ and variance $s_n^2(\BFx)$ that are given by
\begin{align}
& f(\BFx)\,|\, y_1,\ldots,y_n\sim\mathsf{Normal}\big(\widetilde{f}_n(\BFx), s^2_n(\BFx)\big), \label{eq:posterior} \\
& \widetilde{f}_n(\BFx)\coloneqq \mu(\BFx) + \BFk_n^\intercal(\BFx)( \BFK_n+\varsigma^2 \BFI_n)^{-1} (\BFy_n-\BFmu_n), \label{eq:posterior-mean}\\
& s_n^2(\BFx) \coloneqq k(\BFx,\BFx)- \BFk_n^\intercal(\BFx)(\BFK_n + \varsigma^2 \BFI_n)^{-1}\BFk_n(\BFx), \label{eq:posterior-var}
\end{align}
where $\BFmu_n = (\mu(\BFx_1),\ldots, \mu(\BFx_n))^\intercal$.

Now, suppose that $n$ design points have been selected, and observations of $f$ at these design points have been collected.
The EI strategy to determine the next design point---for example, $x_{n+1}$---is based on the \emph{improvement function}, denoted by $I(\BFx,\xi)$, which measures the amount of improvement that a design point $\BFx$ would lead to relative to some threshold $\xi$:
\[
I(\BFx,\xi)\coloneqq   (f(\BFx)-\xi) \ind\{f(\BFx)>\xi\} =
\left\{
\begin{array}{ll}
     f(\BFx)-\xi, &\quad\mbox{if }f(\BFx) > \xi,  \\
     0, &\quad\mbox{otherwise}.
\end{array}
\right.
\]
Because the posterior distribution of $f(\BFx)$ is normal, as given by \eqref{eq:posterior}--\eqref{eq:posterior-var},
it is easy to show via direct calculation that
the expected improvement is
\[
\E[I(\BFx,\xi)\,|\, y_1,\ldots,y_n] = s_n(\BFx)\eta\biggl(\frac{\widetilde{f}_n(\BFx) -\xi }{s_n(\BFx)} \biggr),
\]
where $\eta:\Real\mapsto \Real$ is often called the \emph{information value function}, defined as
$\eta(z)\coloneqq z\Phi(z)+\phi(z)$,
with  $\Phi$ and $\phi$ denoting the cumulative distribution function and the probability density function of the standard normal distribution, respectively.

A typical choice of $\xi$ is to adaptively set it to be $\max_{1\leq i\leq } \widetilde{f}_n(\BFx_i)$, the maximum of the posterior mean at the design points that have been selected so far.
The EI strategy determines the next design point $\BFx_{n+1}$ via maximizing the expected improvement:
\begin{equation*}\label{eq:EIClosedForm}
\BFx_{n+1} = \argmax_{\BFx\in \ScrX}\biggl\{ s_n(\BFx)\eta\biggl(\frac{\widetilde{f}_n(\BFx) -  \max_{1\leq i\leq n} \widetilde{f}_n(\BFx_i)}{s_n(\BFx)} \biggr) \biggr\}.
\end{equation*}
Despite its closed form,
the objective function of the above optimization problem is non-convex.
Thus, it may be  computationally challenging to solve it numerically  when the feasible set $\ScrX$ is high-dimensional.

\end{APPENDICES}

% Acknowledgments here
% \ACKNOWLEDGMENT{}

% References here (outcomment the appropriate case)

% CASE 1: BiBTeX used to constantly update the references
%   (while the paper is being written).
\bibliographystyle{informs2014} % outcomment this and next line in Case 1
\bibliography{High-dim-SO} % if more than one, comma separated

% CASE 2: BiBTeX used to generate mypaper.bbl (to be further fine tuned)
%\input{mypaper.bbl} % outcomment this line in Case 2

%If you don't use BiBTex, you can manually itemize references as shown below.

%% Here starts the e-companion (EC)
%%%%%%%%%%%%%%%%%%%%%%%%%%%%%%%%%%%%%%%%%%%%%%%%%%%%%%%%%%
 \ECSwitch

%\ECDisclaimer
%%%%%%%%%%%%%%%%%%%%%%%%%%%%%%%%%%%%%%%%%%%%%%%%%%%%%%%%%%

%%% Main head for the e-companion
\EquationsNumberedBySection
\ECHead{Supplemental Material}

To make this supplemental material self-contained, we repeat below  the main assumptions in the main body of the paper.

\begin{repeatassumption}[Assumption~\ref{assump:optimum}]
$\ScrX = (0,1)^d$ and $f$ has a global maximum $\BFx^*\in\ScrX$.
\end{repeatassumption}

\begin{repeatassumption}[Assumption~\ref{assump:sub-Gaussian}]
For any $n\in\NatInt$ and any sequence of design points $\{\BFx_{i}\}_{i=1}^n\subset\ScrX$, the noise terms $\{\varepsilon(\BFx_i)\}_{i=1}^n$ are independent zero-mean sub-Gaussian random variables with variance proxy $\sigma^2$, denoted by $\subG(\sigma^2)$.
That is, $\E\bigl[e^{t\varepsilon(\BFx_i)} \bigr] \leq e^{t^2\sigma^2/2}$ for all $t\in\Real$ and $i=1,\ldots,n$.
\end{repeatassumption}

\section{Generalization From Brownian Field to Tensor Markov}

In this supplemental material, we generalize Algorithm~\ref{alg:EI} and the theoretical results regarding its convergence rate to the class of tensor Markov kernels, which include BF kernels as a special case.

\begin{definition}[Tensor Markov Kernel]\label{def:TMK}
For each $j=1,\ldots,d$, let $\ScrI_j\subseteq \Real$ be an interval (open or closed), and let
$p_j$ and $q_j$ be positive functions on $\ScrI_j$ with $p_j/q_j$ strictly increasing.
Then,
\begin{equation*}\label{eq:TMK}
    k(\BFx,\BFx') = \prod_{j=1}^d p_j(x_j\wedge x'_j) q_j(x_j\vee x'_j),
\end{equation*}
is  a  tensor Markov (TM) kernel on $\ScrI = \bigtimes_{j=1}^d \ScrI_j$.
\end{definition}

We first show that BF kernels satisfy the following assumption,
which provides regularity conditions for TM kernels.
These conditions allow us to establish convergence rates of the generalized version of Algorithm~\ref{alg:EI}, in which the BF kernel is replaced with a TM kernel.

\subsection{Assumption on Tensor Markov Kernels}

\begin{assumption}\label{assump:SL}
For each $j=1,\ldots,d$, let $\ScrI_j=(a_j,b_j)\subseteq \Real$ with  $\infty\leq a_j\leq 0$ and $1\leq b_j\leq\infty$ and let
$p_j$ and $q_j$ be positive functions on $\ScrI_j$ with $p_j/q_j$ strictly increasing on $\ScrI_j$.
Let $k(\BFx,\BFx') = \prod_{j=1}^d k_j(x_j, x_j')$ be a TM kernel defined on $\ScrI \coloneqq \bigtimes_{j=1}^d \ScrI_j$, where $k_j(x,x') \coloneqq p_j(x_j\wedge x'_j) q_j(x_j\vee x'_j)$.
\begin{enumerate}[label=(\roman*)]
\item \label{condi:SL-operator}
    For each $j=1,\ldots,d$, both $p_j$ and $q_j$ are continuously differentiable and satisfy the following differential equation:
\begin{equation}
    \label{eq:SL_equation}
    \CalL_j[g](x)\coloneqq -\frac{\partial}{\partial x}\left(u_j(x) \frac{\partial g}{\partial x} \right) + v_j(x) g(x) = 0,\quad x\in (a_j, b_j),
\end{equation}
where $u_j$ is a continuously differentiable function on $(a_j,b_j)$ and $v_j$ is a continuous function on $(a_j,b_j)$.
\item \label{condi:self-ad}
For each $j=1,\ldots,d$, let $\ScrH_{k_j}$ the RKHS induced by $k_j$. For all  $g,h\in\ScrH_{k_j}$,
\[\langle g, h\rangle_{\ScrH_{k_j}} =  \int_{a_j}^{b_j}g(x)\CalL_j[h](x)\dd{x}=\int_{a_j}^{b_j}\CalL_j[g](x)h(x)\dd{x}
= \int_{a_j}^{b_j}\CalL_j^{\frac{1}{2}}[g](x)\CalL_j^{\frac{1}{2}}[h](x)\dd{x}.\]
\item \label{condi:embed}
For each $j=1,\ldots,d$, let $\ScrH_{k_j}(0,1)$ denote the restriction of $\ScrH_{k_j}$ on $(0,1)$ and let $\|\cdot\|_{\ScrH_{k_j}(0,1)}$ denote the RKHS norm of the restriction, that is,
\[\|g\|_{\ScrH_{k_j}(0,1)} \coloneqq \inf_{\{h\in\ScrH_{k_j}: h|_{(0,1)} = g\}} \|h\|_{\ScrH_{k_j}}. \]
There exist  positive constants $C$ and $\tilde{C}$ such that for all $g:(0,1)\mapsto\Real$,
\begin{equation}\label{eq:norm-equiv}
    C\int_{0}^{1}\bigl|D^1g(x)\bigr|^2\dd{x}\leq \|g\|_{\ScrH_{k_j}(0,1)}^2\leq \tilde{C}\int_{0}^{1}\bigl|D^1g(x)\bigr|^2\dd{x},
\end{equation}
where $D^1 g$ denotes the first-order weak derivative of $g$;
moreover, for all $w\in L^2(a_j,b_j)$,
\begin{equation}\label{eq:norm-equiv-2}
    \int_0^1\biggl|\frac{\partial^2}{\partial x^2}\int_{a_j}^{b_j}k_j(x,s)w(s)\dd{s}\bigg|^2\dd{x}<\infty,
\end{equation}
where the partial derivative is understood in the weak sense.
\end{enumerate}
\end{assumption}

To show that  BF kernels satisfy Assumption~\ref{assump:SL}, we need the  Gagliardo–Nirenberg interpolation inequality for estimating the weak derivatives of a function; see, e.g., \citet[Proposition~9]{Haroske17_ec}.
This inequality will also play a vital role in subsequent analysis.

\begin{lemma}[Gagliardo–Nirenberg Interpolation Inequality]
\label{lemma:Gagliardo–Nirenberg}
Let $\ScrI = \bigtimes_{j=1}^d (a_j, b_j)$ where $-\infty\leq a_j < b_j\leq \infty$ for all $j=1,\ldots,d$.
Let $m\geq 0$ and $n\geq 1$ be integers,
$1\leq q,r\leq\infty$, and $ \frac{m}{n}\leq \alpha \leq 1$
such that
\[\frac{1}{p}=\frac{m}{d}+\left(\frac{1}{r}-\frac{n}{d}\right)\alpha+\frac{1-\alpha}{q}.\]
Then, for all $g:\ScrI\to\Real$,
\[\|D^m g\|_{p}\leq C\|D^n g\|^{\alpha}_{r}\|g\|^{1-\alpha}_{q},\]
for some positive constant $C$,
where $D^s g$ denotes an order-$s$ weak partial derivative of $g$ for an integer $s\geq 0$,
and $\|g\|_t$ denotes the $L^t$ norm of $g$ for a real number $1\leq t\leq \infty$.
\end{lemma}

\begin{corollary}
\label{cor:Gagliardo–Nirenberg}
Let $\ScrI = \bigtimes_{j=1}^d (a_j, b_j)$ where $-\infty\leq a_j < b_j\leq \infty$ for all $j=1,\ldots,d$.
Then, there exists a positive constant $C$ such that for all $g:\ScrI\to\Real$,
\[\| g\|_\infty\leq C\sqrt{\|D^d g\|_{2}\|g\|_2}.\]
\end{corollary}
\proof{Proof.}
In Lemma~\ref{lemma:Gagliardo–Nirenberg}, we let $m=0$, $n=d$, $r=q=2$ and $\alpha=\frac{1}{2}$.
\Halmos\endproof

\begin{corollary}
\label{cor:Gaglirdo-Nirenberg-cor2}
Let $d=1$ and $\ScrI=(a,b)$ where $-\infty< a < b < \infty$.
Let $0\leq m< n$ be integers.
Then, there exists a positive constant $C$  such that for  all  $g:\ScrI\to\Real$,
\[\|D^m g\|_2\leq C\|D^n g\|_2.\]
\end{corollary}
\proof{Proof.}
Note that
\begin{align*}
    \|D^m g \|_2
= \int_a^b \abs{D^m g(x)} \dd{x}
\leq \int_a^b \| D^m g\|_\infty \dd{x}
={}&  (b-a)  \| D^m g\|_\infty \\
\leq{}& (b-a) C \sqrt{\| D^{m+1} g\|_2 \|  D^m g\|_2},
\end{align*}
for some constant $C>0$,
where the second inequality follows from
applying Corollary~\ref{lemma:Gagliardo–Nirenberg}.
Thus,
\begin{align*}
\|D^m g \|_2 \leq C^2 (b-a)^2 \| D^{m+1} g\|_2.
\end{align*}
The proof is completed by performing induction on $m$.
\Halmos\endproof

We are now ready to prove that BF kernels satisfy Assumption~\ref{assump:SL}.

\begin{proposition}
\label{prop:BM-kernel-SL-operator}
Let $k(\BFx,\BFx') = \prod_{j=1}^d [\theta_j+\gamma_j(x_j\wedge x_j')]$ be a BF kernel,
where $\theta_j$ and $\gamma_j$ are positive constants for all $j=1,\ldots,d$.
Then, $k$ satisfies Assumption~\ref{assump:SL}.
\end{proposition}

\proof{Proof.}
Without loss of generality, we assume $\gamma_j=\theta_j=1$ for all $j=1,\ldots,d$.
For each $j$,
let  $a_j=-1$, $b_j=\infty$,  $p_j(x)=1+x$, and  $q_j(x)=1$; moreover, define a differential operator $\CalL_j$ as
\[\CalL_j[g](x)=-\frac{\partial^2}{\partial x^2}g(x)=-D^2g(x).\]

\textbf{Condition~\ref{condi:SL-operator}.}
It follows from a direct calculation that
\[\CalL_j[p_j](x)=\CalL_j[q_j](x)=0,\quad  x\in(0,1),\]
for all $j=1,\ldots,d$.
Thus, $k$ satisfies
Condition~\ref{condi:SL-operator}.

\textbf{Condition~\ref{condi:self-ad}.}
Fix $j=1,\ldots,d$.
We first show that
$g(-1) = 0$ for all $g\in\ScrH_{k_j}$.
To see this, note that $k_j(x,x)\geq 0$  if and only if $x\geq 1$,
so
the maximum domain on which $k_j$ is a positive definite kernel is $[-1,\infty)$.
Moreover, given any $n\in\NatInt$, $\{\beta_i\}_{i=1}^n\subset\Real$, and $\{x_i\}_{i=1}^n\subset[-1,\infty)$, we have
\[\sum_{i=1}^n \beta_i k_j(-1,x_i)=\sum_{i=1}^n \beta_i \big[1+(-1)\wedge x_j\big]=0.\]
Therefore, by the constructive definition of RKHSs in Appendix~\ref{sec:RKHS}, $g(-1)=0$ for all $g\in\ScrH_{k_j}$.

Then, through integration-by-parts, we have that for all $g, h\in\ScrH_{k_j}$,
\begin{align*}
    -\int_{-1}^\infty D^2 g(x)h(x)\dd{x}
    ={} &\int_{-1}^\infty D^1g(x)D^1h(x)\dd{x}-\big[D^1g(-1) h(-1) - D^1g(\infty) h(\infty)\big]\\
    ={}&\int_{-1}^\infty D^1g(x) D^1h(x)\dd{x}
    = -\int_{-1}^\infty g(x)D^2h(x)\dd{x},
\end{align*}
where the second equality follows from the fact that $h(-1)=0$ and $D^1g(\infty)=0$, and the last equality can be derived from exactly the same way.
So we have verified that all $g, h\in\ScrH_{k_j}$,
\[
\int_{-1}^\infty \CalL_j[g](x)h(x)\dd{x} =
\int_{-1}^\infty \CalL_j^{\frac{1}{2}}[g](x) \CalL_j^{\frac{1}{2}}[h](x)\dd{x}
= \int_{-1}^\infty g(x)\CalL_j[h](x)h(x)\dd{x}.
\]

We may define an inner product via $\CalL_j$ as $\langle g, h \rangle_{\CalL_j} \coloneqq \int_{-1}^\infty \CalL_j[g](x)h(x)\dd{x}$.
Then,
\begin{align*}
\langle g, k_j(x',\cdot) \rangle_{\CalL_j} ={}&   -\int_{-1}^\infty D^2g(x) k_j(x',x)\dd{x}\\
={} & \int_{-1}^{x'} D^1g(x) \dd{x}- \big[D^1g(-1)k_j(x', -1)-D^1g(\infty)k_j(x',\infty)\big] \\
    ={}& g(x').
\end{align*}
Namely, the reproducing property of $k_j$ holds under $\langle \cdot, \cdot \rangle_{\CalL_j}$.
Hence,  the two inner products $\langle \cdot, \cdot \rangle_{\CalL_j}$ and $\langle \cdot, \cdot \rangle_{\ScrH_{k_j}}$ are identical.
Thus, $k$ satisfies Condition~\ref{condi:self-ad}.

\textbf{Condition~\ref{condi:embed}.}
Fix $j=1,\ldots,d$.
Let $g\in\ScrH_{k_j}(\ScrX)$.
Then, for any $h\in\ScrH_{k_j}$ such that $g = h|_{(0,1)}$,
\begin{align}\label{eq:integral_bounded}
    \|h\|_{\ScrH_{k_j}}^2={} &\int_{-1}^{\infty} \abs{D^1 h(x)}^2\dd{x}
    \geq  \int_{0}^{1} \big|D^1 h(x)\big|^2\dd{x}
    = \int_{0}^{1} \big|D^1 g(x)\big|^2\dd{x},
\end{align}
where the first  equality follows from Condition~\ref{condi:embed}.
Hence,
\begin{equation}\label{eq:norm-equiv-first-half}
\|g\|_{\ScrH_{k_j}(0,1)}^2 =
\inf_{\{h\in\ScrH_{k_j}: h|_{(0,1)} = g\}}
\|h\|_{\ScrH_{k_j}}^2 \geq
\int_{0}^{1} \big|D^1g(x)\big|^2\dd{x},
\end{equation}
which implies
\begin{equation}\label{eq:norm-equiv-first-half'}
\ScrH_{k_j}(0,1) \subseteq \left\{ g:(0,1)\mapsto\Real: \int_{0}^{1} \big|D^1g(x)\big|^2\dd{x} < \infty \right\}.
\end{equation}

On the other hand, let $g:(0,1)\mapsto\Real$ such that $\int_0^1 \abs{D^1 g(x)}^2\dd{x} < \infty$.
Then, $g$ is in the (classical) order-1 Sobolev space of functions with domain $(0,1)$.
Hence, we can apply Theorem~1 on page 268 of \cite{Evans10_ec} to extend $g$ from $(0,1)$ to  a bounded open interval---for example, $(-1,2)$---in the following way.
With $\CalE$ denoting the extension operator, we have
$\CalE[g](x)=g(x)$ for $x\in(0,1)$,
$\CalE[g](x)=0$ for $x\notin (-1,2)$, and
\begin{equation}\label{eq:fun-ext}
    \int_{-1}^2\big|\CalE[g](x)\big|^2+\big|D^1 \CalE[g](x)\big|^2\dd{x}\leq C_1 \int_{0}^1\big|g(x)\big|^2+\big| D^1 g(x)\big|^2\dd{x},
\end{equation}
for  some constant $C_1$  independent of $g$.
Note that
\begin{align*}
    \bigl\| \CalE[g]\big\|_{\ScrH_{k_j}}^2 = \int_{-1}^\infty \abs{D^1 \CalE[g](x)}^2\dd{x}
    ={}& \int_{-1}^2 \abs{D^1 \CalE[g](x)}^2\dd{x} \nonumber \\
    \leq{}& C_1 \int_{0}^1\big|g(x)\big|^2+\big| D^1 g(x)\big|^2\dd{x} \nonumber \\
    \leq{}& C_1 \int_{0}^1 C_2\big|D^1 g(x)\big|^2+ \big| D^1 g(x)\big|^2\dd{x} \nonumber \\
    ={}& C_1(C_2 +1 ) \int_{0}^1 \big| D^1 g(x)\big|^2\dd{x} < \infty,
\end{align*}
for some constant $C_2$ independent of $g$,
where the first inequality follows from \eqref{eq:fun-ext} and
the second inequality from Corollary~\ref{cor:Gaglirdo-Nirenberg-cor2}.
It follows that $\CalE[g] \in \ScrH_{k_j}$, and thus
\begin{align}
    \| g\|^2_{\ScrH_{k_j}(0,1)} ={}&
\inf_{\{h\in\ScrH_{k_j}: h|_{(0,1)} = g\}}
\|h\|_{\ScrH_{k_j}}^2
\leq  \bigl\| \CalE[g]\big\|^2_{\ScrH_{k_j}}
\leq \tilde{C} \int_{0}^1 \big| D^1 g(x)\big|^2\dd{x},  \label{eq:norm-equiv-second-half}
\end{align}
where $\tilde{C} =  C_1(C_2 +1 ) $.
This implies
\begin{equation}\label{eq:norm-equiv-second-half'}
\left\{ g:(0,1)\mapsto\Real: \int_{0}^{1} \big|D^1g(x)\big|^2\dd{x} < \infty \right\} \subseteq  \ScrH_{k_j}(0,1).
\end{equation}

Now, combining \eqref{eq:norm-equiv-first-half'} and \eqref{eq:norm-equiv-second-half'}, we conclude that \[
\ScrH_{k_j}(0,1) = \left\{ g:(0,1)\mapsto\Real: \int_{0}^{1} \big|D^1g(x)\big|^2\dd{x} < \infty \right\}.
\]
Hence, for a function $g:(0,1)\mapsto\Real$,
if $g\in \ScrH_{k_j}(0,1)$,  then
\eqref{eq:norm-equiv} holds because of \eqref{eq:norm-equiv-first-half'} and \eqref{eq:norm-equiv-second-half'};
but if $g\notin \ScrH_{k_j}(0,1)$, then $\|g\|_{\ScrH_{k_j}}(0,1)=\int_{0}^{1} \big|D^1g(x)\big|^2\dd{x} = \infty$, so \eqref{eq:norm-equiv} holds trivially.

Lastly, we verify \eqref{eq:norm-equiv-2}.
Note that for any square-integrable function $w$ on $(-1,\infty)$,
\begin{align*}
    \int_{0}^1\biggl|\frac{\partial^2}{\partial x^2}\int_{-1}^{\infty}k_j(x,s)w(s)\dd{s}\biggr|^2\dd{x}={}& \int_{0}^1\biggl|\frac{\partial^2}{\partial x^2}\int_{-1}^{\infty}\bigl(1+x\wedge s\bigr)w(s)\dd{s}\biggr|^2\dd{x}\\
    ={}& \int_{0}^1\biggl|\frac{\partial}{\partial x}\int_{x}^{\infty}w(s)\dd{s}\biggr|^2\dd{x}\\
    ={}& \int_{0}^1\bigl|w(x)\bigr|^2\dd{x}<\infty.
\end{align*}
Therefore, $k$ satisfies Condition~\ref{condi:embed}.
\Halmos\endproof

\begin{remark}
Laplace kernels of form $k(\BFx,\BFx') = \exp\bigl(-\sum_{j=1}^d\theta_j\abs{x_j-y_j}\bigr)$ for some constants $\{\theta_j\}_{j=1}^d$
are also TM kernels.
One may follow a proof similar to that of Proposition~\ref{prop:BM-kernel-SL-operator} to show that Laplace kernels also satisfy Assumption~\ref{assump:SL}.
\end{remark}

\subsection{Equivalence Between Function Spaces}

Now we can generalize Proposition~\ref{prop:RKHS-norm-equivalence}
to the norm-equivalent between $\ScrH_{\mathsf{mix}}^1$ and $\ScrH_{k}(\ScrX)$, the restriction of $\ScrH_{k}$ on $\ScrX=(0,1)^d$, for any kernel $k$ satisfying Assumption~\ref{assump:SL}.
\begin{proposition}
\label{prop:SL-norm-equivalence}
Let $k$ be a kernel satisfying Assumption~\ref{assump:SL}.
Then,
for a function $g:\ScrX\mapsto\Real$, $g\in \ScrH_{k}(\ScrX)$ if and only if $g\in \ScrH_{\mathsf{mix}}^1$;
moreover,
there exist positive constants $C$ and $\tilde{C}$ such that for all $g:\ScrX\mapsto\Real$,
\[C \|g\|_{\ScrH_{\mathsf{mix}}^1} \leq \|g\|_{\ScrH_k(\ScrX)}\leq \tilde{C}\|g\|_{\ScrH_{\mathsf{mix}}^1}.\]
\end{proposition}
\proof{Proof.}

It follows from the tensor product form of $k$ and
Condition~\ref{condi:self-ad} in Assumption~\ref{assump:SL} that
\begin{align*}
    \|h\|_{\ScrH_{k}}^2   =  \langle h, h \rangle_{\ScrH_k} = {}& \int_{a_d}^{b_d}\CalL_{d}\cdots \int_{a_2}^{b_2}\CalL_{2}\int_{a_1}^{b_1}\CalL_{1}[h] h\dd{x_1}\dd{x_2}\cdots\dd{x_d} \\
    ={}& \int_{\ScrI}\biggl|\prod_{j=1}^d\CalL^{\frac{1}{2}}_{j}[h] (\BFx)\biggr|^2\dd{\BFx}.
\end{align*}
for all $h\in\ScrH_k$,
where $\CalL_j$, as an operator defined on one-dimensional functions, takes effect with respect to $x_j$.
Then, for any $g:\ScrX\mapsto\Real$, with repeated use of both Fubini's theorem and \eqref{eq:norm-equiv} in Assumption~\ref{assump:SL}, we deduce that
\begin{align*}
    \|g\|_{\ScrH_k(\ScrX)}^2 ={}& \inf_{h\in\ScrH_k:h|_{\ScrX}=g} \| h\|_{\ScrH_k}\\
    ={} & \inf_{h\in\ScrH_k:h|_{\ScrX}=g} \int_{\ScrI}\biggl|\prod_{j=1}^d\CalL^{\frac{1}{2}}_{j}[h] (\BFx)\biggr|^2\dd{\BFx}\\
    ={} & \inf_{h\in\ScrH_k:h|_{\ScrX}=g} \int_{a_d}^{b_d}\cdots\int_{a_2}^{b_2}\int_{a_1}^{b_1}\biggl|\CalL_{1}^{\frac{1}{2}}\biggl[\prod_{j=2}^d\CalL^{\frac{1}{2}}_{j}[h](x_2,\ldots,x_d)\biggr](x_1)\biggr|^2\dd{x_1}\dd{x_2}\cdots\dd{x_d}\\
    \geq{} & \inf_{h\in\ScrH_k:h|_{\ScrX}=g} C_1\int_{a_d}^{b_d}\cdots\int_{a_2}^{b_2}\int_{0}^{1}\biggl|D^1_{x_1}\biggl[\prod_{j=2}^d\CalL^{\frac{1}{2}}_{j}[h](x_2,\ldots,x_d)\biggr]\biggr|^2\dd{x_1}\dd{x_2}\cdots\dd{x_d}\\
    ={} & \inf_{h\in\ScrH_k:h|_{\ScrX}=g} C_1\int_{0}^{1}\int_{a_d}^{b_d}\cdots\int_{a_2}^{b_2}\biggl|\CalL^{\frac{1}{2}}_{2}\biggl[\prod_{j=3}^d\CalL^{\frac{1}{2}}_{j}[D^1_{x_1}h](x_3,\ldots,x_d)\biggr](x_2)\biggr|^2\dd{x_2}\cdots\dd{x_d}\dd{x_1}\\
    \geq{} & \inf_{h\in\ScrH_k:h|_{\ScrX}=g} C_1C_2\int_{0}^{1}\int_{a_d}^{b_d}\cdots\int_{0}^{1}\biggl|D^1_{x_2}\biggl[\prod_{j=3}^d\CalL^{\frac{1}{2}}_{j}[D^1_{x_1}h](x_3,\ldots,x_d)\biggr]\biggr|^2\dd{x_2}\cdots\dd{x_d}\dd{x_1}\\
    \cdots{}& \\
    \geq{} & \biggl(\prod_{j=1}^dC_j\biggr)\int_{\ScrX}\big|D^1_{x_1}\cdots D^1_{x_d}g(\BFx)\big|^2\dd{\BFx} =  C \|g \|^2_{\ScrH_{\mathsf{mix}}^1},
\end{align*}
where $C= \prod_{j=1}^dC_j$ and $\{C_j\}_{j=1}^d$ are constants independent of $g$ from the repeated use of the first inequality of \eqref{eq:norm-equiv}.

Likewise, we can show
$\|g\|_{\ScrH_k(\ScrX)}^2 \leq \tilde{C} \|g \|^2_{\ScrH_{\mathsf{mix}}^1}$,
for some constant $\tilde{C}$ independent of $g$.

The equivalence between $\|\cdot\|_{\ScrH_k(\ScrX)} $ and $ \|\cdot \|_{\ScrH_{\mathsf{mix}}^1}$ immediately implies that $\ScrH_k(\ScrX) = \ScrH_{\mathsf{mix}}^1$ as a set of functions.
\Halmos\endproof

In what follows,
we define a subspace of $\ScrH_k$ that consists of functions of a higher-order smoothness,
and show that its restriction on $\ScrX$ is identical to $\ScrH_{\mathsf{mix}}^2$ as a set of functions.

\begin{definition}
Let $k$ be a kernel satisfying Assumption~\ref{assump:SL}. Define
\begin{align*}
    \ScrH_k^2 \coloneqq \left\{g\in\ScrH_k: \ \mbox{There exists }  w\in L^2(\ScrI) \mbox{ such that } g=\int_{\ScrI}k(\cdot,\BFs)w(\BFs)\dd{\BFs}\right\},
\end{align*}
where $\ScrI = \prod_{j=1}^d (a_j,b_j)$ as in Assumption~\ref{assump:SL}.
Also define the restriction of $\ScrH_k^2$ on $\ScrX$ as
\[\ScrH_k^2(\ScrX) \coloneqq  \left\{g:\ScrX\to\Real:\ \mbox{There exists } h\in\ScrH_k^2 \mbox{ such that } h|_{\ScrX}=g\right \}.\]
\end{definition}

\begin{lemma}
\label{lemma:second-order-identity}
For a function $g:\ScrI\to \Real$, $g\in\ScrH_{k}^2$ if and only if
\[\int_{\ScrI}\biggl|\prod_{j=1}^d\CalL_{j}[g](\BFx)\biggr|^2\dd{\BFx}<\infty.\]
\end{lemma}
\proof{Proof.} Suppose $g\in\ScrH^2_k$. Then,  $g=\int_{\ScrI}k(\cdot,\BFx)w(\BFx)\dd{\BFx}$ on $\ScrX$ for some $w\in L^2(\ScrI)$.  Note that by Condition~\ref{condi:self-ad} in Assumption~\ref{assump:SL}, $\CalL_j$ is a self-adjoint operator for each $j$, and it can be easily checked that $\prod_{j=1}^dk_j$ is the Green's function for $\prod_{j=1}^d\CalL_j$. Hence:
\[\int_{\ScrI}\biggl|\prod_{j=1}^d\CalL_j[g](\BFx)\biggr|^2\dd{\BFx}=\int_\ScrI \bigl|{w}(\BFx)\bigr|^2\dd{\BFx}<\infty.\]
Suppose $\int_{\ScrI}\bigl|\prod_{j=1}^d\CalL_j[g](\BFx)\bigr|^2 \dd{\BFx}<\infty$.
Let $w=\prod_{j=1}^d\CalL_j[g](\cdot)$ and $w\in L^2(\ScrI)$.
Then we can apply the Green's function property of $\prod_{j=1}^dk_j$ to get
\[\int_{\ScrI}k(\BFx,\BFs)w(\BFx)\dd{\BFx}=\int_{\ScrI}k(\BFx,\BFs)\prod_{j=1}^d\CalL_j[g](\BFs)\dd{\BFx}=g(\BFx).\quad \Box\]

\begin{proposition}
\label{prop:H-2-mix}
Let $k$ be a kernel satisfying Assumption~\ref{assump:SL}.
Then,
for a function $g:\ScrX\mapsto \Real$,
$g\in \ScrH^2_k(\ScrX)$ if and only if $g\in \ScrH_{\mathsf{mix}}^2$.
\end{proposition}
\proof{Proof.}
Suppose $g\in\ScrH^2_k(\ScrX)$. Then,  $g=\int_{\ScrI}k(\cdot,\BFx)w(\BFx)\dd{\BFx}$ on $\ScrX$ for some $w\in L^2(\ScrI)$.  Hence,
\begin{align*}
     \|g\|^2_{\ScrH_{\mathrm{mix}}^2} = \biggl\|\frac{\partial^2}{\partial x_1^2}\frac{\partial^2}{\partial x_2^2} \cdots \frac{\partial^2}{\partial x_d^2} g\biggr\|_2^2={} & \int_\ScrX \biggl|\frac{\partial^2}{\partial x_1^2}\frac{\partial^2}{\partial x_2^2} \cdots \frac{\partial^2}{\partial x_d^2}\int_{\ScrI}k(\BFx,\BFs)w(\BFs)\dd{\BFs}\biggr|^2\dd{\BFx}\\
     ={} & \int_\ScrX \biggl|\int_{\ScrI}w(\BFs)\prod_{j=1}^d  \frac{\partial^2}{\partial x_j^2} k_j(x_j,s_j)\dd{\BFs}\bigg|^2\dd{\BFx}<\infty,
\end{align*}
where the finiteness follows from Condition~\ref{condi:embed} in Assumptions~\ref{assump:SL}.
Therefore, $g\in\ScrH^2_{\mathsf{mix}}$.

Conversely,
suppose $g\in \ScrH^2_{\mathsf{mix}}$.
We first prove
\begin{equation}
\label{eq:finite-k-2}
    \int_{\ScrX}\biggl|\prod_{j=1}^d\CalL_j[g](\BFx)\biggr|^2\dd{\BFx}<\infty
\end{equation} by induction in the dimensionality $d$.
To stress the dependence on $d$, we write  $\ScrH^2_{k,d} = \ScrH^2_k$.
Let $w=\prod_{j=1}^d\CalL_{j}[g]$ on $\ScrX$ and $w=0$ outside $\ScrX$.

When $d=1$, we have
\begin{align}
    \int_{0}^{1}\bigl|w(x)\bigr|^2\dd{x}
    ={} & \int_{0}^{1}\biggl|\biggl(\frac{\partial}{\partial x}u_1(x)\frac{\partial}{\partial x}+v_1(x)\biggr)[g]\biggr|^2\dd{x} \nonumber \\
    ={} &  \int_{0}^{1}\bigl|D^1 u_1(x) D^1 g(x)+u_1(x)D^2g(x)+v_1(x)g(x)\bigr|^2\dd{x}\\
    \leq{} &\sum_{i_1=0}^2\sum_{i_2=i_1}^2C_{i_1,i_2}\int_{0}^{1}\bigl| D^{i_1}g(x) D^{i_2}g(x)\bigr|\dd{x}, \label{eq:w-finite-1}
\end{align}
for some positive constants $C_{i_1,i_2}$,
where the inequality follows from the fact that $u_1$ is continuously differentiable and $v_1$ is continuous by Assumption~\ref{assump:SL}.

For all $0\leq i_1\leq i_2 \leq 2$,
we may first apply the Cauchy–Schwarz inequality and then apply Corollary~\ref{cor:Gaglirdo-Nirenberg-cor2} to deduce that
\begin{align}
\left(\int_{0}^{1}\bigl|D^{i_1}g(x) D^{i_2}g(x)\bigr|\dd{x}\right)^2
\leq{}& \int_{0}^{1}\bigl|D^{i_1}g(x) \bigr|^2\dd{x}\int_{0}^{1}\bigl|D^{i_2}g(x)\bigr|^2\dd{x} \nonumber \\
\leq{}& C_1 \left(  \int_{0}^{1}\bigl|D^{2}g(x) \bigr|^2\dd{x} \right)^2 < \infty,\label{eq:w-finite-2}
\end{align}
for some positive constant $C_1$,
where the finiteness follows from the assumption that $g\in \ScrH_{\mathsf{mix}}^2$.
Hence, we know from \eqref{eq:w-finite-1} and \eqref{eq:w-finite-2} that $w$ is square-integrable, i.e.,
$w\in L^2(0,1)$.
 We then can complete the case of $d=1$,
\[\int_{0}^1\bigl|w(x)\bigr|^2\dd{x}\leq C_2\int_{0}
^1\bigl|D^{2}g(x) \bigr|^2\dd{x}<\infty\]
for some universal constant $C_2$ independent of $g$.

Let $\BFx_{-d}$ denote the vector $(x_1,\ldots,x_{d-1})$. Now suppose for any $(d-1)$-dimensional function $g:\ScrX\to\Real$, the following inequality is satisfied:
\[\int_{\ScrX}\bigl|w(\BFx_{-d})\bigr|^2\dd{\BFx_{-d}}\leq C_3 \int_{\ScrX}\biggl|\frac{\partial^{2(d-1)}g}{\partial x_1^2\cdots\partial x_{d-1}^2}\biggr|^2\dd{\BFx_{-d}}\]
where $C_3$ is some constant independent of $g$. Then for any $d$-dimensional function $g\in\ScrH_{\mathsf{mix}}^2$,  one can apply Fubini's theorem to check
\begin{align*}
    \int_{\ScrX}\bigl|w(\BFx)\bigr|^2\dd{\BFx}\leq{} & C_2\int_{(0,1)^{d-1}}\int_{0}^1 \biggl|\frac{\partial^2}{\partial x_d^2} \prod_{j=1}^{d-1}\CalL_j[g](\BFx)\biggr|^2\dd{x_d}\dd{\BFx_{-d}} \\
    ={} &C_2\int_{0}^1\int_{(0,1)^{d-1}} \biggl| \prod_{j=1}^{d-1}\CalL_j\biggl[\frac{\partial^2 g}{\partial x_d^2}\biggr](\BFx_{d-1})\biggr|^2\dd{\BFx_{-d}}\dd{x_d}\\
    \leq{} & C_2C_3 \int_{\ScrX}\biggl|\frac{\partial^{2d}g(\BFx)}{\partial x_1^2\cdots\partial x_{d}^2}\biggr|^2\dd{\BFx}<\infty.
\end{align*}
where the first line is from the base case and the last line is from the induction assumption. This finishes the proof for equation~\eqref{eq:finite-k-2}.

Then for any function $g\in\ScrH^2_{\mathsf{mix}}$ we construct a function $h\in\ScrH^2_k$ satisfying $h=g$ on $\ScrX$ to complete th proof. we let $w=\prod_{j=1}^d\CalL_j[g]$ on $\ScrX$ and $w=0$ outside $\ScrX$. The function $h$ is constructed as follows:
\[h=\int_{\ScrI}k(\cdot,\BFx)w(\BFx)\dd{\BFx}.\]
By applying the Green's function property of $k$, one can check that
\[h(\BFx)=\int_{\ScrI}k(\cdot,\BFx)\prod_{j=1}^d\CalL_j[g](\BFx)\dd{\BFx}=g(\BFx),\quad \forall\BFx\in\ScrX.\]
Moreover, we can apply equation~\eqref{eq:finite-k-2} to show:
\[\int_{\ScrI}\biggl|\prod_{j=1}^d\CalL_j[h](\BFx)\biggr|^2\dd{\BFx}=\int_\ScrI\bigl|w(\BFx)\bigr|^2\dd{\BFx}=\int_\ScrX\bigl|w(\BFx)\bigr|^2\dd{\BFx}<\infty.\]
Hence, the equivalent representation for function in $\ScrH_k^2$ provided in Lemma~\ref{lemma:second-order-identity} indicates that $h\in\ScrH_k^2$.\Halmos\endproof

\subsection{General Results on Convergence Rates}

To summarize the analysis so far,
Proposition~\ref{prop:BM-kernel-SL-operator} asserts that BF kernels satisfy Assumption~\ref{assump:SL};
moreover,
Proposition~\ref{prop:SL-norm-equivalence} (resp.,
Proposition~\ref{prop:H-2-mix}) shows that $\ScrH^1_{\mathsf{mix}}$  is identical to $\ScrH_k(\ScrX)$ (resp.,  $\ScrH^2_{\mathsf{mix}}$  is identical to  $\ScrH_k^2(\ScrX)$) as a set of functions for any kernel $k$ that satisfies Assumption~\ref{assump:SL}.

Note that the two Sobolev spaces with dominating mixed smoothness, $\ScrH^1_{\mathsf{mix}}$ and $\ScrH^2_{\mathsf{mix}}$, are the main driving force for our theory.
Note also that the fast kernel matrix inversion algorithms (Algorithms~\ref{alg:TM-1D}--\ref{alg:TM-TSG} in Section~\ref{sec:fast-computation})  that support the computational efficiency of Algorithm~\ref{alg:EI} in high dimensions are designed for  TM kernels in the first place.
Hence, it is natural for us to  generalize Algorithm~\ref{alg:EI} and the theoretical results (Theorems~\ref{thm:EI_noiseless}--\ref{thm:Convergence_EI_misspecification}) from BF kernels to TM kernels.

Throughout the rest of this supplemental material,
we will consider a generalized version of Algorithm~\ref{alg:EI} in which the BF kernel is replaced with a TM kernel that satisfies Assumption~\ref{assump:SL}, but will still refer to it as Algorithm~\ref{alg:EI} for simplicity.
We will also suppress the dependence of $\ScrH^m_k(\ScrX)$ on $\ScrX$ and use $\ScrH_k$ instead.

Recall that both Theorem~\ref{thm:EI_noiseless} and Theorem~\ref{thm:EI_noiseless_misspecification} cover the noise-free case,
with the former assuming $f\in\ScrH_{\mathsf{mix}}^1$ while the latter assuming $f\in\ScrH_{\mathsf{mix}}^2$.
We generalize them to Theorem~\ref{thm:EI_noiseless_sum} and prove it in Section~\ref{appdix:noise-free-convergence}.
Likewise, both Theorem~\ref{thm:Convergence_EI} and Theorem~\ref{thm:Convergence_EI_misspecification} cover the noisy case, and
we generalize them to Theorem~\ref{thm:Convergence_EI_sum} and prove it in Section~\ref{appdix:noisy-convergence}.

\begin{theorem}
\label{thm:EI_noiseless_sum}
Let $k$ be a kernel satisfying Assumption~\ref{assump:SL}.
Suppose that $f\in\ScrH^{m}_{\mathsf{mix}}$ with $m=1,2$, Assumption~\ref{assump:optimum} holds, and Assumption~\ref{assump:sub-Gaussian} holds with $\sigma=0$.
Let $\lambda=0$  and $\delta_n = 1$ in Algorithm~\ref{alg:EI}.
Then,
\[f(\BFx^*)-f(\widehat{\BFx}^*_N)=\CalO\left(N^{-\frac{2m-1}{2}}(\log N)^{\frac{(2m+1)(d-1)}{2}}\right).\]
\end{theorem}

\begin{theorem}
\label{thm:Convergence_EI_sum}
Let $k$ be a kernel satisfying Assumption~\ref{assump:SL}.
Suppose that $f\in\ScrH^{m}_{\mathsf{mix}}$ with $m=1,2$, Assumption~\ref{assump:optimum} holds, and Assumption~\ref{assump:sub-Gaussian} holds with $\sigma>0$.
Let
\begin{align*}
    &\lambda \asymp \sigma^{\frac{4}{2m+1}}N_{\tau}^{-\frac{2}{2m+1}}|\log (\sigma N_\tau)|^{\frac{2d-1}{2m+1}}(\log N_\tau)^{\frac{6(m-1)(1-d)}{5}},\\
    &\delta_n^2\asymp (\sigma^2 n^{-1})^{\frac{2m-1}{2m+1}}|\log (\sigma n)|^{\frac{1-2d}{2m+1}}(\log n)^{\frac{6(m-1)(d-1)}{5}},
\end{align*}
in Algorithm~\ref{alg:EI}, then
\begin{align*}
\E[f(\BFx^*)-f(\widehat{\BFx}^*_N)]
={} \CalO\biggl(\sigma^{\frac{2m-1}{2m+1}}N^{-\frac{2m-1}{4m+2}}|\log (\sigma N)|^{\frac{(2m-1)(2d-1)}{4(4m+1)}}(\log N)^{\frac{(6m-3)(d-1)}{6m-2}}\biggr).
\end{align*}
\end{theorem}

We also generalize Proposition~\ref{prop:FG-convergence} to the following.
\begin{proposition}
\label{prop:FG-convergence-ec}
Let $k$ be a kernel satisfying Assumption~\ref{assump:SL}.
Suppose that $f\in\ScrH^1_{\mathsf{mix}}$, Assumption~\ref{assump:optimum} holds, and Assumption~\ref{assump:sub-Gaussian} holds with $\sigma=0$.
Let the design points  $\{\BFx_1,\ldots,\BFx_N\}$  form a full grid
$\bigtimes_{j=1}^d \CalX_{j,\tau} $ for some $\tau\geq 1$ where $\CalX_{j,\tau} = \{i\cdot 2^{-\tau}:1\leq i\leq 2^\tau-1\}$.
Then, for any $0<L<\infty$, there exists a constant $c>0$ such that for all sufficiently large $N$,
\[
\sup_{\|f\|_{\ScrH_{\mathsf{mix}}^1}\leq L}\biggl\{\max_{\BFx\in\ScrX}f(\BFx)-\max_{x\in\ScrX}f(\BFx)\biggr\} \geq  c N^{-\frac{1}{2d}}.
\]
\end{proposition}

\section{Grid-based Expansions}

Recall the following notations used for defining sparse grids.
Let $\SFc_{l, i}\coloneqq i\cdot 2^{-l}$ for $l\geq 1$ and $i=1,\ldots,2^l -1$,
and let
$\BFc_{\BFl,\BFi}\coloneqq (\SFc_{l_1, i_1},\ldots, \SFc_{l_d, i_d})$.
The design points in \eqref{eq:dyadic} is then written as $\CalX_{j, l} = \{\SFc_{l, i}: i=1,\ldots,2^l-1 \}$ for all $j=1,\ldots,d$ and $l=1,\ldots,\tau$.
For any level multi-index $\BFl=(l_1,\ldots,l_d)\in \NatInt^d$, we define a set for the  multi-index $\BFi=(i_1,\ldots,i_d)$ as follows
\begin{equation}\label{eq:index-set-rho-ec}
\rho(\BFl) \coloneqq \bigtimes_{j=1}^d \{i_j: i_j\mbox{ is an odd number between 1 and }2^{l_j} \}  = \bigtimes_{j=1}^d  \{1, 3, 5, \ldots, 2^{l_j}-1 \}.
\end{equation}

\begin{definition}
\label{def:feature}
Let $k(\BFx,\BFx') = \prod_{j=1}^d p_j(x_j\wedge x_j')q_j(x_j\vee x_j')$ be a TM kernel satisfying Assumption~\ref{assump:SL}.
For each $l\geq 1$ and $i=1,\ldots, 2^l-1$,
let $\SFc_{l,i}\coloneqq i\cdot 2^{-l}$ and define  the following continuous function with  support $(\SFc_{l,i-1},\SFc_{l,i+1})=((i-1)2^{-l},(i+1)2^{-l})$:
\begin{equation}\label{eq:feature}
\phi_{j, l, i}(x) \coloneqq \left\{
\begin{array}{ll}
\displaystyle \frac{p_j(x)\SFq_{j,l,i-1} - \SFp_{j,l,i-1}q_j(x)}{\SFp_{j,l,i}\SFq_{j,l,i-1} - \SFp_{j,l,i-1}\SFq_{j,l,i}},     & \quad\mbox{if } x \in (\SFc_{l, i-1}, \SFc_{l, i}], \\[2ex]
\displaystyle \frac{\SFp_{j,l,i+1} q_j(x) - p_j(x) \SFq_{j,l,i+1} }{\SFp_{j,l,i+1}\SFq_{j,l,i} - \SFp_{j,l,i}\SFq_{j,l,i+1}},
     & \quad\mbox{if } x \in (\SFc_{l, i}, \SFc_{l, i+1}), \\[2ex]
0, & \quad\mbox{otherwise },
\end{array}
\right.
\end{equation}
where
$\SFp_{j,l,i}=p_j(\SFc_{l, i})$ and $\SFq_{j,l,i}=q_j(\SFc_{l, i})$.
For any multi-indices $\BFl,\BFi\in\NatInt^d$, define
\begin{equation}\label{eq:basis-func-tensor}
\phi_{\BFl,\BFi}(\BFx) \coloneqq \prod_{j=1}^d \phi_{j,l_j,i_j}(x_j).
\end{equation}
\end{definition}

We need to estimate the $L^2$ norm and the RKHS norm of $\{\phi_{\BFl,\BFi}\}$ for the subsequent rate analysis.
\begin{lemma}
\label{lemma:norm_estimate}
Let $k$ be a kernel satisfying Assumption~\ref{assump:SL} and
$\{\phi_{\BFl,\BFi}:\BFl\in\NatInt^d,\,\BFi\in\rho(\BFl)\}$ be the orthogonal basis in Definition~\ref{def:feature}. Then,
\begin{align*}
    \|\phi_{\BFl,\BFi}\|_2=\CalO(2^{-\abs{\BFl}/2}) \qq{and} \|\phi_{\BFl,\BFi}\|_{\ScrH_k}\asymp 2^{\abs{\BFl}/2}.
\end{align*}
\end{lemma}
\proof{Proof.}
The
$L^2$ norm estimate is from Lemma~5 in \cite{DingTuoShahrampour20_ec} and the RKHS norm estimate is from Lemma~EC.2 in \cite{DingZhang21_ec}.
\Halmos\endproof

To distinguish the KI estimator from the KRR estimator,
assuming the design points $\{\BFx_1,\ldots,\BFx_n\}$ form a TSG $\CalX^{\mathsf{TSG}}_{n}$,
we use the following notation:
\begin{equation}
    \label{eq:Kriging-ec}
    \breve{f}_n(\BFx)\coloneqq \BFk^\intercal(\BFx)\BFK^{-1}f(\CalX^{\mathsf{TSG}}_{n}),
\end{equation}
where $f(\CalX^{\mathsf{TSG}}_{n}) = (f(\BFx_1),\ldots,f(\BFx_n))^\intercal$,
$\BFk(\BFx) = (k(\BFx_1,\BFx),\ldots, k(\BFx_n,\BFx))^\intercal$, and $\BFK$ is the kernel matrix $(k(\BFx_i,\BFx_j))_{i,j=1}^n$.
Similarly, given a multi-index $\BFl$ and a full grid $\CalX^{\mathsf{FG}}_{\BFl}$,
we let
\begin{equation}\label{eq:Kriging-ec-FG}
\breve{f}_{\BFl}^{\mathsf{FG}}(\BFx)\coloneqq \BFk_{\BFl}^\intercal(\BFx) \BFK^{-1}_{\BFl}f(\CalX^{\mathsf{FG}}_{\BFl}),
\end{equation}
where  $\BFk_{\BFl}(\BFx)$ is the vector composed of $k(\BFx,\BFx')$ for all $\BFx'\in \CalX^{\mathsf{FG}}_{\BFl}$, and
$\BFK_{\BFl}$ is the matrix composed of $k(\BFx',\BFx)$ for all $\BFx,\BFx'\in \CalX^{\mathsf{FG}}_{\BFl}$.

\begin{lemma}
\label{lemma:orthogonalExpansion}
Let $k$ be a kernel satisfying Assumption~\ref{assump:SL},
$\{\phi_{\BFl,\BFi}:\BFl\in\NatInt^d,\,\BFi\in\rho(\BFl)\}$ be the orthogonal basis in Definition~\ref{def:feature}, and
$\breve{f}_n$ be the KI estimator \eqref{eq:Kriging-ec} on a TSG $\CalX^{\mathsf{TSG}}_n$. Then,
\[\breve{f}_n(\BFx)=\sum_{(\BFl,\BFi):\BFl\in\NatInt^d,\BFi\in\rho(\BFl),\BFc_{\BFl,\BFi}\in\CalX_n^{\mathsf{TSG}}}\langle f,\phi_{\BFl,\BFi}\rangle_{\ScrH_k}\frac{\phi_{\BFl,\BFi}(\BFx)}{\|\phi_{\BFl,\BFi}\|_{\ScrH_k}^2}.\]
\end{lemma}
\proof{Proof.}
According to Proposition~EC.1 in \cite{DingZhang21_ec}, for any kernel $k$ satisfying Assumption~\ref{assump:SL}, we have the following identities for $k$
\begin{align*}
    &\BFk^\intercal(\BFx')\BFK\BFk(\BFx'') = \sum_{(\BFl,\BFi):\BFl\in\NatInt^d,\BFi\in\rho(\BFl),\BFc_{\BFl,\BFi}\in\CalX_n^{\mathsf{TSG}}}
\frac{\phi_{\BFl,\BFi}(\BFx')\phi_{\BFl,\BFi}(\BFx'')}{\|\phi_{\BFl,\BFi}\|_{\ScrH_k}^2}, \\
&k(\BFx',\BFx'') =\sum_{(\BFl,\BFi):\BFl\in\NatInt^d,\BFi\in\rho(\BFl),\BFc_{\BFl,\BFi}\in\CalX_n^{\mathsf{TSG}}}
\frac{\phi_{\BFl,\BFi}(\BFx')\phi_{\BFl,\BFi}(\BFx'')}{\|\phi_{\BFl,\BFi}\|_{\ScrH_k}^2}+\sum_{(\BFl,\BFi):\BFl\in\NatInt^d,\BFi\notin\rho(\BFl),\BFc_{\BFl,\BFi}\in\CalX_n^{\mathsf{TSG}}}
\frac{\phi_{\BFl,\BFi}(\BFx')\phi_{\BFl,\BFi}(\BFx'')}{\|\phi_{\BFl,\BFi}\|_{\ScrH_k}^2}.
\end{align*}
Moreover, we can notice that for any $\phi_{\BFl,\BFi}$ with $\BFc_{\BFl,\BFi}\notin\CalX_n^{\mathsf{TSG}}$, $\phi_{\BFl,\BFi}=0$ on $\CalX_n^{\mathsf{TSG}}$ because $k(\BFx,\BFx)-\BFk^\intercal(\BFx)\BFK\BFk(\BFx)=0$ for any $\BFx\in\CalX^{\mathsf{TSG}}_n$. Hence, by the representer theorem (Lemma~\ref{lemma:representation-thm}) gives
\begin{align*}
    \breve{f}_n(\BFx)={} & \sum_{(\BFl,\BFi):\BFl\in\NatInt^d,\BFi\in\rho(\BFl),\BFc_{\BFl,\BFi}\in\CalX_n^{\mathsf{TSG}}}\beta_{\BFl,\BFi}k(\BFc_{\BFl,\BFi},\BFx)\\
    ={} & \sum_{(\BFl,\BFi):\BFl\in\NatInt^d,\BFi\in\rho(\BFl),\BFc_{\BFl,\BFi}\in\CalX_n^{\mathsf{TSG}}}\beta_{\BFl,\BFi}\sum_{(\BFl',\BFi'):\BFl'\in\NatInt^d,\BFi'\in\rho(\BFl'),\BFx_{\BFl',\BFi'}\in\CalX_n^{\mathsf{TSG}}}
\frac{\phi_{\BFl',\BFi'}(\BFx)\phi_{\BFl',\BFi'}(\BFc_{\BFl,\BFi})}{\|\phi_{\BFl',\BFi'}\|_{\ScrH_k}^2}\\
    ={} &\sum_{(\BFl',\BFi'):\BFl'\in\NatInt^d,\BFi'\in\rho(\BFl'),\BFx_{\BFl',\BFi'}\in\CalX_n^{\mathsf{TSG}}}\Biggl[\sum_{(\BFl,\BFi):\BFl\in\NatInt^d,\BFi\in\rho(\BFl),\BFc_{\BFl,\BFi}\in\CalX_n^{\mathsf{TSG}}}\beta_{\BFl,\BFi}\frac{\phi_{\BFl',\BFi'}(\BFc_{\BFl,\BFi})}{\|\phi_{\BFl',\BFi'}\|_{\ScrH_k}^2}\Biggr]\phi_{\BFl',\BFi'}(\BFx)\\
    \coloneqq{} & \sum_{(\BFl',\BFi'):\BFl'\in\NatInt^d,\BFi'\in\rho(\BFl'),\BFx_{\BFl',\BFi'}\in\CalX_n^{\mathsf{TSG}}}\beta'_{\BFl',\BFi'}\phi_{\BFl',\BFi'}(\BFx).
\end{align*}

Our next step is to determine the constant vector $\boldsymbol{\beta'}$. Notice that  any $\BFx\in\CalX^{\mathsf{TSG}}_n$, $f(\BFx)=\breve{f}_n(\BFx)$, so we can apply the reproducing property of RKHS:
 \begin{align*}
    \sum_{(\BFl,\BFi):\BFl\in\NatInt^d,\BFi\in\rho(\BFl'),\BFc_{\BFl,\BFi}\in\CalX_n^{\mathsf{TSG}}}\beta'_{\BFl,\BFi}\phi_{\BFl,\BFi}(\BFx)= \breve{f}_n(\BFx)
     =  f(\BFx)
     ={} & \langle f,k(\BFx,\cdot)\rangle_{\ScrH_k}  \\
     ={}&\langle f,\BFk^\intercal(\BFx)\BFK^{-1}\BFk(\cdot)\rangle_{\ScrH_k}\\
     ={} & \sum_{(\BFl,\BFi):\BFl\in\NatInt^d,\BFi\in\rho(\BFl),\BFc_{\BFl,\BFi}\in\CalX_n^{\mathsf{TSG}}}
\frac{\langle f,\phi_{\BFl,\BFi}\rangle_{\ScrH_k}}{\|\phi_{\BFl,\BFi}\|_{\ScrH_k}^2}\phi_{\BFl,\BFi}(\BFx),
\end{align*}
which is the desired result.
\Halmos\endproof

\begin{lemma}\label{lemma:FGexpansion}
Let $k$ be a kernel satisfying Assumption~\ref{assump:SL},
$\{\phi_{\BFl,\BFi}:\BFl\in\NatInt^d,\,\BFi\in\rho(\BFl)\}$ be the orthogonal basis in Definition~\ref{def:feature}, and $\breve{f}_{\BFl}^{\mathsf{FG}}(\BFx)$ be defined in \eqref{eq:Kriging-ec-FG}.
Then,
\begin{equation}\label{eq:FGexpansion}
    \breve{f}_{\BFl}^{\mathsf{FG}}(\BFx)=\sum_{(\BFl,\BFi):\BFc_{\BFl,\BFi}\in\CalX^{\mathsf{FG}}_{\BFl}}f(\BFc_{\BFl,\BFi})\phi_{\BFl,\BFi}(\BFx).
\end{equation}
\end{lemma}
\proof{Proof.}
See Theorem 1 in \cite{ding2019bdrygp_ec}.
\Halmos\endproof

\section{Convergence Rates When Samples are Noise-free}
\label{appdix:noise-free-convergence}

The main idea of the proof relies on the following inequality:
\begin{align}
f(\BFx^*)-f(\widehat{\BFx}^*_N)={}& [f(\BFx^*)-\widetilde{f}_N(\BFx^*)] + [\widetilde{f}_N(\BFx^*) - \widetilde{f}_N(\widehat{\BFx}^*_N)]
+ [\widetilde{f}_N(\widehat{\BFx}^*_N)  - f(\widehat{\BFx}^*_N)] \nonumber \\
\leq{}& \|f-\widetilde{f}_N\|_\infty + 0 + \|f-\widetilde{f}_N\|_\infty  = 2 \|f-\widetilde{f}_N\|_\infty, \label{eq:decomp-norm}
\end{align}
where $\widetilde{f}_N(\BFx^*) - \widetilde{f}_N(\widehat{\BFx}^*_N) \leq 0$ because  $\widehat{\BFx}^*_N$ maximizes $\widetilde{f}_N$. Therefore, it suffices to estimate the $L^\infty$ error of $\widetilde{f}_N$.

We first introduce two lemmas that are useful for calculations involving the size of a classical SG.

\begin{lemma}
\label{lem:number-SG}
The number of grid points in a $d$-dimensional level-$\tau$ classical SG is given by
\begin{align*}
    \lvert \CalX^{\mathsf{SG}}_\tau\rvert=\sum_{i=0}^{\tau-1}2^i{\binom{d-1+i}{d-1}}=\CalO(2^\tau \tau^{d-1}).
\end{align*}
\end{lemma}
\proof{Proof.}
See Lemma~3.6 in \cite{BungartzGriebel04_ec}.
\Halmos\endproof

\begin{lemma}
\label{lem:bungartz-Lemma-3-7}
For any $s>0$ and $\tau\in\NatInt$,
\begin{align*}
   \sum_{\abs{\BFl}>\tau+d-1}2^{-s\abs{\BFl}}\leq \CalO(2^{-s\tau}\tau^{d-1}).
\end{align*}
\end{lemma}
\proof{Proof.}
From a direct calculation, we have
\begin{align*}
    \sum_{\abs{\BFl}>\tau+d-1}2^{-s\abs{\BFl}}={} &\sum_{i=\tau+d}^\infty 2^{-s i}\sum_{\abs{\BFl}=i}1\\
    ={} & \sum_{i=\tau+d}^\infty 2^{-s i}{\binom{i-1}{d-1}}\\
    ={}& 2^{-s\tau}\cdot 2^{-sd}\sum_{i=0}^{\infty}2^{-si}{\binom{\tau +i + d- 1}{ d-1}}.
\end{align*}
For equation~(3.67) in \cite{BungartzGriebel04_ec}, we get for any $x\in\Real$
\[\sum_{i=0}^{\infty}x^i{\binom{\tau +i + d- 1}{d-1}}=\sum_{j=0}^{d-1}{\binom{\tau+d-1}{ j}}\biggl(\frac{x}{1-x}\biggr)^{d-1-j}\frac{1}{1-x}.\]
Let $x=2^{-s}$ and make the substitution, we can get
\begin{align*}
    \sum_{\abs{\BFl}>\tau+d-1}2^{-s\abs{\BFl}}={} &
    2^{-s\tau}\Biggl(2^{-sd}\sum_{i=0}^{\infty}2^{-si}{\binom{\tau +i + d- 1}{ d-1}}\Biggr)\\
    {}\leq &  2^{-s\tau}\max_{j=0,\ldots,d-1}\Biggl\{\frac{2^{-sd}}{1-2^{-s}}\left(\frac{2^{-s}}{1-2^{-s}}\right)^{d-1-j}\Biggr\}\sum_{j=0}^{d-1}{\binom{\tau+d-1}{ j}}\\
    ={}& \CalO(2^{-s\tau}\tau^{d-1})
\end{align*}
where the last equality is from equation~(3.65) in \cite{BungartzGriebel04_ec}, which states that
$\sum_{j=0}^{d-1}{\binom{\tau+d-1}{j}}=\CalO(\tau^{d-1})$.
\Halmos
\endproof

\subsection{Convergence Rate of Kernel Interpolation}

\begin{proposition}
\label{prop:Convergence_SGI}
Let $k$ be a kernel satisfying Assumption~\ref{assump:SL}.
Suppose that $f\in\ScrH^{m}_{\mathsf{mix}}$ with $m=1,2$, Assumption~\ref{assump:optimum} holds, and Assumption~\ref{assump:sub-Gaussian} holds with $\sigma=0$.
Let  $\breve{f}_n$ be the KI estimator \eqref{eq:Kriging-ec} on a TSG $\CalX^{\mathsf{TSG}}_n$.
Then,
\begin{align*}
    &\|f-\breve{f}_n\|_\infty=\CalO\left(n^{-\frac{2m-1}{2}}(\log n)^{\frac{(2m+1)(d-1)}{2}}\right).
\end{align*}
\end{proposition}

\proof{Proof.}
Without loss of generality, we can only consider function $f\in\ScrH^m_k$, $m=1,2$ because $\ScrH_k^m$ is equivalent to $\ScrH^m_{\mathsf{mix}}$ for $m=1,2$.

From Lemma~\ref{lemma:orthogonalExpansion} and the definition of feature functions $\{\phi_{\BFl,\BFi}\}$, we can see that  if the index $(\BFl,\BFi)$ of a feature $\phi_{\BFl,\BFi}$  satisfies $\BFl\in\NatInt^d$, $\BFi\in\rho(\BFl)$ and $\BFc_{\BFl,\BFi}\not\in \CalX^{\mathsf{TSG}}_n$, $\phi_{\BFl,\BFi}(\CalX^{\mathsf{TSG}}_n)=0$. As a result, $\breve{f}_n$ can be see as a projection of $f$ onto the function spaces $\{\phi_{\BFl,\BFi}: \BFc_{\BFl,\BFi}\in\CalX^{\mathsf{TSG}}_n\}$:
 \[\breve{f}_n=\sum_{(\BFl,\BFi):\BFc_{\BFl,\BFi}\in\CalX^{\mathsf{TSG}}_n}\langle f,\phi_{\BFl,\BFi}\rangle_{\ScrH_k}\frac{\phi_{\BFl,\BFi}}{\|\phi_{\BFl,\BFi}\|^2_{\ScrH_k}}.\]

\textbf{Case (i): $f\in\ScrH^2_{k}$.}   We have the following $L^2$ error estimate of $\breve{f}_n$:
\begin{align*}
    \|f-\breve{f}_n\|_2={}& \biggl \|\sum_{(\BFl,\BFi):\BFl\in\NatInt^d,\BFi\in\rho(\BFl),\BFc_{\BFl,\BFi}\not\in\CalX^{\mathsf{TSG}}_n}\langle f,\phi_{\BFl,\BFi}\rangle_{\ScrH^1_{\mathsf{mix}}}\frac{\phi_{\BFl,\BFi}}{\|\phi_{\BFl,\BFi}\|_{\ScrH^1_{\mathsf{mix}}}^2}\biggr\|_2\\
    \leq{}& \sum_{(\BFl,\BFi):\BFl\in\NatInt^d,\BFi\in\rho(\BFl),\BFc_{\BFl,\BFi}\not\in\CalX^{\mathsf{TSG}}_n}\abs{\langle f,\phi_{\BFl,\BFi}\rangle_{\ScrH^1_{\mathsf{mix}}}}\frac{\|\phi_{\BFl,\BFi}\|_2}{\|\phi_{\BFl,\BFi}\|_{\ScrH^1_{\mathsf{mix}}}^2}\\
    ={}&\sum_{(\BFl,\BFi):\BFl\in\NatInt^d,\BFi\in\rho(\BFl),\BFc_{\BFl,\BFi}\not\in\CalX^{\mathsf{TSG}}_n}\biggl|\int_\ScrI\prod_{j=1}^d\CalL_j[f](\BFx)\phi^*_{\BFl,\BFi}(\BFx)\dd{\BFx}\biggr|\frac{\|\phi_{\BFl,\BFi}\|_2}{\|\phi_{\BFl,\BFi}\|_{\ScrH_k}^2}\\
    \leq{}& \sum_{(\BFl,\BFi):\BFl\in\NatInt^d,\BFi\in\rho(\BFl),\BFc_{\BFl,\BFi}\not\in\CalX^{\mathsf{TSG}}_n}\bigg\| \prod_{j=1}^d\CalL_j [f]\bigg|_{\supp[\phi_{\BFl,\BFi}]}\bigg\|_2\frac{\|\phi_{\BFl,\BFi}\|_2^2}{\|\phi_{\BFl,\BFi}\|_{\ScrH_k}^2}\\
    \leq{}& \sum_{(\BFl,\BFi):\BFl\in\NatInt^d,\BFi\in\rho(\BFl),\BFc_{\BFl,\BFi}\not\in\CalX^{\mathsf{SG}}_\tau}\bigg\| \prod_{j=1}^d\CalL_j [f]\bigg|_{\supp[\phi_{\BFl,\BFi}]}\bigg\|_2\frac{\|\phi_{\BFl,\BFi}\|_2^2}{\|\phi_{\BFl,\BFi}\|_{\ScrH_k}^2}\\
    \leq{}& \CalO\biggl( \sum_{\abs{\BFl}>\tau+d-1}2^{-2\abs{\BFl}}\biggr)\\
    ={}&\CalO\left(2^{-2\tau}\tau^{d-1}\right)
\end{align*}
where $g\big|_{A}$ denotes a version of function $g$ such that $g\big|_{A}=g$ on set $A$ and $g\big|_{A}=0$ outside of $A$, the second line is from triangular inequality over the summation, the fourth line is from Lemma~\ref{lemma:second-order-identity} and the Cauchy–Schwarz  inequality, the fifth line is because the TSG $\CalX^{\mathsf{{TSG}}}_n$ is between SG $\CalX^{\mathsf{SG}}_\tau$ and SG $\CalX^{\mathsf{SG}}_{\tau+1}$: \[\abs{\CalX^{\mathsf{SG}}_\tau}\leq n <\abs{\CalX^{\mathsf{SG}}_{\tau+1}}, \]
the sixth line is from the definition of $\CalX^{\mathsf{SG}}_{\tau}:$
\[\CalX^{\mathsf{SG}}_{\tau}=\{\BFc_{\BFl,\BFi}:\abs{\BFl}\leq \tau+d-1,\BFi\in\rho(\BFl)\},\]
the fact that supports of $\{\phi_{\BFl,\BFi}: \BFi\in\rho(\BFl)\}$ form a partition of $\ScrX$:
\[ \sum_{\BFi: \BFi\in\rho(\BFl)}\bigg\| \prod_{j=1}^d\CalL_j [f]\bigg|_{\supp[\phi_{\BFl,\BFi}]}\bigg\|_2=\bigg\| \prod_{j=1}^d\CalL_j [f]\bigg\|_2<\infty\]
and Lemma~\ref{lemma:norm_estimate}, the last line is from Lemma \ref{lem:bungartz-Lemma-3-7}.

We now prove the convergence rate of $\breve{f}_n$ to $f$ under the RKHS norm when $f\in\ScrH^{2}_{k}$.   We can  apply the same inequalities on the expansion  of $f-\breve{f}_n$:
\begin{align*}
    \|f-\breve{f}_n\|_{\ScrH_{k}}={}&\bigg\|\sum_{(\BFl,\BFi):\BFl\in\NatInt^d,\BFi\in\rho(\BFl),\BFc_{\BFl,\BFi}\not\in\CalX^{\mathsf{TSG}}_n}\langle f,\phi_{\BFl,\BFi}\rangle_{\ScrH_k}\frac{\phi_{\BFl,\BFi}}{\|\phi_{\BFl,\BFi}\|_{\ScrH_k}^2}\bigg\|_{\ScrH_k}\\
    \leq{}&\sum_{(\BFl,\BFi):\BFl\in\NatInt^d,\BFi\in\rho(\BFl),\BFc_{\BFl,\BFi}\not\in\CalX^{\mathsf{TSG}}_n}\abs{\langle f,\phi_{\BFl,\BFi}\rangle_{\ScrH_k}}\frac{\|\phi_{\BFl,\BFi}\|_{\ScrH_k}}{\|\phi_{\BFl,\BFi}\|_{\ScrH_k}^2}\\
     \leq{}&\sum_{(\BFl,\BFi):\BFl\in\NatInt^d,\BFi\in\rho(\BFl),\BFc_{\BFl,\BFi}\not\in\CalX^{\mathsf{TSG}}_n}\bigg\| \prod_{j=1}^d\CalL_j [f]\bigg|_{\supp[\phi_{\BFl,\BFi}]}\bigg\|_2\frac{\|\phi_{\BFl,\BFi}\|_2}{\|\phi_{\BFl,\BFi}\|_{\ScrH_k}}\\
     \leq{}& \CalO\biggl( \sum_{\abs{\BFl}>\tau+d-1}2^{-\abs{l}}\biggr)\\
     ={}&\CalO\left(2^{-\tau}\tau^{d-1}\right)
\end{align*}
According to Proposition~\ref{prop:SL-norm-equivalence}, we have the following norm equivalence:
\[ \|f\|_{\ScrH_k}\asymp \biggl\|\frac{\partial^d f}{\partial x_1\cdots\partial x_d} \biggr\|_2.\]
Now we can apply Corollary~\ref{cor:Gagliardo–Nirenberg} to get:
\begin{equation}\label{eq:GN-inequality-case}
    \|f-\breve{f}_n\|_{\infty}\leq C_1\biggl\|\frac{\partial^d(f-\breve{f}_n)}{\partial x_1\cdots\partial x_d}\biggr\|_2^{\frac{1}{2}}\bigl\|f-\breve{f}_n\bigr\|_{2}^{\frac{1}{2}}=C_2\bigl\|f-\breve{f}_n\bigr\|_{\ScrH_k}^{\frac{1}{2}}\bigl\|f-\breve{f}_n\bigr\|_{2}^{\frac{1}{2}}=\CalO(2^{-3\tau/2}\tau^{d-1}).
\end{equation}
\textbf{Case (ii): $f\in \ScrH_{k}^1$ but $f\not \in \ScrH^2_{k^2}$.} For any $\BFx\in \ScrX$, we have
\begin{align*}
    f(\BFx)-\breve{f}_n(\BFx)={}&\sum_{(\BFl,\BFi):\BFl\in\NatInt^d,\BFi\in\rho(\BFl),\BFc_{\BFl,\BFi}\not\in\CalX^{\mathsf{TSG}}_n}\biggl\langle \frac{\phi_{\BFl,\BFi}}{\|\phi_{\BFl,\BFi}\|_{\ScrH_k}},f\biggr\rangle_{\ScrH_k}\frac{\phi_{\BFl,\BFi}(\BFx)}{\|\phi_{\BFl,\BFi}\|_{\ScrH_k}}\\
    \leq{}& \sum_{(\BFl,\BFi):\BFl\in\NatInt^d,\BFi\in\rho(\BFl),\BFc_{\BFl,\BFi}\not\in\CalX^{\mathsf{TSG}}_n}\frac{\phi_{\BFl,\BFi}(\BFx)}{\|\phi_{\BFl,\BFi}\|_{\ScrH_k}}\|f\|_{\ScrH_k}\\
    \leq{}&  \sum_{\abs{\BFl}>\tau+d-1}\frac{C}{\|\phi_{\BFl,\BFi}\|_{\ScrH_{k}}}\|f\|_{\ScrH_k}\\
    ={}&\CalO\biggl( \sum_{\abs{\BFl}>\tau+d-1}2^{-\abs{\BFl}/2}\|f\|_{\ScrH_k}\biggr)\\
    ={}&\CalO(2^{-\tau/2}\tau^{d-1})
\end{align*}
where the second line is from the Cauchy–Schwarz  inequality, the third line is from  disjoint support property of $\{\phi_{\BFl,\BFi}\}$, the last line is again form Lemma~\ref{lem:bungartz-Lemma-3-7}. Therefore, if $f\in\ScrH^m_{k}$ with $m=1$ or $2$, we can summarize the $L^\infty$ convergence rate as:
\[\|f-\breve{f}_n\|_\infty=\CalO\left(2^{-(m-\frac{1}{2})\tau}\tau^{d-1}\right).\]
By Lemma~\ref{lem:number-SG}, the total number of point in $ {\ScrX}^{\mathsf{SG}}_{\tau}$ is $n=\CalO(2^{\tau}\tau^{d-1})$, we can obtain the result.
\Halmos\endproof

\subsection{Proof of Theorem~\ref{thm:EI_noiseless_sum}}
\proof{Proof.}
It follows from Proposition~\ref{prop:SL-norm-equivalence} and
Proposition~\ref{prop:H-2-mix} that  $f\in\ScrH_{k}^m$ for $m=1,2$.

By \eqref{eq:decomp-norm},
it suffices to bound
$\|f-\widetilde{f}_N\|_\infty$, which can be written as:
 \begin{align}
     \|f-\widetilde{f}_N\|_\infty ={}& \| f- \breve{f}_{N_\tau} - \BFk_N^\intercal(\cdot) \BFK_N ^{-1}(f(\CalS_N) - \breve{f}_{N_\tau}(\CalS_N))  \|_\infty \nonumber \\
     ={}& \| \underbrace{f- \BFk_N^\intercal(\cdot) \BFK_N ^{-1}f(\CalS_N)}_{J_1}  + \underbrace{\BFk_N^\intercal(\cdot) \BFK_N ^{-1} \breve{f}_{N_\tau}(\CalS_N) - \breve{f}_{N_\tau}}_{J_2}  \|_\infty, \label{eq:decomp-noise-free}
\end{align}
where $\CalS_N\coloneqq\{\BFx_i\}_{i=1}^N$ is the set of design points selected by Algorithm~\ref{alg:EI}, which forms a TSG $\CalX^{\mathsf{TSG}}_N$.

For $J_1$, note that by the definition \eqref{eq:Kriging-ec}, $\BFk_N^\intercal(\cdot) \BFK_N ^{-1}f(\CalS_N)=\breve{f}_N$.
Then, by Proposition \ref{prop:Convergence_SGI},
\begin{equation}\label{eq:J1-bound}
\|J_1\|_\infty=  \| f - \breve{f}_N \|_\infty = \CalO\left(N^{-\frac{2m-1}{2}}(\log N)^{\frac{(2m+1)(d-1)}{2}}\right).
\end{equation}

For $J_2$, we first apply Lemma \ref{lemma:orthogonalExpansion} to  $\breve{f}_{N_\tau}$:
\begin{equation}
    \label{eq:interpolation-expansion-mu-tau}
    \breve{f}_{N_\tau}(\BFx)=\sum_{(\BFl,\BFi):\BFl\in\NatInt^d,\BFi\in\rho(\BFl),\BFc_{\BFl,\BFi}\in\CalX^{\mathsf{SG}}_\tau}\biggl\langle \frac{\phi_{\BFl,\BFi}}{\|\phi_{\BFl,\BFi}\|_{\ScrH_k}},f\biggr\rangle_{\ScrH_k}\frac{\phi_{\BFl,\BFi}(\BFx)}{\|\phi_{\BFl,\BFi}\|_{\ScrH_k}}.
\end{equation}

Next, we note that by the definition \eqref{eq:Kriging-ec}, $\BFk_N(\cdot)^\intercal \BFK_N ^{-1}\breve{f}_{N_\tau}(\CalS_N)$ is the KI estimator of $\breve{f}_{N_\tau}$ based on data collected from $\CalS_N$.
Hence,
applying Lemma~\ref{lemma:orthogonalExpansion}  to $\BFk_N(\cdot)^\intercal \BFK_N ^{-1}\breve{f}_{N_\tau}(\CalS_N)$, we have:
\begin{align*}
    &\BFk_N(\BFx)^\intercal \BFK_N ^{-1}\breve{f}_{N_\tau}(\CalS_N)\\
    ={} & \sum_{(\BFl,\BFi):\BFl\in\NatInt^d,\BFi\in\rho(\BFl),\BFc_{\BFl,\BFi}\in\CalX^{\mathsf{TSG}}_N}\biggl\langle \frac{\phi_{\BFl,\BFi}}{\|\phi_{\BFl,\BFi}\|_{\ScrH_k}},\breve{f}_{N_\tau}\biggr\rangle_{\ScrH_k}\frac{\phi_{\BFl,\BFi}(\BFx)}{\|\phi_{\BFl,\BFi}\|_{\ScrH_k}}\\
    ={} & \sum_{(\BFl,\BFi):\BFl\in\NatInt^d,\BFi\in\rho(\BFl),\BFc_{\BFl,\BFi}\in\CalX^{\mathsf{TSG}}_N}\biggl\langle \frac{\phi_{\BFl,\BFi}}{\|\phi_{\BFl,\BFi}\|_{\ScrH_k}},\sum_{(\BFl,\BFi):\BFl\in\NatInt^d,\BFi\in\rho(\BFl),\BFc_{\BFl,\BFi}\in\CalX^{\mathsf{SG}}_\tau}\biggl\langle \frac{\phi_{\BFl,\BFi}}{\|\phi_{\BFl,\BFi}\|_{\ScrH_k}},f\biggr\rangle_{\ScrH_k}\frac{\phi_{\BFl,\BFi}(\BFx)}{\|\phi_{\BFl,\BFi}\|_{\ScrH_k}}\biggr\rangle_{\ScrH_k}\frac{\phi_{\BFl,\BFi}(\BFx)}{\|\phi_{\BFl,\BFi}\|_{\ScrH_k}}\\
    ={} &\sum_{(\BFl,\BFi):\BFl\in\NatInt^d,\BFi\in\rho(\BFl),\BFc_{\BFl,\BFi}\in\CalX^{\mathsf{SG}}_\tau}\biggl\langle \frac{\phi_{\BFl,\BFi}}{\|\phi_{\BFl,\BFi}\|_{\ScrH_k}},f\biggr\rangle_{\ScrH_k}\frac{\phi_{\BFl,\BFi}(\BFx)}{\|\phi_{\BFl,\BFi}\|_{\ScrH_k}} \\
    ={}& \breve{f}_{N_\tau}(\BFx),
\end{align*}
for all $\BFx\in\ScrX$,
where the second equality follows \eqref{eq:interpolation-expansion-mu-tau},
the third equality follows from the fact that $\{\phi_{\BFl,\BFi}:\BFl\in\NatInt^d,\BFi\in\rho(\BFl)\}$ are mutually orthogonal and the fact that $\CalX^{\mathsf{SG}}_\tau \subseteq \CalX^{\mathsf{TSG}}_N$,
and the last equality also follows from \eqref{eq:interpolation-expansion-mu-tau}.
Therefore, $J_2=0$, and thus
the proof is completed by combining  \eqref{eq:decomp-norm}, \eqref{eq:decomp-noise-free}, and \eqref{eq:J1-bound}.
\Halmos \endproof

\subsection{Proof of Proposition \ref{prop:FG-convergence-ec}}

\proof{Proof.}

\begin{figure}[ht]
\FIGURE{
\includegraphics[width=0.4\textwidth]{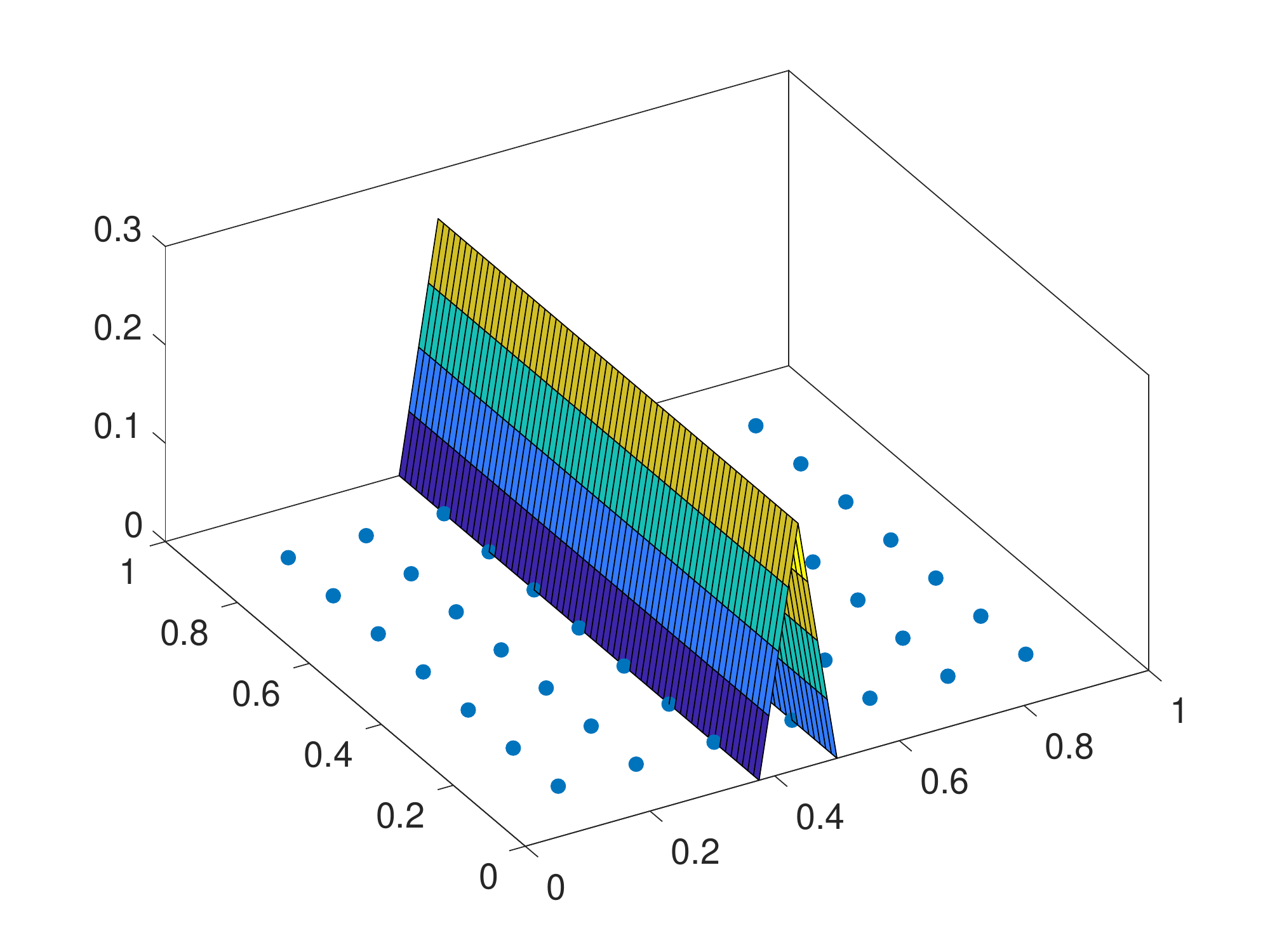}
}
{Two-dimensional $f_\tau$ \label{fig:ec_LowerBounds}}
{}
\end{figure}

Let $\CalX_{\tau}^{\mathsf{FG}} = \bigtimes_{j=1}^d\CalX_{j,\tau}$ denote the full grid of level $\tau$. We prove our statement by constructing the following function  $f_\tau$:
\[f_\tau(\BFx)=2^{-\frac{\tau+1}{2}}\max\biggl\{0, 1-\frac{\abs{x_1-0,5+2^{-\tau-1}}}{2^{-\tau-1}}\biggr\}\]
such that $f_\tau=0$ on $\CalX_\tau^{\mathsf{FG}}$ but $\max_{\BFx\in\ScrX}f_\tau(\BFx)>0$. In this case, any reasonable interpolation estimator of $f_\tau$ conditioned on observations on $\CalX^{\mathsf{FG}}_\tau$ can only equal to 0 and, hence, fail to estimate the maximizer of $f_\tau$.
An illustration of $f_\tau$ is shown in Figure~\ref{fig:ec_LowerBounds} for $\tau=4$ and $d=2$.

Firstly, for a point $\BFx^*\in\ScrX$, if $x^*_1=0.5-2^{-\tau-1}$,
then $\BFx^*$ maximizes $f_\tau$:
\[f_{\tau}(\BFx^*)=2^{-\frac{\tau+1}{2}}\max\bigl\{0, 1\bigr\}=2^{-\frac{\tau+1}{2}}=\max_{\BFx\in\ScrX}f_\tau(\BFx).\]

On the other hand, for a point $\BFx'\in\ScrX$, if $\abs{x'_1-0.5+2^{-\tau-1}}\geq2^{-\tau-1}$, then
\[f_\tau(\BFx')=2^{-\frac{\tau+1}{2}}\max\biggl\{0, 1-\frac{\abs{x_1'-0,5+2^{-\tau-1}}}{2^{-\tau-1}}\biggr\}=0.\]

Moreover, we need to check $f_\tau\in\ScrH_{\mathsf{mix}}^1$ for any $\tau\in\NatInt$. Because $f_{\tau}$ is independent of $x_2,\ldots,x_d$, we can treat $f_\tau$ as a function of $x_1$ and  verify that $\|\frac{\partial f_\tau}{\partial x_1}\|_2<\infty$ for any $\tau\in\NatInt$ via  the following calculation:
\begin{align*}
    \biggl\|\frac{\partial f_\tau}{\partial x_1}\biggr\|^2_2={} &\int_{0}^1\biggl|\frac{\partial f_\tau(x_1)}{\partial x_1}\biggr|^2\dd{x_1}
    =  2^{-({\tau+1})}\int_{0.5-2^{-\tau}}^{0.5} 2^{2(\tau+1)}\dd{x_1}=2.
\end{align*}
For any $\BFx\in\CalX_{\tau}^{\mathsf{FG}}$, we have $x_1=i^{2^{-\tau}}$ for some $i\in\NatInt$. So it is straightforward to check that $\abs{x_1-0.5+2^{-\tau-1}}\geq2^{-\tau-1}$ and
\[\max_{\BFx\in\ScrX}f_{\tau}(\BFx)-\max_{\BFx\in\CalX_{\tau}^{\mathsf{FG}}}f_\tau(\BFx)=2^{-(\tau+1)/2}.\]

Lastly, note that  $n=\abs{\CalX_\tau^{\mathsf{FG}}}=(2^\tau-1)^d\asymp 2^{\tau d}$, so $2^\tau \asymp n^{1/d}$ as $n\to\infty$. \Halmos\endproof

\section{Convergence Rates When Samples are Noisy}\label{appdix:noisy-convergence}

To simplify notations, for each integer $n$, we define the  \emph{empirical inner product} for any pair of functions $g$ and $h$ as
\[\langle g,h\rangle_n=\frac{1}{n}\sum_{i=1}^ng(\BFx_i)h(\BFx_i),\]
and define the associated \emph{empirical semi-norm} for any function $g$ as
\[\|g\|_n^2\coloneqq\frac{1}{n}\sum_{i=1}^n g^2(\BFx_i).\]

The main idea of our proof is as follows. We first establish the convergence rate of the KRR estimator under the empirical semi-norm by applying tools from empirical process theory \citep{vanderGeer00_ec}.
We then use Gagliardo–Nirenberg interpolation inequality (Corollary~\ref{cor:Gagliardo–Nirenberg}) and
an inequality (Lemma~\ref{lem:norming-ineqality}) which connects empirical semi-norm and other norms to convert the convergence rate under empirical semi-norm to the convergence rate  under the $L^\infty$ norm.

\subsection{Empirical Processes}

In this subsection, assuming that the design points $\{\BFx_1,\ldots,\BFx_n\}$ form a TSG $\CalX_n^{\mathsf{TSG}}$,
we prove the convergence rate of the KRR estimator
\begin{align}\label{eq:KRR-solution-ec}
\widehat{f}_{n,\lambda}=\BFk^\intercal(\BFx)\big[\BFK+\lambda n\BFI \big]^{-1}y(\CalX_n^{\mathsf{TSG}}),
\end{align}
under empirical semi-norm,
where $y(\CalX_n^{\mathsf{TSG}})=(y(\BFx_1),\ldots,y(\BFx_n))^\intercal$.
To this end, we estimate the distribution of the random variable $\sup_{\|g\|_{\ScrH_k}\leq 1}\langle g,\varepsilon\rangle_n$, which is closely related to the following concept.

\begin{definition}[Metric Entropy]\label{def:metric-entropy}
Let $\ScrG$ be a function space equipped with a norm $\|\cdot\|$.
For any $\epsilon>0$, let $\ScrB_\epsilon(f,\|\cdot\|)\coloneqq\{g\in\ScrG: \|g-h\|\leq\epsilon\}$ be an $\epsilon$-ball that is centered at $h\in\ScrG$.
The \emph{covering number}  $\CalN(\epsilon,\ScrG,\|\cdot\|)$ is defined as
\[\CalN(\epsilon,\ScrG,\|\cdot\|)\coloneqq\min\biggl\{n: \ \mbox{There exist } g_1,\ldots, g_n\in \ScrG \mbox{ such that } \ScrG\subseteq\bigcup_{i=1}^n \ScrB_{\epsilon}(g_i,\|\cdot\|)\biggr\}.\]
Then, $\CalH(\epsilon,\ScrG,\|\cdot\|) \coloneqq \log_2\CalN(\epsilon,\ScrG,\|\cdot\|)$ is called the \emph{metric entropy} of $\ScrG$.
\end{definition}

\begin{lemma}\label{lem:entropy}
Let $k$ be a kernel satisfying Assumption~\ref{assump:SL}.
Let $\ScrF\coloneqq\{g\in\ScrH_k: \|g\|_{\ScrH_k}\leq 1\}$. Then, there exists a positive constant $C$ such that
\[\CalH(\epsilon,\ScrF,\|\cdot\|_{\infty})\leq C\epsilon^{-1}\bigl|\log \epsilon|^{\frac{2d-1}{2}}. \]
\end{lemma}
\proof{Proof.}
See Lemma~2 in \cite{DingTuoShahrampour20_ec}
\Halmos\endproof

\begin{lemma}\label{lemma:geer8_3}
Let $a>0$, $R>0$, and $\ScrG$ be a function space.
Suppose that $\sup_{g\in\ScrG}\|f\|_n\leq R$ for and  $W\coloneqq\{W_i\}_{i=1}^n$ are independent zero-mean $\subG(\sigma^2)$ random variables.
If there exists some positive constant $C$ depending only on $\sigma$ and satisfying
\begin{equation}
    \label{eq:entropy_condition}
    a\geq Cn^{-\frac{1}{2}} \bigg(\int^R_0 \CalH^{\frac{1}{2}}(u,\ScrG,\|\cdot\|_n)du\vee R\bigg),
\end{equation}
then we have
\begin{equation}
    \label{eq:geer_modulus_continuity}
    \pr\bigg(\sup_{g\in\ScrG}\abs{\langle g, W \rangle_n}\geq a\bigg)\leq 2 \exp\biggl(-\tilde{C}\frac{na^2}{\sigma^2R^2}\biggr),
\end{equation}
for some positive $\tilde{C}$.
\end{lemma}
\proof{Proof.}
See Corollary~8.3 in \cite{vanderGeer00_ec}.
\Halmos\endproof

\begin{lemma}\label{lem:Modulus-Continuity}

Let $k$ be a kernel satisfying Assumption~\ref{assump:SL}.
Suppose that $\{\BFx_i\}_{i=1}^n$ form a TSG  $\CalX^{\mathsf{TSG}}_{n}$ and  $W\coloneqq\{W_i\}_{i=1}^n$ are independent  zero-mean $\subG(\sigma^2)$ random variables.
Then,
for all $t$ large enough,
\begin{equation}\label{eq:Modulus-of-Continuity}
        \sup_{g\in\ScrH_k}\frac{\sigma^{-1} n^{\frac{1}{2}}\abs{\langle g,W\rangle_n}}{\sqrt{\|g\|_n\|g\|_{\ScrH_k}} \Bigl|\log\frac{\|g\|_n}{\|g\|_{\ScrH_k}}\Bigr|^{\frac{2d-1}{4}}} > t ,
\end{equation}
with probability at most $C_2\exp(-C_1t^2)$ with some positive constants $C_1$ and $C_2$.
\end{lemma}
\proof{Proof.}

Let $\ScrF\coloneqq\{g:\|g\|_{\ScrH_k}\leq 1\}$ be the unit ball in $\ScrH_k$.
Notice that the empirical semi-norm $\|\cdot\|_n$ is upper bounded by the $L^\infty$ norm $\|\cdot\|_\infty$. Therefore, the metric entropy associated to the empirical semi-norm is upper bounded by the one associated to the $L^\infty$ norm:
\[\CalH(\epsilon,\ScrF,\|\cdot\|_{n})\leq \CalH(\epsilon,\ScrF,\|\cdot\|_{\infty})\leq C\epsilon^{-1}\bigl|\log \epsilon|^{\frac{2d-1}{2}},\]
for some positive constant $C$.
Hence,
\begin{align*}
    \int_0^{R}\left(\CalH(u,\ScrF,\|\cdot\|_{n})\right)^{\frac{1}{2}}\dd{u}\leq C\int_0^R u^{-1/2}\bigl|\log u\bigr|^{d/2-1/4}\dd{u}\leq C R^{\frac{1}{2}}\bigl|\log R\bigr|^{\frac{2d-1}{4}}.
\end{align*}

One may readily check that
for all $R<1$ and $a\geq n^{-1/2}\sigma R^{\frac{1}{2}}\bigl|\log R\bigr|^{\frac{2d-1}{4}}$, the condition \eqref{eq:entropy_condition} is met,
and thus by \eqref{eq:geer_modulus_continuity}, we have
\begin{align}
  \pr\biggl(\sup_{g\in\ScrF,\|g\|_n\leq R}\sigma^{-1}n^{\frac{1}{2}}\abs{\langle h,W\rangle_n}\geq R^{\frac{1}{2}}|\log R|^{\frac{2d-1}{4}} \biggr)&=\pr\biggl(\sup_{g\in\ScrF,\|g\|_n\leq R}\bigl|\langle f,W\rangle_n\bigr|\geq a\biggr) \nonumber \\
  &\leq 2\exp(-C_1\frac{na^2}{\sigma^2 R^2}),\label{eq:prob-empirical-inner-prod}
\end{align}
for some positive constant $C_1$.

Let $h\coloneqq g/\|g\|_{\ScrH_k}$. Then, the left-hand-side of the inequality \eqref{eq:Modulus-of-Continuity} becomes
\[\sup_{h\in\ScrF}\frac{\sigma^{-1}\abs{\langle h,W\rangle_n}}{\sqrt{\|h\|_n}\bigl|\log \|h\|_n\bigr|^{\frac{2d-1}{4}}}.\]
We then can give the following upper bound for its tail distribution:
\begin{align*}
    \quad{} & \pr\Biggl(\sup_{h\in\ScrF}\frac{\sigma^{-1}n^{\frac{1}{2}}\abs{\langle h,W\rangle_n}}{\sqrt{\|h\|_n}\bigl|\log \|h\|_n\bigr|^{\frac{2d-1}{4}}}> t \Biggr)\\
    ={} & \pr\Biggl(\bigcup_{i=0}^{\infty} \biggl\{\sup_{h\in\ScrF,\|h\|_n\in(2^{-i-1}, 2^{-i}]}\frac{\sigma^{-1} n^{\frac{1}{2}}\abs{\langle h,W\rangle_n}}{\sqrt{\|h\|_n}\bigl|\log \|h\|_n\bigr|^{\frac{2d-1}{4}}}> t \biggr\}\Biggr)\\
    \leq{} & \pr\Biggl(\bigcup_{i=0}^{\infty} \biggl\{\sup_{h\in\ScrF,\|h\|_n\in(2^{-i-1}, 2^{-i}]}\sigma^{-1} n^{\frac{1}{2}}\abs{\langle h,W\rangle_n}> t 2^{-\frac{i+1}{2}}(i+1)^{\frac{2d-1}{4}}\biggr\}\Biggr)\\
    \leq{} & \sum_{i=0}^\infty\pr\biggl(\sup_{h\in\ScrF, \|h\|_n\leq 2^{-i}}n^{\frac{1}{2}}\sigma^{-1}\abs{\langle h,W\rangle_n}> t2^{-\frac{i+1}{2}}i^{\frac{2d-1}{4}}\biggr)\\
    \leq{} & \sum_{i=0}^{\infty}2\exp(-C_1t^2 2^{i}i^{\frac{2d-1}{2}})\\
    \leq {} & \sum_{i=1}^\infty 2\bigl(\exp(-C_1t^2)\bigr)^i\\
    \leq{} & C_2 \exp(-C_1 t^2),
\end{align*}
where the fifth line follows from applying \eqref{eq:prob-empirical-inner-prod}  with $R=2^{-i}$ and $a=tn^{-1/2}\sigma 2^{-\frac{i+1}{2}}i^{\frac{2d-1}{4}}$.
\Halmos \endproof

\begin{remark}\label{remark:sub-Gaussian}
Lemma~\ref{lem:Modulus-Continuity} shows that the left-hand-side of the inequality \eqref{eq:Modulus-of-Continuity} is a positive sub-Gaussian random variable.
\end{remark}

\subsection{Connecting Empirical Semi-norm and Other Norms}

\begin{lemma}\label{lem:norming-ineqality}
Let $k$ be a kernel satisfying Assumption~\ref{assump:SL}.
Suppose that $f\in\ScrH_k$ and $\{\BFx_i\}_{i=1}^n$ form a TSG  $\CalX^{\mathsf{TSG}}_{n}$
Then, there exists a positive constant $C$ such that
\[\|f\|_{2}\leq C\left((\log n)^{\frac{3(d-1)}{2}}\|f\|_n+n^{-1}(\log n)^{2(d-1)}\|f\|_{\ScrH_k}\right).\]
\end{lemma}
\proof{Proof.}
Let $\tau$ satisfy the condition that $\abs{\CalX^{\mathsf{SG}}_{\tau}}\leq n< \abs{\CalX^{\mathsf{SG}}_{\tau+1}}$. According to Lemma~\ref{lem:number-SG}, we can easily check that $\abs{\CalX^{\mathsf{SG}}_{\tau}}$ and $\abs{\CalX^{\mathsf{SG}}_{\tau}}$ have the same order. Therefore, we can assume $\abs{\CalX^{\mathsf{SG}}_{\tau}}=n$ without loss of generality.

Let $\breve{f}_n$ be the KI estimator \eqref{eq:Kriging-ec} with noise-free observations of $f$ on $\CalX^{\mathsf{SG}}_{\tau}$. It follows from triangular inequality that:
\[\|f\|_2\leq\|f-\breve{f}_n\|_{2}+\|\breve{f}_n\|_2.\]
From Lemma~\ref{lemma:orthogonalExpansion}, we can expand $f-\breve{f}_n$ as follows:
\begin{align*}
    f-\breve{f}_n={}& \sum_{\abs{\BFl}> \tau+d-1}\sum_{\BFi\in\rho(\BFl)}\langle f,\phi_{\BFl,\BFi}\rangle_{\ScrH_k}\frac{\phi_{\BFl,\BFi}}{\|\phi_{\BFl,\BFi}\|_{\ScrH_k}^2},
\end{align*}
and we can have the following upper bound for the $L^2$ norm of the following summation:
\begin{align*}
    \bigg\|\sum_{\BFi\in\rho(\BFl)}\langle f,\phi_{\BFl,\BFi}\rangle_{\ScrH_k}\frac{\phi_{\BFl,\BFi}}{\|\phi_{\BFl,\BFi}\|_{\ScrH_k}^2}\bigg\|_2^2
    ={} &  \bigg\|\sum_{\BFi\in\rho(\BFl)}\bigl\langle f\big|_{\supp[\phi_{\BFl,\BFi}]},\phi_{\BFl,\BFi}\bigr\rangle_{\ScrH_k}\frac{\phi_{\BFl,\BFi}}{\|\phi_{\BFl,\BFi}\|_{\ScrH_k}^2}\bigg\|_2^2\\
    ={} & \sum_{\BFi\in\rho(\BFl)}\bigl\langle f\big|_{\supp[\phi_{\BFl,\BFi}]},\phi_{\BFl,\BFi}\bigr\rangle_{\ScrH_k}^2\frac{\|\phi_{\BFl,\BFi}\|_2^2}{\|\phi_{\BFl,\BFi}\|_{\ScrH_k}^4}\\
    \leq{} & C_12^{-3\abs{\BFl}}\sum_{\BFi\in\rho(\BFl)}\bigl\langle f\big|_{\supp[\phi_{\BFl,\BFi}]},\phi_{\BFl,\BFi}\bigr\rangle_{\ScrH_k}^2\\
    \leq{} &C_12^{-3\abs{\BFl}}\sum_{\BFi\in\rho(\BFl)}\big\| f\big|_{\supp[\phi_{\BFl,\BFi}]}\big\|_{\ScrH_k}^2\|\phi_{\BFl,\BFi}\|_{\ScrH_k}^2\\
    \leq {} & C_22^{-2\abs{\BFl}}\|f\|_{\ScrH_k}^2,
\end{align*}
where the second line follows from the fact that the supports of basis function $\|\phi_{\BFl,\BFi}:\BFi\in\rho(\BFl)\|$ form a partition of $\ScrX$ and are mutually disjoint, the third line follows from the estimate of $\|\phi_{\BFl,\BFi}\|_2$ and $\|\phi_{\BFl,\BFi}\|_{\ScrH_k}$ in Lemma~\ref{lemma:norm_estimate}, and  the fourth line follows from the Cauchy–Schwarz inequality.

So we can use Lemma~\ref{lem:bungartz-Lemma-3-7} to get the estimate for the $L^2$ norm of $f-\breve{f}_{n}$:
\begin{align*}
     \|f-\breve{f}_n\|_2={}& \bigg\|\sum_{\abs{\BFl}> \tau+d-1}\sum_{\BFi\in\rho(\BFl)}\langle f,\phi_{\BFl,\BFi}\rangle_{\ScrH_k}\frac{\phi_{\BFl,\BFi}}{\|\phi_{\BFl,\BFi}\|_{\ScrH_k}^2}\bigg\|_2\\
     \leq {} &\sum_{\abs{\BFl}> \tau+d-1}\bigg\|\sum_{\BFi\in\rho(\BFl)}\langle f,\phi_{\BFl,\BFi}\rangle_{\ScrH_k}\frac{\phi_{\BFl,\BFi}}{\|\phi_{\BFl,\BFi}\|_{\ScrH_k}^2}\bigg\|_2\\
     \leq{} & \sqrt{C_2}\|f\|_{\ScrH_k}\sum_{\abs{\BFl}> \tau+d-1}2^{-\abs{\BFl}}\\
     \leq{} & C_3 2^{-\tau}\tau^{d-1}\|f\|_{\ScrH_k}.
\end{align*}
On the other hand, according to Algorithm 1 in \cite{Plumlee14_ec}, the KI estimator $\breve{f}_n$  can be represented as linear combination of KI estimators conditioned on observations on full grid:
\begin{equation}\label{eq:SG-FG-relation}
    \breve{f}_n(\BFx)=\sum_{\tau\leq\abs{\BFl}\leq \tau+d-1}(-1)^{\tau+d-1-\abs{\BFl}}\binom{d-1}{\tau+d-1-\abs{\BFl}}\breve{f}^{\mathsf{FG}}_{\BFl}(\BFx).
\end{equation}
We first prove a relation between $\|\breve{f}^{\mathsf{FG}}_{\BFl}\|_n$ and $\|\breve{f}^{\mathsf{FG}}_{\BFl}\|_2$ when the observations are on full grid $\CalX^{\mathsf{FG}}_{\BFl}$. From equation \eqref{eq:FGexpansion}, we can see that
\begin{align}
   \|\breve{f}^{\mathsf{FG}}_{\BFl}\|_2^2\nonumber
   ={} &\int_{\ScrX}\biggl|\sum_{(\BFl,\BFi):\BFc_{\BFl,\BFi}\in\CalX^{\mathsf{FG}}_{\BFl}}f(\BFc_{\BFl,\BFi})\phi_{\BFl,\BFi}(\BFx)\biggr|^2\dd{\BFx}\nonumber\\
    ={} &
    \sum_{(\BFl,\BFi):\BFc_{\BFl,\BFi}\in\CalX^{\mathsf{FG}}_{\BFl}}\sum_{\BFx_{\BFl,\BFi'}\in\mathsf{Ne}(\BFl,\BFi)}f(\BFc_{\BFl,\BFi})f(\BFx_{\BFl,\BFi'})\int_{\ScrX}\phi_{\BFl,\BFi}(\BFx)\phi_{\BFl,\BFi'}(\BFx)\dd{\BFx},\label{eq:FG-squared-decomp}
\end{align}
where
\[\mathsf{Ne}(\BFl,\BFi)\coloneqq\{\BFx_{\BFl,\BFi'}: i'_j\in\{i_j-1,i_j,i_j+1\},j=1,\ldots,d\}.\]
The second term of equation \eqref{eq:FG-squared-decomp} is from the fact that if supports of two basis function $\phi_{\BFl,\BFi}$ and $\phi_{\BFl,\BFi'}$ are disjoint, then the product  $\phi_{\BFl,\BFi}\phi_{\BFl,\BFi'}$ is identically 0. As a result, for each index $(\BFl,\BFi)$, only basis functions with supports that overlap with the support of $\phi_{\BFl,\BFi}$ are left in the summation. From the definition of $\phi_{\BFl,\BFi}$, we can see that basis functions with supports that overlaps with the support of $\phi_{\BFl,\BFi}$ are all the basis functions centered at the neighboring points of $\BFc_{\BFl,\BFi}$.

From direct calculation, for any pair $\BFi,\BFi'$, we can also have
\begin{equation}\label{eq:phi-L2-estimate}
    \int_{\ScrX}\phi_{\BFl,\BFi}(\BFx)\phi_{\BFl,\BFi'}(\BFx)\dd{\BFx}\leq C_42^{-\abs{\BFl}},
\end{equation}
where $C_4$ is some constant. It follows that
\begin{align}
    \quad &\sum_{(\BFl,\BFi):\BFc_{\BFl,\BFi}\in\CalX^{\mathsf{FG}}_{\BFl}}\sum_{\BFx_{\BFl,\BFi'}\in\mathsf{Ne}(\BFl,\BFi)}f(\BFc_{\BFl,\BFi})f(\BFx_{\BFl,\BFi'})\int_{\ScrX}\phi_{\BFl,\BFi}(\BFx)\phi_{\BFl,\BFi'}(\BFx)\dd{\BFx}\nonumber\\
    \leq{} & C_4 \sum_{(\BFl,\BFi):\BFc_{\BFl,\BFi}\in\CalX^{\mathsf{FG}}_{\BFl}}\sum_{\BFx_{\BFl,\BFi'}\in\mathsf{Ne}(\BFl,\BFi)}f(\BFc_{\BFl,\BFi})f(\BFx_{\BFl,\BFi'})2^{-\abs{\BFl}}\nonumber\\
    \leq{} & C_42^{-\abs{\BFl}-1}\sum_{(\BFl,\BFi):\BFc_{\BFl,\BFi}\in\CalX^{\mathsf{FG}}_{\BFl}}\sum_{\BFx_{\BFl,\BFi'}\in\mathsf{Ne}(\BFl,\BFi)}\bigl(f^2(\BFc_{\BFl,\BFi})+f^2(\BFx_{\BFl,\BFi'})\bigr)\nonumber\\
    ={}& C_42^{-\abs{\BFl}-1}\Biggl[ \sum_{(\BFl,\BFi):\BFc_{\BFl,\BFi}\in\CalX^{\mathsf{FG}}_{\BFl}}3^df^2(\BFc_{\BFl,\BFi})+ \sum_{(\BFl,\BFi):\BFc_{\BFl,\BFi}\in\CalX^{\mathsf{FG}}_{\BFl}}\sum_{\BFx_{\BFl,\BFi'}\in\mathsf{Ne}(\BFl,\BFi)}f^2(\BFx_{\BFl,\BFi'}) \Biggr],\label{eq:FG-interaction}
\end{align}
where the third line is from the
inequality of arithmetic and geometric means and the fourth line is from the fact that the size of $\mathsf{Ne}(\BFl,\BFi)$ is $3^d$. Moreover,
\begin{align*}
    \sum_{(\BFl,\BFi):\BFc_{\BFl,\BFi}\in\CalX^{\mathsf{FG}}_{\BFl}}\sum_{\BFx_{\BFl,\BFi'}\in\mathsf{Ne}(\BFl,\BFi)}f^2(\BFx_{\BFl,\BFi'})={} & \sum_{(\BFl,\BFi):\BFc_{\BFl,\BFi}\in\CalX^{\mathsf{FG}}_{\BFl}}\sum_{(\BFl',\BFi'):\BFx_{\BFl',\BFi'}\in\CalX^{\mathsf{FG}}_{\BFl}}\boldsymbol{1}_{\{\BFc_{\BFl,\BFi}\in\mathsf{Ne}(\BFl',\BFi')\}}f^2(\BFc_{\BFl,\BFi})\\
    \leq{} & 3^{d}\sum_{(\BFl,\BFi):\BFc_{\BFl,\BFi}\in\CalX^{\mathsf{FG}}_{\BFl}} f^2(\BFc_{\BFl,\BFi}),
\end{align*}
where the second line is from the following reasoning. For each $\BFc_{\BFl,\BFi}$, the number of its neighboring point is at most $3^d$. Therefore, for each $\BFc_{\BFl,\BFi}$, it can also be the neighboring point of at most $3^d$ distinct points. If we sum over all the $\{f^2(\BFc_{\BFl,\BFi}):\BFx_{\BFl,\BFi\in\CalX^{\mathsf{FG}}_{\BFl}}\}$ on each support, then there must be at most $3^d$  neighboring sets $\{f^2\bigl(\mathsf{Ne}({\BFl,\BFi})\bigr)\}$ that can share the same $f(\BFc_{\BFl,\BFi})$. Hence, the sum must be bounded by the sum of all $\{f(\BFc_{\BFl,\BFi})\}$ multiplied by $3^d$.

We then can substitute the above estimate into \eqref{eq:FG-interaction} to get:
\begin{align}
    \quad &\sum_{(\BFl,\BFi):\BFc_{\BFl,\BFi}\in\CalX^{\mathsf{FG}}_{\BFl}}\sum_{\BFx_{\BFl,\BFi'}\in\mathsf{Ne}(\BFl,\BFi)}f(\BFc_{\BFl,\BFi})f(\BFx_{\BFl,\BFi'})\int_{\ScrX}\phi_{\BFl,\BFi}(\BFx)\phi_{\BFl,\BFi'}(\BFx)\dd{\BFx}\nonumber \\
    \leq{} &  C_42^{-\abs{\BFl}-1}\Biggl[ \sum_{(\BFl,\BFi):\BFc_{\BFl,\BFi}\in\CalX^{\mathsf{FG}}_{\BFl}}3^df^2(\BFc_{\BFl,\BFi})+ \sum_{(\BFl,\BFi):\BFc_{\BFl,\BFi}\in\CalX^{\mathsf{FG}}_{\BFl}}3^df^2(\BFc_{\BFl,\BFi}) \Biggr]\nonumber \\
    ={} & C_52^{-\abs{\BFl}}\sum_{(\BFl,\BFi):\BFc_{\BFl,\BFi}\in\CalX^{\mathsf{FG}}_{\BFl}}f^2(\BFc_{\BFl,\BFi}).\label{eq:FG-interaction-estimate}
\end{align}
Notice that the number of point in FG $\CalX^{\mathsf{FG}}_{\BFl}$ is exactly $2^{\abs{\BFl}}$. Putting \eqref{eq:FG-squared-decomp},\eqref{eq:phi-L2-estimate} and \eqref{eq:FG-interaction-estimate} together, we can get
\begin{equation}\label{eq:FG-L2-empirical}
    \|\breve{f}^{\mathsf{FG}}_{\BFl}\|_2^2\leq C_6\|\breve{f}^{\mathsf{FG}}_{\BFl}\|_n^2.
\end{equation}
for some constant $C_6$ independent of $f$. We then use equation \eqref{eq:SG-FG-relation} to get
\begin{align*}
     \|\breve{f}_n\|_2^2={} &\int_\ScrX\biggl|\sum_{\tau\leq\abs{\BFl}\leq \tau+d-1}(-1)^{\tau+d-1-\abs{\BFl}}\binom{d-1}{\tau+d-1-\abs{\BFl}}\breve{f}^{\mathsf{FG}}_{\BFl}(\BFx)\biggr|^2\dd(\BFx)\\
     \leq{} & \int_\ScrX\sum_{\tau\leq\abs{\BFl}\leq \tau+d-1}\sum_{\tau\leq\abs{\BFl'}\leq \tau+d-1}\binom{d-1}{\tau+d-1-\abs{\BFl}}\binom{d-1}{\tau+d-1-\abs{\BFl'}}\bigl|\breve{f}^{\mathsf{FG}}_{\BFl}(\BFx)\breve{f}^{\mathsf{FG}}_{\BFl'}(\BFx)\bigr|\dd(\BFx)\\
     \leq{} & \int_\ScrX\sum_{\tau\leq\abs{\BFl}\leq \tau+d-1}\sum_{\tau\leq\abs{\BFl'}\leq \tau+d-1}\binom{d-1}{\tau+d-1-\abs{\BFl}}\binom{d-1}{\tau+d-1-\abs{\BFl'}}\frac{\bigl|\breve{f}^{\mathsf{FG}}_{\BFl}(\BFx)\bigr|^2+\bigl|\breve{f}^{\mathsf{FG}}_{\BFl'}(\BFx)\bigr|^2}{2}\dd(\BFx)\\
     ={} & \Biggl[\sum_{\tau\leq\abs{\BFl'}\leq \tau+d-1}\binom{d-1}{\tau+d-1-\abs{\BFl'}}\Biggr]\sum_{\tau\leq\abs{\BFl}\leq \tau+d-1}\binom{d-1}{\tau+d-1-\abs{\BFl}} \|\breve{f}^{\mathsf{FG}}_{\BFl}\|_2^2\\
     \leq{} & \Biggl[\sum_{\tau\leq\abs{\BFl'}\leq \tau+d-1}\binom{d-1}{\tau+d-1-\abs{\BFl'}}\Biggr]C_6\sum_{\tau\leq\abs{\BFl}\leq \tau+d-1}\binom{d-1}{\tau+d-1-\abs{\BFl}}  2^{-\abs{\BFl}}\sum_{(\BFl,\BFi):\BFc_{\BFl,\BFi}\in\CalX^{\mathsf{FG}}_{\BFl}}f^2(\BFc_{\BFl,\BFi}),
\end{align*}
where the third line is from the
inequality of arithmetic and geometric means and the last line is from \eqref{eq:FG-L2-empirical}.
Note that
\begin{align}
    \sum_{\tau\leq\abs{\BFl'}\leq{}\tau+d-1}\binom{d-1}{\tau+d-1-\abs{\BFl'}}={} &\sum_{\ell=\tau}^{\tau+d-1}\binom{\ell-1}{ d-1}\binom{d-1}{\tau+d-1-\ell}\nonumber\\
    \leq{} & \max_{1\leq l\leq d}\binom{d-1}{\ell}\sum_{\ell=\tau}^{\tau+d-1}\binom{\ell-1}{ d-1}\nonumber\\
    \leq{} & C_7 \tau^{d-1},\label{eq:first-term-SG_empirical}
\end{align}
where the last line can be shown via Stirling's approximation for factorials. Moreover
\begin{align}
    \quad{} & \sum_{\tau\leq\abs{\BFl}\leq \tau+d-1}\binom{d-1}{\tau+d-1-\abs{\BFl}}  2^{-\abs{\BFl}}\sum_{(\BFl,\BFi):\BFc_{\BFl,\BFi}\in\CalX^{\mathsf{FG}}_{\BFl}}f^2(\BFc_{\BFl,\BFi})\nonumber\\
    \leq {} & \max_{1\leq l\leq d}\binom{d-1}{\ell}2^{-\tau}\sum_{\tau\leq\abs{\BFl}\leq \tau+d-1}\sum_{(\BFl,\BFi):\BFc_{\BFl,\BFi}\in\CalX^{\mathsf{FG}}_{\BFl}}f^2(\BFc_{\BFl,\BFi})\nonumber\\
    \leq{} & C_8 2^{-\tau}\tau^{d-1} \sum_{\BFc_{\BFl,\BFi}\in\CalX^{\mathsf{SG}}_{\tau}}f^2(\BFc_{\BFl,\BFi}),\label{eq:second-term-SG_empirical}
\end{align}
where the last time is from the fact that any point $\BFc_{\BFl,\BFi}\in\CalX^{\mathsf{SG}}_{\tau}$ can appear at most on $\sum_{\tau\leq\abs{\BFl}\leq \tau+d-1}1=\CalO(\tau^{d-1})$ different FG's. As a result, \eqref{eq:first-term-SG_empirical} and \eqref{eq:second-term-SG_empirical} gives:
\[ \|\breve{f}_n\|_2^2\leq C_7C_8\tau^{2(d-1)}2^{-\tau}\sum_{\BFc_{\BFl,\BFi}\in\CalX^{\mathsf{SG}}_{\tau}}f^2(\BFc_{\BFl,\BFi})\leq C_9\tau^{3(d-1)}\|f\|_n^2,\]
where the last inequality is from Lemma~\ref{lem:number-SG} which states that total number of point if $ {\ScrX}^{\mathsf{SG}}_{\tau}$ is $n=\CalO(2^{\tau}\tau^{d-1})$.
It follows that:
\[\|f\|_2\leq\|f-\breve{f}_n\|_{2}+\|\breve{f}_n\|_2\leq C_3 2^{-\tau}\tau^{d-1}\|f\|_{\ScrH_k}+\sqrt{C_9}\tau^{\frac{3(d-1)}{2}}\|f\|_n.\]
We can substitute the identity $n=\CalO(2^{\tau}\tau^{d-1})$ into the above equation to get the final result.
\Halmos \endproof

\subsection{Convergence Rate of Kernel Ridge Regression}
\begin{proposition}
\label{prop:Convergence_KRR}
Let $k$ be a kernel satisfying Assumption~\ref{assump:SL}.
Suppose that $f\in\ScrH^{m}_{\mathsf{mix}}$ with $m=1,2$, Assumption~\ref{assump:optimum} holds, and Assumption~\ref{assump:sub-Gaussian} holds with $\sigma>0$.
Let  $\widehat{f}_{n,\lambda}$ be the KRR estimator \eqref{eq:KRR-solution-ec} on a TSG $\CalX^{\mathsf{TSG}}_{n}$.
If $\lambda\asymp \sigma^{\frac{4}{2m+1}}n^{-\frac{2}{2m+1}}|\log (\sigma n)|^{\frac{2d-1}{2m+1}}(\log n)^{\frac{6(m-1)(1-d)}{5}}$, then
\begin{align*}
    &\|f-\widehat{f}_{n,\lambda}\|_n=\CalO_p\left(\sigma^{\frac{2m}{2m+1}}n^{-\frac{m}{2m+1}}|\log (\sigma n)|^{\frac{2d-1}{7-m}}(\log n)^{\frac{3m-3}{10}(d-1)}\right),\\
    & \|f-\widehat{f}_{n,\lambda}\|_{\ScrH_k}=\CalO_p\left(\sigma^{\frac{2(m-1)}{5}}n^{-\frac{m-1}{5}}|\log (\sigma n)|^{\frac{(m-1)(2d-1)}{10}}(\log n)^{\frac{9(m-1)(d-1)}{10}}\right),\\
    &\|f-\widehat{f}_{n,\lambda}\|_\infty=\CalO_p\Bigl(\sigma^{\frac{2m-1}{2m+1}}n^{-\frac{2m-1}{4m+2}}|\log (\sigma n)|^{\frac{(2m-1)(2d-1)}{4(2m+1)}}(\log n)^{\frac{(6m-3)(d-1)}{6m-2}}\Bigr).
\end{align*}
\end{proposition}
\proof{Proof.}
It follows from Proposition~\ref{prop:SL-norm-equivalence} and
Proposition~\ref{prop:H-2-mix} that  $f\in\ScrH_{k}^m$ for $m=1,2$.

Because $\widehat{f}_{n,\lambda}$ is the optimizer of the optimization problem \eqref{eq:KRR-solution-ec}, we can derive that
\begin{equation}\label{eq:optimizer-condition}
    \frac{1}{n}\sum_{\BFc_{\BFl,\BFi}\in \CalX^{\mathsf{TSG}}_{n} }\big|y(\BFc_{\BFl,\BFi})-\widehat{f}_{n,\lambda}(\BFc_{\BFl,\BFi})\big|^2+\lambda \|\widehat{f}_{n,\lambda}\|^2_{\ScrH_k}\leq \frac{1}{n}\sum_{\BFc_{\BFl,\BFi}\in \CalX^{\mathsf{TSG}}_{n} }\big|y(\BFc_{\BFl,\BFi})-f(\BFc_{\BFl,\BFi})\big|^2+\lambda \|f\|^2_{\ScrH_k}.
\end{equation}
We  rearrange  \eqref{eq:optimizer-condition} as:
 \begin{align}
     &\|\widehat{f}_{n,\lambda}-f\|_n^2+\lambda\|\widehat{f}_{n,\lambda}\|^2_{\ScrH_k} \nonumber\\
     \leq{} & 2\langle \varepsilon,\widehat{f}_{n,\lambda}-f\rangle_n+\lambda\|f\|^2_{\ScrH_k}\nonumber\\
     \leq{} & C_1\sigma n^{-\frac{1}{2}}\sqrt{\|\widehat{f}_{n,\lambda}-f\|_n\|\widehat{f}_{n,\lambda}-f\|_{\ScrH_k}}  \biggl|\log\frac{\|\widehat{f}_{n,\lambda}-f\|_n}{\|\widehat{f}_{n,\lambda}-f\|_{\ScrH_k}}\biggr|^{\frac{2d-1}{4}}+\lambda\|f\|^2_{\ScrH_k},\label{eq:rearrange-1}
 \end{align}
where the second inequality is from Lemma \ref{lem:Modulus-Continuity},
and $C_1$ is some positive sub-Gaussian random variable (see Remark~\ref{remark:sub-Gaussian}).

\textbf{Case (i): $f\in\ScrH_k$ but $f\not\in\ScrH_k^2$.}
Note that
\begin{align}
    \|f\|^2_{\ScrH_k}-\|\widehat{f}_{n,\lambda}\|^2_{\ScrH_k}={} &\langle f-\widehat{f}_{n,\lambda},f+\widehat{f}_{n,\lambda}\rangle_{\ScrH_k}\nonumber\\
    ={} &2 \langle f,f-\widehat{f}_{n,\lambda}\rangle_{\ScrH_k}-\|f-\widehat{f}_{n,\lambda}\|^2_{\ScrH_k}\nonumber\\
    \leq{}& 2\|f\|_{\ScrH_k}\| \widehat{f}_{n,\lambda}-f\|_{\ScrH_k}-\|\widehat{f}_{n,\lambda}-f\|^2_{\ScrH_k}\label{eq:f-f_hat-rkhs-difference}.
\end{align}
Therefore \eqref{eq:rearrange-1} can be further rearranged as
\begin{align}
    \quad{} & \|\widehat{f}_{n,\lambda}-f\|_n^2+\lambda\|\widehat{f}_{n,\lambda}-f\|^2_{\ScrH_k}\nonumber\\
    \leq{} & \underbrace{2\lambda\|f\|_{\ScrH_k}\| \widehat{f}_{n,\lambda}-f\|_{\ScrH_k}}_{I_1}+\underbrace{C_1\sigma n^{-\frac{1}{2}}\sqrt{\|\widehat{f}_{n,\lambda}-f\|_n\|\widehat{f}_{n,\lambda}-f\|_{\ScrH_k}} \biggl|\log\frac{\|\widehat{f}_{n,\lambda}-f\|_n}{\|\widehat{f}_{n,\lambda}-f\|_{\ScrH_k}}\biggr|^{\frac{2d-1}{4}}}_{I_2}.\label{eq:rearrange-2}
\end{align}
For notation simplicity, let $P= \|\widehat{f}_{n,\lambda}-f\|_n$ and $Q=\|\widehat{f}_{n,\lambda}-f\|_{\ScrH_k}$. From \eqref{eq:rearrange-2}, we can derive that $P^2+\lambda Q^2$ is less than or equal either $2I_1$ or $2I_2$. In the case $P^2+\lambda Q^2\leq 2I_1$, we have:
\begin{align}
    P\leq C_2\sqrt{\lambda}\|f\|_{\ScrH_k}\qq{and}
    Q\leq C_2 \|f\|_{\ScrH_k},\label{eq:case-A}
\end{align}
where $C_2=4$. Note that
\[\|\widehat{f}_{n,\lambda}-f\|_n\leq \|\widehat{f}_{n,\lambda}-f\|_\infty\leq \max_{\BFx_\in\ScrX}\sqrt{k(\BFx,\BFx)}\|\widehat{f}_{n,\lambda}-f\|_{\ScrH_k},\]
so both $P$ and $Q$ can be upper bounded if we select $\lambda$ correctly and
\[|\log(PQ^{-1})|\leq C_3 |\log P |\]
for some constant $C_3$. Therefore, in the case $P^2+\lambda Q^2\leq 2I_2$, the inequality can be rewritten as
\[P^2+\lambda Q^2\leq 2C_1C_3\sigma n^{-\frac{1}{2}}\sqrt{PQ}\bigl|\log P\bigr|^{\frac{2d-1}{4}},\]
which leads to
\begin{align*}
    &P^{\frac{3}{2}}\leq C_4\sigma n^{-\frac{1}{2}}\sqrt{Q}\bigl|\log P\bigr|^{\frac{2d-1}{4}},\\
    &Q^{\frac{3}{2}}\leq C_4\sigma\lambda ^{-1}n^{-\frac{1}{2}}\sqrt{P}\bigl|\log P\bigr|^{\frac{2d-1}{4}},
\end{align*}
where $C_4=2C_1C_3$. We then  can solve the above equations to get:
\begin{equation}
    \label{eq:case-B}
\begin{aligned}
    & P\leq C_5 \sigma n^{-\frac{1}{2}}\lambda^{-\frac{1}{4}}\bigl|\log (\sigma n)^{-\frac{1}{2}}\lambda^{-\frac{1}{4}} \bigr|^{\frac{2d-1}{4}} ,\\
    & Q\leq C_6 \sigma  n^{-\frac{1}{2}}\lambda ^{-\frac{3}{4}}\bigl|\log (\sigma n)^{-\frac{1}{2}}\lambda^{-\frac{1}{4}} \bigr|^{\frac{2d-1}{4}} ,
\end{aligned}
\end{equation}
where $C_5$ and $C_6$ are some constants. As a result, we can summarize \eqref{eq:case-A} and \eqref{eq:case-B} as
\begin{align*}
    & P\leq \max\biggl\{C_2\sqrt{\lambda}\|f\|_{\ScrH_k},\  C_5 \sigma n^{-\frac{1}{2}}\lambda^{-\frac{1}{4}}\left|\log (\sigma n^{-\frac{1}{2}}\lambda^{-\frac{1}{4}}) \right|^{\frac{2d-1}{4}}\biggr\},\\
    & Q\leq \max\biggl\{C_2\|f\|_{\ScrH_k},\ C_6\sigma  n^{-\frac{1}{2}}\lambda ^{-\frac{3}{4}}\left|\log (\sigma n^{-\frac{1}{2}}\lambda^{-\frac{1}{4}} )\right|^{\frac{2d-1}{4}}\biggr\}.
\end{align*}
Set $\lambda\asymp \sigma^{\frac{4}{3}}n^{-\frac{2}{3}}(\log (\sigma n))^{\frac{2d-1}{3}}$, we get
\begin{align}
    & P=\|f-\widehat{f}_{n,\lambda}\|_n\leq \CalO_p\left(\sigma^{\frac{2}{3}} n^{-\frac{1}{3}}(\log (\sigma n))^{\frac{2d-1}{6}}\right),\label{eq:KRR-upper-bound-1-empirical}\\
    & Q=\|f-\widehat{f}_{n,\lambda}\|_{\ScrH_k}\leq \CalO_p\bigl(1\bigr).\label{eq:KRR-upper-bound-1-RKHS}
\end{align}
So we invoke Lemma \ref{lem:norming-ineqality} to get
\begin{align}
     \|f-\widehat{f}_{n,\lambda}\|_\infty\leq{} &C_7\|f-\widehat{f}_{n,\lambda}\|_2^{\frac{1}{2}}\nonumber\\
     \leq{} &  \left(C_8\bigl((\log n)^{\frac{3(d-1)}{2}}P+n^{-1}(\log n)^{2(d-1)}Q\bigr)\right)^{\frac{1}{2}}\nonumber\\
    \leq{} & \CalO_p\left( \sigma^{\frac{1}{3}} n^{-\frac{1}{6}}|\log (\sigma n)|^{\frac{2d-1}{12}}(\log n)^{\frac{3(d-1)}{4}}\right)\label{eq:KRR-upper-bound-1}.
\end{align}
This finishes the proof for the case $f\in\ScrH_k$ but $f\not\in\ScrH^2_k$.

\textbf{Case (ii): $f\in\ScrH^2_k$.}
According to our assumption, the RKHS inner product between $f$ and $g$ for any $g\in\ScrH_k$ can be written as
\[\langle f,g\rangle_{\ScrH_k}= \biggl\langle \prod_{j=1}^d\CalL_j[f],g\biggr\rangle_2\eqqcolon\langle w,g\rangle_2 \]
with
\[\|w\|^2_2\coloneqq\biggl\langle \prod_{j=1}^d\CalL_j[f],\prod_{j=1}^d\CalL_j[f]\biggr\rangle_2<\infty.\]

So we have a new form of equation \eqref{eq:f-f_hat-rkhs-difference}:
\begin{align}
    \|f\|^2_{\ScrH_k}-\|\widehat{f}_{n,\lambda}\|^2_{\ScrH_k}={} &\langle f-\widehat{f}_{n,\lambda},f+\widehat{f}_{n,\lambda}\rangle_{\ScrH_k}\nonumber\\
    ={} &2 \langle f,f-\widehat{f}_{n,\lambda}\rangle_{\ScrH_k}-\|f-\widehat{f}_{n,\lambda}\|^2_{\ScrH_k}\nonumber\\
    ={}& \langle w,f-\widehat{f}_{n,\lambda}\rangle_2-\|\widehat{f}_{n,\lambda}-f\|^2_{\ScrH_k}\nonumber\\
    \leq{} & \|w\|_2\|f-\widehat{f}_{n,\lambda}\|_2-\|\widehat{f}_{n,\lambda}-f\|^2_{\ScrH_k} \label{eq:f-f_hat-rkhs-difference-H2}.
\end{align}
Let $\bar{w}$ denote $\|w\|_2$. We can invoke \eqref{eq:f-f_hat-rkhs-difference-H2} and Lemma \ref{lem:norming-ineqality} to rewrite \eqref{eq:rearrange-1} as
\begin{align}
    \quad{} & \|\widehat{f}_{n,\lambda}-f\|_n^2+\lambda\|\widehat{f}_{n,\lambda}-f\|^2_{\ScrH_k}\nonumber\\
    \leq{} & \underbrace{\lambda\bar{w}\CalK_1(\log n)^{\frac{3(d-1)}{2}}\|\widehat{f}_{n,\lambda}-f\|_n}_{I_3}+ \underbrace{\lambda\bar{w}\CalK_1n^{-1}(\log n)^{2(d-1)}\|\widehat{f}_{n,\lambda}-f\|_{\ScrH_k}}_{I_4}\nonumber\\
    \quad{} & + \underbrace{C_1\sigma n^{-\frac{1}{2}}\sqrt{\|\widehat{f}_{n,\lambda}-f\|_n\|\widehat{f}_{n,\lambda}-f\|_{\ScrH_k}} \biggl|\log\frac{\|\widehat{f}_{n,\lambda}-f\|_n}{\|\widehat{f}_{n,\lambda}-f\|_{\ScrH_k}}\biggr|^{\frac{2d-1}{4}}}_{I_2},\label{eq:rearrange-3}
\end{align}
where $\CalK_1$ is some constant.

We keep the notations $P\coloneqq \|\widehat{f}_{n,\lambda}-f\|_n$ and $Q\coloneqq\|\widehat{f}_{n,\lambda}-f\|_{\ScrH_k}$ and split the equation \eqref{eq:rearrange-3} into three cases:
\begin{enumerate}[label=\emph{Case \arabic*.}, left=0pt]
\item  $I_3\geq I_4$.
\item $I_3\leq I_4$ and $Q\geq n^{-1}(\log n)^{2(d-1)}$.
\item $I_3\leq I_4$ and $Q\leq n^{-1}(\log n)^{2(d-1)}$.
\end{enumerate}
Notice that both Case 1 and Case 2 lead to
\[P^2+\lambda Q^2\leq \underbrace{\lambda\bar{w}\CalK_2(\log n)^{\frac{3(d-1)}{2}}P}_{I_5}+\underbrace{C_1\sigma n^{-\frac{1}{2}}\sqrt{PQ}\bigl|\log (PQ^{-1})\bigr|^{\frac{2d-1}{4}}}_{I_6}.\]
where $\CalK_2$ is some constant.

Case 3 implies that  the stochastic estimator $\widehat{f}_{n,\lambda}$ converges to the true function $f$ in a rate  faster than $n^{-\frac{1}{2}}$ but this violates central limit theorem so this case will never be true.

Similar to our previous analysis, $P^2+\lambda Q^2$ is less than or equal to either $2I_5$ or $2I_6$. For the case $P^2+\lambda Q^2\leq 2I_5$, we can immediately derive:
\begin{align*}
    & P\leq 2\lambda \bar{w}\CalK_2(\log n)^{\frac{3(d-1)}{2}},\\
    & Q\leq 2\sqrt{\lambda} \bar{w}\CalK_2(\log n)^{\frac{3(d-1)}{2}}.
\end{align*}
The case $P^2+\lambda Q^2\leq 2I_6$ has been solved in our previous analysis with its solution given in \eqref{eq:case-B}. Therefore, we have the following upper bounds:
\begin{align*}
    & P\leq \max\biggl\{2\lambda \bar{w}\CalK_2(\log n)^{\frac{3(d-1)}{2}},\  C_5 \sigma n^{-\frac{1}{2}}\lambda^{-\frac{1}{4}}\left|\log (\sigma n^{-\frac{1}{2}}\lambda^{-\frac{1}{4}}) \right|^{\frac{2d-1}{4}}\biggr\},\\
    & Q\leq \max\biggr\{2\sqrt{\lambda} \bar{w}\CalK_2(\log n)^{\frac{3(d-1)}{2}},\ C_6 \sigma n^{-\frac{1}{2}}\lambda ^{-\frac{3}{4}}\left|\log(\sigma n^{-\frac{1}{2}}\lambda^{-\frac{1}{4}})\right|^{\frac{2d-1}{4}}\biggr\}.
\end{align*}
Set $\lambda\asymp \sigma^{\frac{4}{5}}n^{-\frac{2}{5}}|\log (\sigma n)|^{\frac{2d-1}{5}}(\log n)^{\frac{6(1-d)}{5}} $ we get
\begin{align}
    &P=\|f-\widehat{f}_{n,\lambda}\|_n\leq \CalO_p\left(\sigma^{\frac{4}{5}}n^{-\frac{2}{5}}|\log (\sigma n)|^{\frac{2d-1}{5}}(\log n)^{\frac{3(d-1)}{10}}\right),\label{eq:KRR-upper-bound-2-empirical}\\
    &Q=\|f-\widehat{f}_{n,\lambda}\|_{\ScrH_k}\leq\CalO_p\left(\sigma^{2/5}n^{-1/5}|\log (\sigma n)|^{\frac{2d-1}{10}}(\log n)^{\frac{9(d-1)}{10}}\right).\label{eq:KRR-upper-bound-2-RKHS}
\end{align}
So we invoke Corollary~\ref{cor:Gagliardo–Nirenberg} as \eqref{eq:GN-inequality-case} and Lemma \ref{lem:norming-ineqality} to get
\begin{align}
      \|f-\widehat{f}_{n,\lambda}\|_\infty\leq {} & \CalK_3\sqrt{\|f-\widehat{f}_{n,\lambda}\|_2\|f-\widehat{f}_{n,\lambda}\|_{\ScrH_k}}\nonumber\\
      \leq{} & \CalK_3\sqrt{\left((\log n)^{\frac{3(d-1)}{2}}P+n^{-1}(\log n)^{2(d-1)}Q\right)Q}\nonumber\\
      ={} &\CalO_p\left(\sigma^{\frac{3}{5}}n^{-\frac{3}{10}}|\log (\sigma n)|^{\frac{3}{20}(2d-1)}(\log n)^{\frac{9(d-1)}{10}}\right).\label{eq:KRR-upper-bound-2}
\end{align}
This finishes the proof for $f\in\ScrH_k^2$.

Finally, we can put together \eqref{eq:KRR-upper-bound-1-empirical}, \eqref{eq:KRR-upper-bound-1-RKHS},  \eqref{eq:KRR-upper-bound-1},  \eqref{eq:KRR-upper-bound-2-empirical}, \eqref{eq:KRR-upper-bound-2-RKHS}, and \eqref{eq:KRR-upper-bound-2} and their associated $\lambda$ to get the final result.
\Halmos \endproof

\subsection{Proof of Theorem~\ref{thm:Convergence_EI_sum} }
\proof{Proof.}
It follows from Proposition~\ref{prop:SL-norm-equivalence} and
Proposition~\ref{prop:H-2-mix} that  $f\in\ScrH_{k}^m$ for $m=1,2$.

By \eqref{eq:decomp-norm},
it suffices to bound
$\|f-\widetilde{f}_N\|_\infty$, which can be  written as:
 \begin{align*}
     \|f-\widetilde{f}_N\|_\infty
      ={}  & \|\underbrace{(f-\widehat{f}_{N_\tau,\lambda})}_{U_1}-\underbrace{\widetilde{f}_N - \widehat{f}_{N_\tau,\lambda})}_{U_2}\|_\infty.
\end{align*}

We first calculate the RKHS norm of  term $U_1$. Because $\widehat{f}_{N_\tau,\lambda}$ is a KRR estimator with observations on the level-$\tau$ SG ${\ScrX}_{\tau}^{\mathsf{SG}}$ and penalty being specified as
\[\lambda\asymp \sigma^{\frac{4}{2m+1}}N_{\tau}^{-\frac{2}{2m+1}}|\log (\sigma N_\tau)|^{\frac{2d-1}{2m+1}}(\log N_\tau)^{\frac{6(m-1)(1-d)}{5}},\]
we can invoke
Proposition~\ref{prop:Convergence_KRR} to show that $U_1$ is upper bounded as follows:
\[\|f-\widehat{f}_{N_\tau,\lambda}\|_{\ScrH_k}=\CalO_p\left(\sigma^{\frac{2(m-1)}{5}}N_\tau^{-\frac{m-1}{5}}|\log (\sigma N_\tau)|^{\frac{(m-1)(2d-1)}{10}}(\log N_\tau)^{\frac{9(m-1)(d-1)}{10}}\right).\]
Moreover,
because $N_{\tau} \leq N < N_{\tau+1}$ and $N_{\tau}\asymp 2^{\tau}\tau^{d}$ and $N_{\tau+1}\asymp 2^{\tau+1}(\tau+1)^d$ have the same order
by Lemma~\ref{lem:number-SG}, we can see that $N$ has the same order as $N_{\tau}$.
Therefore,
\begin{equation}\label{eq:EI-decomposition-mu}
    \|f-\widehat{f}_{N_\tau,\lambda}\|_{\ScrH_k}=\CalO_p\left(\sigma^{\frac{2(m-1)}{5}}N^{-\frac{m-1}{5}}|\log (\sigma N)|^{\frac{(m-1)(2d-1)}{10}}(\log N)^{\frac{9(m-1)(d-1)}{10}}\right).
\end{equation}

Note that
\[
U_2 = \BFk_N^\intercal(\cdot) (\BFK_N+ \delta_N^{-2} \sigma^2  \BFI_N )^{-1}(y(\CalS_N)-\widehat{f}_{N_\tau,\lambda}(\CalS_N)).
\]
Because  $y(\CalS_N)-\widehat{f}_{N_\tau,\lambda}(\CalS_N)$ is a collection of samples of $f-\widehat{f}_{N_\tau,\lambda}$,
$U_2$ may be recognized as the KRR estimator---with a regularization parameter that equals $\sigma^2/(N\delta_N^2)$---of $U_1$ based on these samples. So we can define
\[
P\coloneqq \|(f-\widehat{f}_{N_\tau,\lambda})-(\widetilde{f}_N-\widehat{f}_{N_\tau,\lambda})\|_n
\qq{and}
Q\coloneqq \|(f-\widehat{f}_{N_\tau,\lambda})-(\widetilde{f}_N-\widehat{f}_{N_\tau,\lambda})\|_{\ScrH_k},
\]
and run through the proofs in Proposition~\ref{prop:Convergence_KRR} again, which gives:
\begin{align}
    & P\leq \max\biggl\{C_2\sqrt{\lambda}\|f-\widehat{f}_{N_\tau,\lambda}\|_{\ScrH_k},\  C_5 \sigma n^{-\frac{1}{2}}\lambda^{-\frac{1}{4}}\left|\log (\sigma n^{-\frac{1}{2}}\lambda^{-\frac{1}{4}}) \right|^{\frac{2d-1}{4}}\biggr\},\label{eq:EI-decomposition-P}\\
    & Q\leq \max\biggl\{C_2\|f-\widehat{f}_{N_\tau,\lambda}\|_{\ScrH_k},\ C_6\sigma  n^{-\frac{1}{2}}\lambda ^{-\frac{3}{4}}\left|\log (\sigma n^{-\frac{1}{2}}\lambda^{-\frac{1}{4}} )\right|^{\frac{2d-1}{4}}\biggr\}.\label{eq:EI-decomposition-Q}
\end{align}
According to our assumption, we have the following identity of $\lambda$:
\begin{equation}
\label{eq:EI-decomposition-lambda}
    \lambda=N^{-1}\delta_N^{-2} \sigma^2 \asymp \sigma^{\frac{4}{2m+1}}N^{-\frac{2}{2m+1}}|\log (\sigma N)|^{\frac{2d-1}{2m+1}}(\log N)^{\frac{6(m-1)(1-d)}{5}}.
\end{equation}
We can substitute equation \eqref{eq:EI-decomposition-mu} and \eqref{eq:EI-decomposition-lambda} into equation \eqref{eq:EI-decomposition-P} and \eqref{eq:EI-decomposition-Q} to get:
\begin{align*}
    &P=\CalO_p\left(\sigma^{\frac{2m}{2m+1}}N^{-\frac{m}{2m+1}}|\log (\sigma N)|^{\frac{2d-1}{7-m}}(\log N)^{\frac{3m-3}{10}(d-1)}\right),\\
    &Q=\CalO_p\left(\sigma^{\frac{2(m-1)}{5}}N^{-\frac{m-1}{5}}|\log (\sigma N)|^{\frac{(m-1)(2d-1)}{10}}(\log N)^{\frac{9(m-1)(d-1)}{10}}\right).
\end{align*}
Similar to the proof of Proposition~\ref{prop:Convergence_KRR},
we invoke Corollary~\ref{cor:Gagliardo–Nirenberg} and Lemma \ref{lem:norming-ineqality}  to get
\begin{align*}
      \|f-\widetilde{f}_N\|_\infty={} &\|(f-\widehat{f}_{N_\tau,\lambda})-(\widetilde{f}_N-\widehat{f}_{N_\tau,\lambda})\|_\infty \\
      \leq {} & C\sqrt{\|(f-\widehat{f}_{N_\tau,\lambda})-(\widetilde{f}_N-\widehat{f}_{N_\tau,\lambda})\|_2\|(f-\widehat{f}_{N_\tau,\lambda})-(\widetilde{f}_N-\widehat{f}_{N_\tau,\lambda})\|_{\ScrH_k}}\nonumber\\
      \leq{} & C\sqrt{\bigl((\log N)^{\frac{3(d-1)}{2}}P+N^{-1}(\log N)^{2(d-1)}Q\bigr)Q}\nonumber\\
      ={} &\CalO_p\biggl(\sigma^{\frac{2m-1}{2m+1}}N^{-\frac{2m-1}{4m+2}}|\log (\sigma N)|^{\frac{(2m-1)(2d-1)}{4(2m+1)}}(\log N)^{\frac{(6m-3)(d-1)}{6m-2}}\biggr),\label{eq:EI-decomposition-upp-bound-Linfty}
\end{align*}
where $C$ is some constant. Therefore, we can get the final result:
\begin{align*}
   \E[f(\BFx^*)-f(\widehat{\BFx}^*_N)]
    ={} &\CalO\biggl(\sigma^{\frac{2m-1}{2m+1}}N^{-\frac{2m-1}{4m+2}}|\log (\sigma N)|^{\frac{(2m-1)(2d-1)}{4(2m+1)}}(\log N)^{\frac{(6m-3)(d-1)}{6m-2}}\biggr).
\end{align*}
\Halmos \endproof

\section{Fast Matrix Inversion for TM Kernels and TSG Designs}\label{sec:fast-computation}

Two steps in the Algorithm~\ref{alg:EI} that are computationally intensive.
One computes the KRR estimator at the end of Stage~1 of the algorithm,
while the other computes the expected improvement in each iteration of Stage~2 of the algorithm.
Both steps involve computing inverse matrices of the form $(\BFK + \BFSigma)^{-1}$, where $\BFSigma$ is a diagonal matrix.
Specifically, $\BFSigma = n\lambda \BFI$ for the KRR estimator \eqref{eq:KRR-estimator}, whereas
$\BFSigma = \varsigma^2 \BFI$ when computing the expected improvement that involves \eqref{eq:posterior-mean}--\eqref{eq:posterior-var}.

\cite{DingZhang21_ec} developed algorithms for fast computation of $(\BFK+\BFSigma)^{-1}$ when (i) the kernel $k$ is of the TM class and (ii) the design points form a TSG.
To make the present paper self-contained,
we summarize the algorithms below.

A critical property that stems from the joint use of TM kernels and TSG designs is not only is the resulting  $\BFK^{-1}$ is sparse, but also its non-zero entries can be calculated explicitly.
Once  $\BFK^{-1}$ is computed, one may apply the Woodbury matrix identity \cite[page~19]{HornJohnson12_ec}:
\[(\BFK+\BFSigma)^{-1} = \BFSigma^{-1} - \BFSigma^{-1}(\BFK^{-1} + \BFSigma^{-1})^{-1}\BFSigma^{-1}.\]
Note that $(\BFK^{-1}+\BFSigma^{-1})$ is a sparse matrix, so computing its inverse can benefit from sparse linear algebra.
The matrix multiplications involved in the above identity are also easy to compute because $\BFSigma$ is a diagonal matrix.
Hence, we focus on the computation of $\BFK^{-1}$.

Given a TM kernel,
the computation for the case of TSG designs in multiple dimensions is reduced---through several intermediate steps---to the computation for the case of one-dimensional grids.
There are four cases involved. In ascending order of generality, they are (i) one-dimensional grids, (ii) full grids, (iii) classical SGs, and (iv) TSGs,
resulting in Algorithms~\ref{alg:TM-1D}--\ref{alg:TM-TSG}, respectively.
The algorithm developed for a simpler case becomes a subroutine for a more general case.

\subsection{One-Dimensional Grids}
Suppose $d=$1.
Let $k(x, x') = p(x\wedge  x')q(x\vee x')$ be a TM kernel in one dimension.
Let $n\geq 3$, $x_0=-\infty$, $x_{n+1}=\infty$, and $\{x_1,\ldots,x_n\}$ be an increasing sequence.
Let $\SFp_0=\SFq_{n+1}=0$, $\SFp_{n+1}=\SFq_0=1$,
$\SFp_i = p(x_i)$, and $\SFq_i=q(x_i)$ for $i=1,\ldots,n$.
Then, Proposition~1 in \cite{DingZhang21_ec} asserts that  $\BFK^{-1}$ and $\BFK^{-1}\BFk(x)$ are specified as follows.

\begin{enumerate}[label=(\roman*)]
    \item
$\BFK^{-1}$ is a tridiagonal matrix, i.e., $(\BFK^{-1})_{i,i+2}= (\BFK^{-1})_{i+2,i} =0 $ for all $i=1,\ldots,n-2$.
Moreover,
\begin{equation}\label{eq:K-inverse}
\begin{array}{ll}
    (\BFK^{-1})_{i,i} = \displaystyle \frac{\SFp_{i+1}\SFq_{i-1}-\SFp_{i-1}\SFq_{i+1}}{(\SFp_i\SFq_{i-1}-\SFp_{i-1}\SFq_i)(\SFp_{i+1}\SFq_i-\SFp_i\SFq_{i+1})}, &\quad i = 1, \ldots, n, \\[2ex]
    (\BFK^{-1})_{i,i+1} = (\BFK^{-1})_{i+1,i} = \displaystyle \frac{-1}{\SFp_{i+1} \SFq_{i}-\SFp_{i}\SFq_{i+1}},&\quad  i = 1, \ldots, n-1.
\end{array}
\end{equation}
\item Let $i^*=0,1,\ldots,n$ such that $x\in [x_{i^*}, x_{i^*+1})$. Then,
\begin{equation}\label{eq:K-inverse-times-k}
(\BFK^{-1}\BFk(x))_i = \left\{
\begin{array}{ll}
\displaystyle\frac{\SFp_{i^*+1}q(x) - p(x) \SFq_{i^*+1} }{\SFp_{i^*+1}\SFq_{i^*} - \SFp_{i^*} \SFq_{i^*+1} },     &  \mbox{ if } i = i^*,\\[2ex]
\displaystyle\frac{p(x)\SFq_{i^*} - \SFp_{i^*} q(x) }{\SFp_{i^*+1}\SFq_{i^*} - \SFp_{i^*} \SFq_{i^*+1} },     & \mbox{ if } i=i^*+1, \\[2ex]
0, & \mbox{ otherwise}.
\end{array}
\right.
\end{equation}
\end{enumerate}

\begin{algorithm}[ht]
\SingleSpacedXI
\SetAlgoLined
\DontPrintSemicolon
\SetKwInOut{Input}{Input}
\SetKwInOut{Output}{Output}
\Input{TM kernel $k$ in one dimension, design points $x_1<\ldots<x_n$, and prediction point $x$ }
\Output{$\BFK^{-1}$ and $\BFK^{-1}\BFk(x)$}
\BlankLine
Initialize $\BFA \leftarrow \BFzero\in\Real^{n\times n}$ and $\BFb \leftarrow \BFzero \in \Real^{n \times 1}$ \;
Update the tridiagonal entries of $\BFA$ using \eqref{eq:K-inverse}\;
Search for $i^*\in\{0,1,\ldots,n\}$ such that $x\in [x_{i^*}, x_{i^*+1})$, where $x_0 = -\infty$ and $x_{n+1}=\infty$ \;
Update $\BFb_{i^*}$ and $\BFb_{i^*+1}$ using \eqref{eq:K-inverse-times-k}\;
Return $\BFK^{-1}\leftarrow \BFA$ and $\BFK^{-1}\BFk(x)\leftarrow \BFb$\;
\caption{Computing $\BFK^{-1}$ and $\BFK^{-1}\BFk(x)$ for One-Dimensional Grids}\label{alg:TM-1D}
\end{algorithm}

\subsection{Full Grids}

By definition, TM kernels are in the tensor product form.
It is straightforward to generalize the computation of $\BFK^{-1}$ and $\BFK^{-1}\BFk(\BFx)$ from one-dimensional grids to multidimensional full grids.
In Algorithm~\ref{alg:TM-lattice},
$\BFK_j$ denotes the matrix composed of $k_j(x,  x')$ for all $x, x'\in \CalX_j$, $\BFk_j(x)$ denotes the vector composed of entries $k_j(x,  x')$ for all $  x' \in \CalX_j $, and
$\mathrm{vec}(\cdot)$ denotes the vectorization of a matrix.

\begin{algorithm}[ht]
\SingleSpacedXI
\SetAlgoLined
\DontPrintSemicolon
\SetKwInOut{Input}{Input}
\SetKwInOut{Output}{Output}
\Input{TM kernel $k$ in $d$ dimensions,  full grid design  $\CalX=\bigtimes_{j=1}^d \CalX_j$ with $\abs{\CalX_j}=n_j$, and prediction point $\BFx=(x_1,\ldots,x_d)$ }
\Output{$\BFK^{-1}$ and $\BFK^{-1}\BFk(\BFx)$}
\BlankLine
\For{$j\leftarrow 1$ \KwTo $d$}{
Compute $\BFK_j^{-1} \in \Real^{n_j\times n_j}$ and $\BFK_j^{-1}\BFk_j(x_j)\in \Real^{n_j\times 1}$ via Algorithm~\ref{alg:TM-1D} with inputs $(k_j,\CalX_j, x_j)$ \;
}
Return $\BFK^{-1} = \bigotimes_{j=1}^d \BFK_j^{-1}$ and
$\BFK^{-1}\BFk(\BFx) = \mathrm{vec}\left(\bigotimes_{j=1}^d \BFK_j^{-1}\BFk_j(x_j) \right)$ \;
\caption{Computing $\BFK^{-1}$ and $\BFK^{-1}\BFk(\BFx)$ for Full Grids}\label{alg:TM-lattice}
\end{algorithm}

\subsection{Classical  Sparse  Grids}

Given the nested sequence  $\{\CalX_{j,l}:j=1,\ldots,d,\; l=1,\ldots,\tau\}$ in \eqref{eq:dyadic},
for any  multi-index $\BFl=(l_1,\ldots,l_j)\in\NatInt^d$,
we define $\CalX^{\mathsf{FG}}_{\BFl}\coloneqq \CalX_{1,l_1} \times \cdots \times \CalX_{d,l_d}$.
From the definition of classical SGs in \eqref{eq:SG},
we can see that
\[
\CalX^{\mathsf{SG}}_\tau = \bigcup_{\abs{\BFl}\leq \tau+d-1} \CalX_{\BFl}^{\mathsf{FG}}.
\]
This drives the updating schemes in Algorithm~\ref{alg:TM-SG}.

Let
$\BFK_{\BFl} \coloneqq k(\CalX_{\BFl}^{\mathsf{FG}},\CalX_{\BFl}^{\mathsf{FG}})$
and $\BFk_\BFl(\BFx) \coloneqq k(\CalX_{\BFl}^{\mathsf{FG}}, \{\BFx\}) $.
For any matrix $\BFA\in\Real^{n\times n}$ with $n=\abs{\CalX^{\mathsf{SG}}_\tau}$ whose entries are indexed by $(\BFx,\BFx')$ for $\BFx,\BFx'\in\CalX_\tau^{\mathsf{SG}}$, let $\BFA_\BFl$  denote the part of $\BFA$ having entries indexed by $(\BFx,\BFx')$ for all $\BFx,\BFx'\in \CalX_\BFl^{\mathsf{FG}}$. Let  $\BFb_\BFl$ denote the subvector  of any vector $\BFb\in\Real^n$ in a similar manner.
In  each iteration $\BFl$ of Algorithm~\ref{alg:TM-SG}, the part of $\BFK^{-1}$ that corresponds to $\CalX_\BFl^{\mathsf{FG}}$ is updated, as is $\BFK^{-1}\BFk(\BFx)$.
Note that both $\BFK_{\BFl}$ and $\BFk_\BFl(\BFx)$ are defined on a full grid.
Thus, they can be computed via Algorithm~\ref{alg:TM-lattice}.
\begin{algorithm}[ht]
\SingleSpacedXI
\SetAlgoLined
\DontPrintSemicolon
\SetKwInOut{Input}{Input}
\SetKwInOut{Output}{Output}
\Input{TM kernel $k$ in $d$ dimensions,  classical SG design  $\CalX_\tau^{\mathsf{SG}}$ with size $n$, and prediction point $\BFx$ }
\Output{$\BFK^{-1}$ and $\BFK^{-1}\BFk(\BFx)$}
\BlankLine
Initialize $\BFA\leftarrow \BFzero \in\Real^{n\times n}$ and $\BFb\leftarrow \BFzero\in\Real^{n\times 1}$ \;
\For{all $\BFl\in\NatInt^d$ with $ \tau\leq \abs{\BFl} \leq \tau+d-1$ }{
Compute $\BFK_\BFl^{-1}$ and $\BFK_\BFl^{-1}\BFk_\BFl(\BFx)$ via Algorithm~\ref{alg:TM-lattice} with inputs   $(k,\bigtimes_{j=1}^d \CalX_{j, l_j}, \BFx) $ \;
Update $\BFA_\BFl$ and $\BFb_\BFl$ via
\vspace{-2ex}
\begin{align}
\BFA_\BFl \leftarrow{}& \BFA_\BFl + (-1)^{\tau+d-1-\abs{\BFl}} \binom{d-1}{\tau+d-1-\abs{\BFl}} \BFK_\BFl^{-1} \label{eq:update-A}\\[0.5ex]
\BFb_\BFl \leftarrow{}& \BFb_\BFl + (-1)^{\tau+d-1-\abs{\BFl}} \binom{d-1}{\tau+d-1-\abs{\BFl}} \BFK_\BFl^{-1} \BFk_\BFl(\BFx)\label{eq:update-b}
\end{align}
\vspace{-2ex}
}
Return $\BFK^{-1} \leftarrow \BFA$ and
$\BFK^{-1}\BFk(\BFx) \leftarrow \BFb$ \;
\caption{Computing $\BFK^{-1}$ and $\BFK^{-1}\BFk(\BFx)$ for Classical Sparse Grids}\label{alg:TM-SG}
\end{algorithm}

\subsection{Truncated Sparse Grids}

Let $\SFc_{l, i}\coloneqq i\cdot 2^{-l}$ for $l\geq 1$ and $i=1,\ldots,2^l -1$,
and let
$\BFc_{\BFl,\BFi}\coloneqq (\SFc_{l_1, i_1},\ldots, \SFc_{l_d, i_d})$.
By definition, a TSG of size $n$ is the union of two disjoint sets:
$\CalX^{\mathsf{TSG}}_n = \CalX^{\mathsf{SG}}_\tau \cup \mathcal{A}_{\Tilde{n}}$. Here, $\CalX^{\mathsf{SG}}_\tau$ is the classical SG of level $\tau$ such that
$\abs{\CalX^{\mathsf{SG}}_\tau}\leq n <\abs{\CalX^{\mathsf{SG}}_{\tau+1}}$, and  $\mathcal{A}_{\Tilde{n}}$ is a size-$\Tilde{n}$ subset of $\CalX_{\tau+1}^{\mathsf{SG}} \setminus \CalX_{\tau}^{\mathsf{SG}} = \{\BFc_{\BFl,\BFi}: \abs{\BFl}=\tau+d, \BFi\in\rho(\BFl)\}$,
where $\Tilde{n}=n-\abs{\CalX^{\mathsf{SG}}_\tau}$ and $\rho(\BFl)$ is defined in \eqref{eq:index-set-rho-ec}.

Given a TM kernel $k(\BFx,\BFx')=\prod_{j=1}^d p_j(x_j\wedge x'_j) q_j(x_j\vee x'_j)$ and a TSG $\CalX_n^{\mathsf{TSG}}$,
Theorem~5 in \cite{DingZhang21_ec} states that
 $\BFK^{-1}$  can  be expressed as the following block matrix with sparsity:
\begin{equation}\label{eq:K-inverse-TSG}
\BFK^{-1} =
\begin{pNiceMatrix}[first-row, columns-width=auto]
\scriptstyle{|\CalX_\tau^{\mathsf{SG}}|\;\,\mathrm{dim.}} & \scriptstyle{\Tilde{n} \;\,\mathrm{dim.}} \\[2pt]
\BFE & - \BFB \BFD \\[3pt]
- \BFD \BFB^\intercal &  \BFD
\end{pNiceMatrix},
\end{equation}
where
$\BFE = \BFA^{-1} + \BFB \BFD \BFB^\intercal$,
$\BFA = k(\CalX_\tau^{\mathsf{SG}},\CalX_\tau^{\mathsf{SG}})$,
$\BFB = \BFA^{-1} k(\CalX_\tau^{\mathsf{SG}},\CalA_{\tilde n})$,
and $\BFD$ is an $\tilde n\times \tilde n$ diagonal matrix whose diagonal entries are given by
\begin{equation}\label{eq:cond-var}
\BFD_{(\BFl,\BFi),(\BFl,\BFi)} =
\prod_{j=1}^d \left(\frac{\SFp_{j, l_j,i_j+1} \SFq_{j,l_j, i_j-1} - \SFp_{j,l_j,i_j-1} \SFq_{j,l_j,i_j+1}}{\left(\SFp_{j,l_j,i_j} \SFq_{j,l_j,i_j-1} - \SFp_{j,l_j,i_j-1}  \SFq_{j,l_j,i_j} \right) \left(\SFp_{j,l_j,i_j+1}  \SFq_{j,l_j,i_j} - \SFp_{j,l_j,i_j} \SFq_{j,l_j,i_j+1}\right) }\right)
\end{equation}
for all $(\BFl,\BFi)$ such that $\BFc_{\BFl,\BFi} \in \mathcal{A}_{\Tilde{n}}$,
where $\SFp_{j,l,i}= p_j(\SFc_{l, i})$ and
$\SFq_{j,l,i}= q_j(\SFc_{l, i})$ for all $j$, $l$, and $i$.

\begin{algorithm}[ht]
\SingleSpacedXI
\SetAlgoLined
\DontPrintSemicolon
\SetKwInOut{Input}{Input}
\SetKwInOut{Output}{Output}
\Input{TM kernel $k$ in $d$ dimensions,  TSG design  $\CalX_n^{\mathsf{TSG}} = \CalX_\tau^{\mathsf{SG}}\cup \mathcal{A}_{\Tilde{n}}$, observations $\bar{\BFy}$, and prediction point $\BFx$ }
\Output{$\BFK^{-1}$}
\BlankLine
With $\BFA\coloneqq k(\CalX_\tau^{\mathsf{SG}}, \CalX_\tau^{\mathsf{SG}})$, compute $\BFA^{-1}$ via Algorithm~\ref{alg:TM-SG} with inputs $(k, \CalX_\tau^{\mathsf{SG}})$\;
Initialize $\BFB \leftarrow \BFzero \in\Real^{|\CalX_{\tau}^{\mathsf{SG}}| \times \Tilde{n} }$ and $\BFD\leftarrow \BFzero \in\Real^{\Tilde{n}\times \Tilde{n}} $\;
\For{all $\BFx\in \CalA_{\Tilde{n}}$}{
Compute $\BFb \leftarrow \BFA^{-1}k(\CalX_{\tau}^{\mathsf{SG}}, \{\BFx\})$ via Algorithm~\ref{alg:TM-SG} with inputs $(k, \CalX_\tau^{\mathsf{SG}}, \BFx)$ \;
Update the $\BFx$-th column of $\BFB$ to $\BFb$\;
Update the $\BFx$-th diagonal entry of $\BFD$ via  \eqref{eq:cond-var} \;
}
Compute $\BFK^{-1}$ via \eqref{eq:K-inverse-TSG} \;
\caption{Computing $\BFK^{-1}$ for Truncated Sparse Grids}\label{alg:TM-TSG}
\end{algorithm}

%%%%%%%%%%%%%%%%%
\end{document}